\documentclass[dvipsnames]{article}

\usepackage[final]{lmrm}
\usepackage[utf8]{inputenc} 
\usepackage[T1]{fontenc}    
\usepackage{url}            
\usepackage{booktabs}       
\usepackage{amsfonts}       
\usepackage{nicefrac}       
\usepackage{booktabs} 
\usepackage{multirow}
\usepackage{wrapfig}
\usepackage{amssymb}
\usepackage{epigraph}
\usepackage{makecell}
\usepackage{listings}

\usepackage[labelfont=bf]{caption} 

\usepackage[ruled,vlined]{algorithm2e}
\usepackage{color, soul}
\usepackage{microtype}      
\usepackage{xcolor}         
\usepackage{geometry}
\usepackage{times}
\usepackage{ragged2e}
\usepackage{amsmath}
\usepackage{bm}
\usepackage{latexsym}
\usepackage{subfigure}
\usepackage{graphicx}
\usepackage{float}
\usepackage{longtable}
\usepackage{booktabs} 
\usepackage{multirow}
\usepackage{amssymb}
\usepackage{longtable}
\usepackage{tabu}
\usepackage{bbding}
\usepackage{awesomebox}
\usepackage{bbding}
\usepackage[most,breakable]{tcolorbox}
\usepackage{etoolbox}
\usepackage{fancyhdr}
\usepackage{lipsum}
\usepackage{CJKutf8}
\usepackage{pdfpages}
\usepackage{multicol}
\usepackage{pgffor} 
\usepackage{fontawesome5}
\usepackage{nicematrix} 
\usepackage{paralist}
\usepackage[edges]{forest}
\usepackage{siunitx}
\usepackage{tikz-dependency}
\usepackage{tikz}
\usepackage{bbm}
\usepackage{amssymb}
\usepackage[english]{babel}  
\usepackage{hyphenat}

\newcommand{\Rmnum}[1]{\expandafter\@slowromancap\romannumeral #1@}

\usepackage{caption} 
\usepackage{chngcntr}

\usepackage[colorlinks, linkcolor=RoyalBlue, anchorcolor=BrickRed, citecolor=RoyalBlue, urlcolor=RoyalBlue]{hyperref}

\usepackage{cleveref}
\crefname{section}{§}{§§}
\Crefname{section}{§}{§§}

\newcommand\refsec[1]{Section~\hyperref[sec:#1]{\ref{sec:#1}}}
\newcommand\refsecs[2]{\hyperref[sec:#1]{§\ref{sec:#1}:~\textsc{#1}}, \hyperref[sec:#2]{§\ref{sec:#2}:~\textsc{#2}}}

\usepackage[tikz]{bclogo}
\definecolor{msftBlue}{RGB}{0,164,239}
\definecolor{msftGreen}{RGB}{127,186,0}
\definecolor{msftYello}{RGB}{255,185,0}
\definecolor{mypurple}{RGB}{138,43,226} 
\definecolor{msftBlack}{RGB}{0,0,0}

\newtcolorbox{myboxnote}[1][]{
  breakable,
  title=#1,
  colback=cyan!0,
  colbacktitle=cyan!0,
  coltitle=black,
  fonttitle=\bfseries,
  bottomrule=0pt,
  toprule=0pt,
  leftrule=1.5pt,
  rightrule=1.5pt,
  titlerule=0pt,
  arc=0pt,
  outer arc=0pt,
  colframe=lightgray,
}
\usepackage{mdframed}

\definecolor{academicblue}{RGB}{54, 95, 145}

\newtcbox{\smybox}[1][red]{on line,
arc=1pt,colback=#1!10!white,colframe=#1!100!black,
before upper={\rule[-3pt]{0pt}{10pt}},
boxsep=0pt,left=6pt,right=6pt,top=2pt,bottom=0pt,boxrule=0pt,leftrule=1pt,rightrule=1pt}

\tcbset{
  iclrtakeawaybox/.style={
    width=\linewidth,
    colback=gray!5,                 
    colframe=gray!60,              
    colbacktitle=gray!30,            
    coltitle=black,                
    fonttitle=\bfseries,     
    boxrule=0.5pt,                 
    arc=2pt,                       
    sharp corners=south,           
    attach boxed title to top left={yshift=-0.1in,xshift=0.15in},
    boxed title style={boxrule=0pt, colframe=white},
    enhanced,
    drop shadow southeast           
  }
}
\newtcolorbox{TakeawayBox}[2][]{iclrtakeawaybox,title=#2,#1}

\newenvironment{itemsize*}%
 {\leftmargini=20pt\begin{itemize}%
  \setlength{\itemsep}{3pt}%
  \setlength{\parskip}{0pt}%
  }%
 {\end{itemize}}
\newenvironment{enumerate*}%
 {\begin{enumerate}%
  \setlength{\itemsep}{0pt}%
  \setlength{\parskip}{0pt}}%
 {\end{enumerate}}

\usepackage{fancyhdr} 
\usepackage{blindtext} 
\usepackage{ulem}

\usepackage{xparse}
\NewDocumentCommand{\heng}
{ mO{} }{\textcolor{red}{\textsuperscript{\textit{Heng}}\textsf{\textbf{\small[#1]}}}}

\usepackage{colortbl}

\title{
\textcolor{mypurple}{P}erception, \textcolor{mypurple}{R}eason, \textcolor{mypurple}{T}hink, and \textcolor{mypurple}{P}lan: \\ A Survey on Large Multimodal Reasoning Models
}

\author{
\small{Yunxin Li\thanks{Equal contribution $\ddagger$ Corresponding author, email: hubaotian@hit.edu.cn}\hspace{0.5em}, Zhenyu Liu$^*$, Zitao Li$^*$, Xuanyu Zhang, Zhenran Xu, Xinyu Chen, Haoyuan Shi, Shenyuan Jiang} \\ 
\small{\textbf{Xintong Wang, Jifang Wang, Shouzheng Huang, Xinping Zhao, Borui Jiang, Lanqing Hong, Longyue Wang}}\\  
\small{\textbf{Zhuotao Tian, Baoxing Huai, Wenhan Luo, Weihua Luo, Zheng Zhang, Baotian Hu$^\ddagger$, Min Zhang}}\\ 
\small{‌Harbin Institute of Technology, Shenzhen}\\
\footnotesize{Project: \textcolor{blue}{\url{https://github.com/HITsz-TMG/Awesome-Large-Multimodal-Reasoning-Models}}}
}

\begin{document}

\maketitle

\begin{abstract}

Reasoning lies at the heart of intelligence, shaping the ability to make decisions, draw conclusions, and generalize across domains. In artificial intelligence, as systems increasingly operate in open, uncertain, and multimodal environments, reasoning becomes essential for enabling robust and adaptive behavior. 
Large Multimodal Reasoning Models (LMRMs) have emerged as a promising paradigm, integrating modalities such as text, images, audio, and video to support complex reasoning capabilities. It aims to achieve comprehensive perception, precise understanding, and deep reasoning. 
As research advances, multimodal reasoning has rapidly evolved from modular, perception-driven pipelines to unified, language-centric frameworks that offer more coherent cross-modal understanding. While instruction tuning and reinforcement learning have improved model reasoning, significant challenges remain in omni-modal generalization, reasoning depth, and agentic behavior. 
To address these issues, we present a comprehensive and structured survey of multimodal reasoning research, organized around a four-stage developmental roadmap that reflects the field’s shifting design philosophies and emerging capabilities. 
First, we review early efforts based on task-specific modules, where reasoning was implicitly embedded across stages of representation, alignment, and fusion. 
Next, we examine recent approaches that unify reasoning into multimodal LLMs, with advances such as Multimodal Chain-of-Thought (MCoT) and multimodal reinforcement learning enabling richer and more structured reasoning chains. 
Finally, drawing on empirical insights from challenging benchmarks and experimental cases of OpenAI O3 and O4-mini,  we discuss the conceptual direction of native large multimodal reasoning models (N-LMRMs), which aim to support scalable, agentic, and adaptive reasoning and planning in complex, real-world environments. 
By synthesizing historical trends and emerging research, this survey aims to clarify the current landscape and inform the design of next-generation multimodal reasoning systems.
\end{abstract}

\begin{figure}[htbp]
    \centering
    \includegraphics[width=0.72\linewidth]{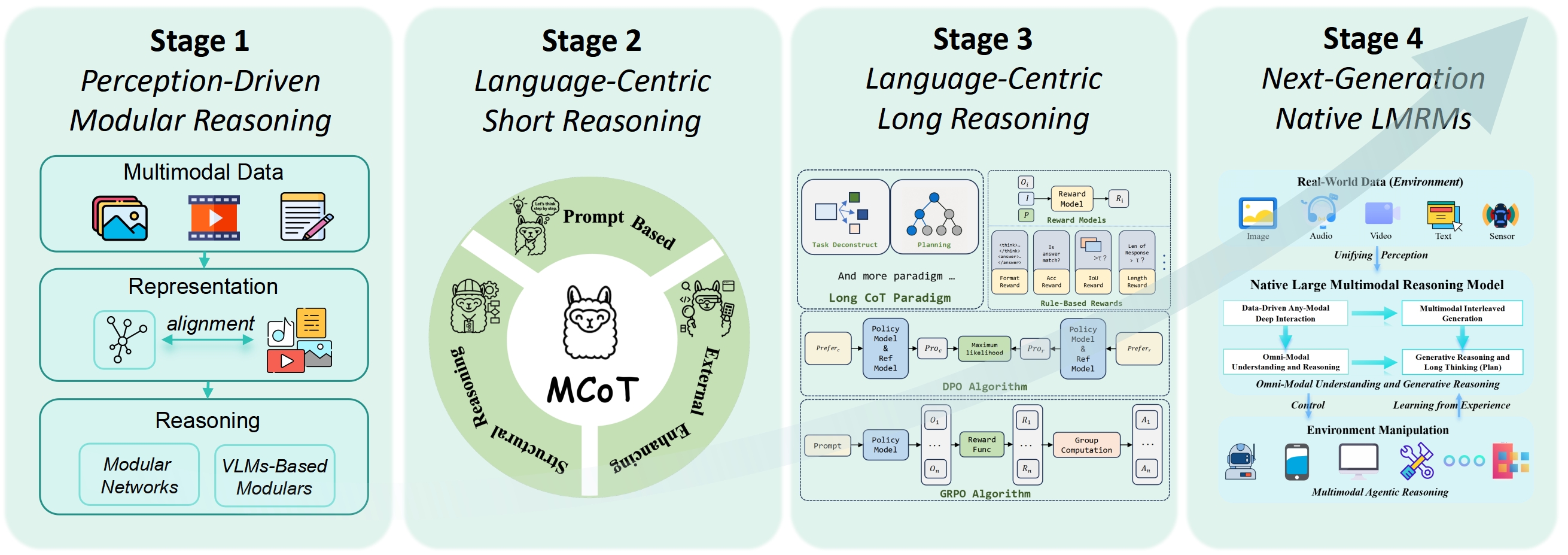}
    \caption{The core path of large multimodal reasoning models}
    \label{fig:enter-label}
\end{figure}

\newpage
{
  \hypersetup{linkcolor=RoyalBlue, linktoc=page}
  \tableofcontents
}

\newpage

\section{Introduction}
\label{sec:introduction}

In both philosophy and artificial intelligence, reasoning is widely regarded as a cornerstone of intelligent behavior~\citep{kahneman2011thinking,DBLP:journals/corr/abs-2410-09918,DBLP:journals/corr/abs-2410-07114,bi2025reasoning}. It enables agents not only to adaptively respond to their environments but also to draw logical inferences, generalize knowledge across diverse contexts, and navigate complex challenges. As AI systems increasingly interact with dynamic, uncertain, and multimodal settings, the ability to perform right reasoning under various environments becomes essential for achieving robust and adaptive intelligence~\citep{yang2025magma,DBLP:journals/corr/abs-2410-08328}.
In this context, Large Multimodal Reasoning Models (LMRMs) have emerged as a promising direction~\citep{wang2024exploring,DBLP:conf/acl/ZhangY0L0C024,yin2023survey}, which integrate multiple data modalities, such as text, images, audio, and video, and exhibit complex reasoning abilities, including logical deduction, causal inference, analogical mapping, and long-horizon thinking. The core objective of LMRMs is to enable \textit{\underline{comprehensive perception, precise understanding, and deep reasoning}}, supporting the decision-making process in diverse environments.

\begin{figure}[htbp]
    \centering
    \includegraphics[width=0.95\linewidth]{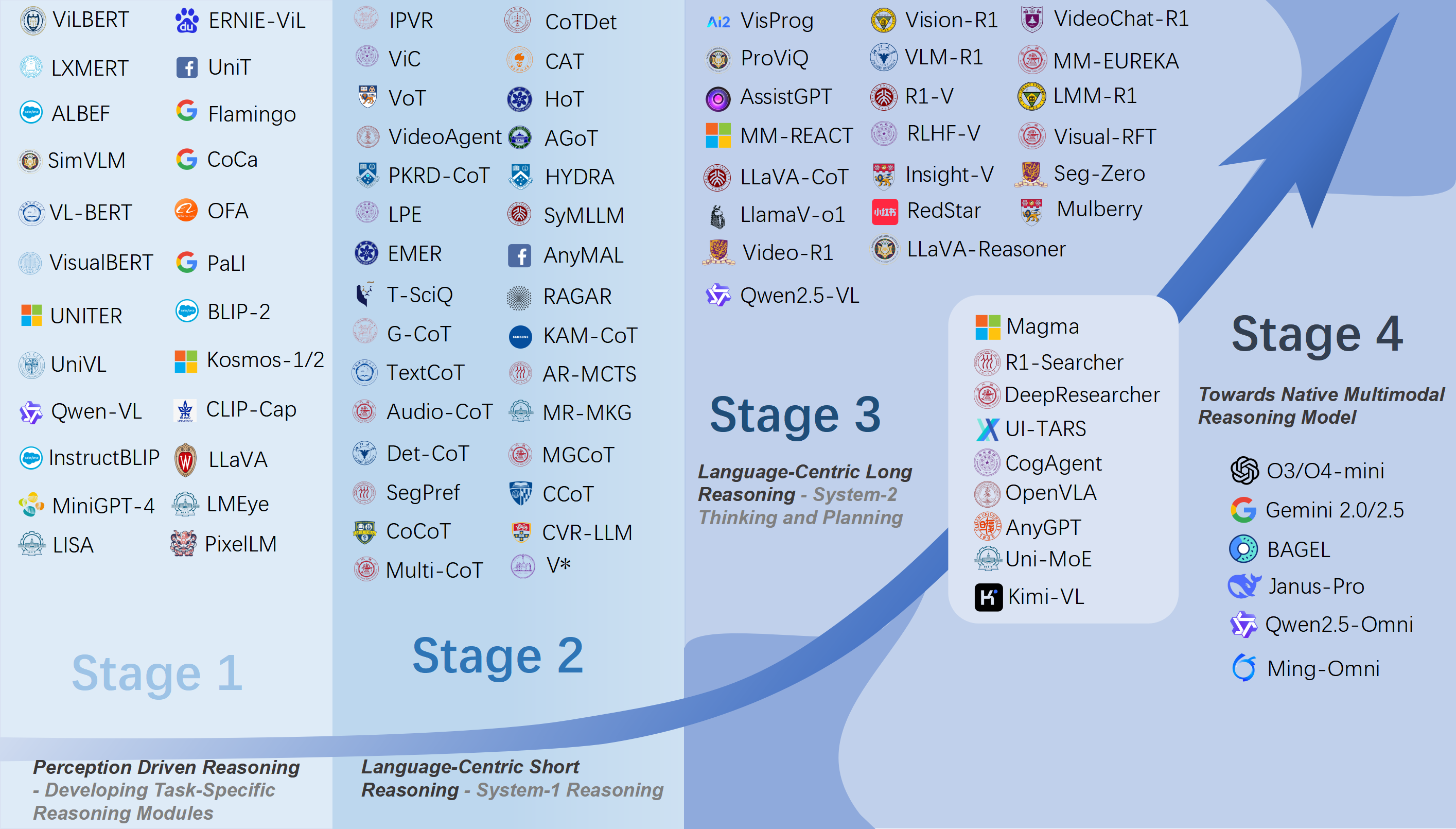}
    \caption{The roadmap of large multimodal reasoning models. The models highlighted in the box are representative models transitioning from Stage 3 towards Stage 4, as indicated by the directional arrow.}
    \label{fig:raodmap_timeline}
\end{figure}

Research in multimodal reasoning has progressed rapidly. Early efforts relied on perception-driven, modular pipelines, while recent advances leverage large language models to unify multimodal understanding and reasoning~\citep{DBLP:conf/nips/Huang0WHSML0MPL23,driess2023palm}. Instruction tuning~\citep{DBLP:conf/nips/LiuLWL23a} and reinforcement learning~\citep{DBLP:journals/corr/abs-2501-12948} further enhance models' reasoning performance, bringing them closer to human-like deliberative behaviour. Despite this rapid progress, multimodal reasoning is still a core bottleneck of large multimodal models, where they show limiting generalization, depth of reasoning, and agent-like behavior~\citep{DBLP:conf/cvpr/YueNZ0LZSJRSWYY24,DBLP:journals/corr/abs-2408-13257,DBLP:conf/eccv/LiuDZLZZYWHLCL24}.

Previous surveys in the field have largely focused on either multimodal large language models or the analysis of reasoning methods primarily centred on language, lacking a detailed analysis of recent reinforcement-enhanced multimodal reasoning and technical prospects of LMRMs. Hence, the multimodal reasoning area needs a coherent framework to understand how multimodal reasoning has evolved and where it is heading. Our work addresses a critical gap by providing a comprehensive review and analysis of the entire roadmap of multimodal reasoning models, encompassing early modular designs and state-of-the-art LMRMs. Furthermore, we project future developments of LMRMs, grounded in experimental findings and technical scrutiny.

Specifically, we propose a structured roadmap of multimodal reasoning, organized into three stages (Figure~\ref{fig:raodmap_timeline}): \textit{Perception-Driven Modular Reasoning}, where reasoning is implicit within task-specific modules; \textit{Language-Centric Short Reasoning (System-1)}, where multimodal reasoning emerges via prompt-based and structured short CoT with LLMs; and \textit{Language-Centric Long Reasoning (System-2)}, where long thinking, planning, and agentic behaviors are enabled through extended reasoning chains and reinforcement learning.

Building upon this developmental trajectory, we introduce the notion of \textit{Native Large Multimodal Reasoning Models (N-LMRMs)}, a forward-looking paradigm where reasoning is no longer retrofitted onto language models, but instead natively emerges from omnimodal perception and interaction, and goal-driven cognition. By grounding this vision in recent progress on unified representations, training data synthesis, learning from world experience, and benchmark construction, we outline possible directions for advancing multimodal intelligence beyond current architectural constraints.

\noindent {Our contributions are mainly threefold:}
\begin{itemsize*}
    \item This paper presents a comprehensive survey of the Large Multimodal Reasoning Model (LMRM) landscape, encompassing about 700 publications. Our analysis contextualizes and addresses key reasoning limitations in current models (Sec. \ref{sec:evolving}).
    
    \item We propose a three-stage roadmap for the development of LMRMs from modular reasoning to multimodal chain-of-thought (MCoT), and finally to long-horizon, system-2 reasoning. Each stage is further analyzed with detailed taxonomies and representative methods (Sec. \ref{roadmap}).

    \item We introduce and analyze Native Large Multimodal Reasoning Models (N-LMRMs), providing a thorough overview of initial progress, including architectures, learning methods, datasets, and benchmarks, thus setting the stage for future multimodal agentic reasoning (Sec. \ref{sec:native}).
    
    \item We reorganize existing datasets and benchmarks (update to 2025.06.30) of multimodal understanding and reasoning (Sec. \ref{sec:dataset}) to clarify their categories and evaluation dimensions.
\end{itemsize*}

\section{Evolving Paradigms of Multimodal Reasoning and Discussion}
\label{sec:evolving}

The evolution of multimodal reasoning has undergone a series of significant paradigm shifts, reflecting a deeper integration of perceptual inputs with structured cognitive processes. In this section, we outline \textbf{four} key stages in the development of multimodal reasoning systems, each embodying distinct model design, capabilities, and technical challenges. This historical perspective not only situates the current state of the field but also clarifies the motivations for the directions explored in later sections of this survey.

\textbf{Stage 1: \textcolor{mypurple}{Perception}-Driven Modular Reasoning - Designing Task-Specific Reasoning Systems} 

In the initial stage, multimodal reasoning capabilities were developed through modular, reasoning modules~\citep{andreas2016neural,DBLP:conf/cvpr/YangHGDS16,xiong2016dynamic}. These systems typically employed convolutional neural networks (CNNs) and recurrent architectures such as long short-term memory (LSTM) networks within supervised learning frameworks. Due to challenges such as limited multimodal data, immature neural architectures, and underdeveloped learning methodologies, early research adopted modular designs that decomposed the reasoning process into separate components: representation, alignment, fusion, and reasoning ({\hypersetup{linkcolor=Emerald}\S\ref{sec: stage1.1}}). As the field gradually shifted toward a pretraining-finetuning paradigm~\citep{devlin2018bert,radford2018improving,DBLP:conf/icml/RadfordKHRGASAM21}, the emergence of large-scale multimodal datasets and deeper neural networks facilitated the rise of pretrained vision–language models (VLMs)~\citep{DBLP:conf/eccv/ChenLYK0G0020,DBLP:conf/eccv/Li0LZHZWH0WCG20,DBLP:journals/tmlr/YuWVYSW22,yu2021ernie}, which aimed to unify the processes of representation, alignment, and fusion ({\hypersetup{linkcolor=Emerald}\S\ref{sec:vlms-basedmr}}). 

However, this unification primarily emphasized visual representation and cross-modal fusion, often at the expense of deeper semantic modelling of language. As a result, the reasoning process frequently defaulted to a classification-based paradigm, limiting context-aware and generalized reasoning. The multimodal reasoning systems still rely on additional modules or task-specific enhancements. Overall, reasoning at this stage remained largely implicit, primarily driven by foundational perceptual processing and neural computation. The emerging multimodal language models will enhance implicit reasoning by introducing powerful language models and large-scale visual data.

\textbf{Stage 2: Language-Centric Short Reasoning - System-1 \textcolor{mypurple}{Reasoning}}

The advent of multimodal large language models (MLLMs)~\citep{DBLP:conf/nips/LiuLWL23a,bai2023qwenvlversatilevisionlanguagemodel,chen2024internvl,zhang2023videollama} marked a pivotal shift in multimodal reasoning: moving from modular systems to end-to-end language-centric frameworks. These models achieved strong performance in tasks such as visual commonsense reasoning (VCR)~\citep{DBLP:conf/cvpr/ZellersBFC19,DBLP:conf/icml/YuYLWL0WW24}, visual question answering (VQA)~\citep{DBLP:conf/cvpr/GoyalKSBP17,DBLP:conf/cvpr/SinghNSJCBPR19}, and visual grounding~\citep{DBLP:journals/corr/abs-2306-14824,DBLP:conf/cvpr/Rasheed0MS0CAX024,DBLP:conf/eccv/LiuDZLZZYWHLCL24,lai2024lisa,DBLP:conf/cvpr/Rasheed0MS0CAX024,ren2024pixellm}.

However, early MLLM architectures largely relied on surface-level pattern matching and static knowledge retrieval, falling short in dynamic hypothesis generation, multi-step logical progression, and context-sensitive adaptation. 
This limitation catalyzed the development of Chain-of-Thought (CoT) reasoning~\citep{DBLP:conf/nips/KojimaGRMI22}, which transforms implicit reasoning into explicit intermediate steps, internalizing the thought processes within end-to-end generation. By aligning the representational capacity of Stage 1’s multimodal fusion with the linguistic expressiveness of LLMs, CoT enables more contextualized and interpretable reasoning.

Building on CoT’s success in pure language models, researchers extended it to the multimodal domain through the development of Multimodal Chain-of-Thought (MCoT)~\citep{zhang2023multimodal,fei2024video,zhang2023speechgpt,shao2024visualcotadvancingmultimodal}. 
Early approaches primarily focused on prompt-based adaptations ({\hypersetup{linkcolor=Emerald}\S\ref{sec:prompt}}), enabling models to produce step-by-step multimodal reasoning traces by carefully crafted instructions. 
Subsequent efforts enhanced the reasoning process itself, either by introducing structured decomposition of reasoning paths ({\hypersetup{linkcolor=Emerald}\S\ref{sec: structural_reasoning}}) or by leveraging external tools and retrieval augmentation to expand inference capabilities beyond the model's static knowledge ({\hypersetup{linkcolor=Emerald}\S\ref{sec: stage_2_3}}).

Nevertheless, reasoning at this stage predominantly remained short and reactive—characteristic of fast, intuitive System-1 reasoning. Models are effective for familiar or bounded tasks but struggle with abstraction, compositionality, and planning. These challenges spurred the development of more deliberate, structured reasoning paradigms, setting the stage for the next major transition.

\textbf{Stage 3: Language-Centric Long Reasoning - System-2 \textcolor{mypurple}{Thinking} and \textcolor{mypurple}{Planning}}

While MCoT has significantly advanced the reasoning capabilities of MLLMs, it remains insufficient for addressing the complexity of real-world multimodal tasks~\citep{DBLP:journals/corr/abs-2408-13257,DBLP:conf/icml/YuYLWL0WW24,DBLP:conf/cvpr/YueNZ0LZSJRSWYY24}.
Most MCoT methods operate through short, reactive chains—resembling fast, intuitive System-1 reasoning. These approaches are effective for familiar or bounded problems but struggle with abstraction, compositionality, long-horizon reasoning, and adaptive planning~\citep{DBLP:journals/corr/abs-2501-12948}. To bridge this gap, recent research has turned toward System-2-inspired reasoning~\citep{YaoYZS00N23,kahneman2011thinking}, emphasizing slower, deliberate, and methodologically structured cognitive processes. In this view, the reasoning is no longer treated as a mere function but as a core component of intelligent behaviour itself.
Extending MCoT along three critical dimensions--\textit{reasoning modalities}, \textit{reasoning paradigms}, and \textit{learning methods}--has become a key trajectory toward a new class of models: \textbf{Large Multimodal Reasoning Models (LMRMs)}, capable of deeper, transferable, and cognitively grounded reasoning.

First, from the perspective of reasoning modality, relying solely on textual representations constrains the model’s ability to capture modality-specific knowledge. Recent studies~\citep{lin2025investigating,gao2024interleaved,li2025imagine,zhou2024image,rose2023visual} introduce \textit{cross-modal reasoning chains} that leverage visual, auditory, and linguistic signals as joint substrates for inference, enabling richer semantic grounding and more faithful information integration ({\hypersetup{linkcolor=Emerald}\S\ref{sec: stage3.1}}).

Second, regarding reasoning paradigms, researchers construct longer, higher-quality chains and introduce generalized, methodologically guided reasoning strategies~\citep{DBLP:journals/corr/abs-2412-16720,yao2024mulberry}. These approaches allow models to autonomously decompose complex tasks and apply transferable procedures across diverse contexts. Notably, the O1 family (e.g., GPT-4o~\citep{hurst2024gpt4o}) exemplifies near-human-level performance on a broad range of cognitively demanding multimodal tasks ({\hypersetup{linkcolor=Emerald}\S\ref{sec: stage3.2}}).

Finally, from a learning method perspective, reinforcement learning-enhanced multimodal reasoning has gained increasing momentum. By incorporating agentic data, iterative feedback, and long-horizon optimization objectives, models like DeepSeek-R1~\citep{DBLP:journals/corr/abs-2501-12948} improve their planning, robustness, and adaptive generalization. This line of work has catalyzed the emergence of a new generation of R1-like models emphasizing scalable, methodologically grounded multimodal reasoning ({\hypersetup{linkcolor=Emerald}\S\ref{sec: stage3.3}}).

Together, these developments reflect a broader transition from reactive to deliberative reasoning paradigms, bringing LMRMs closer to achieving adaptive, system-level intelligence in open and dynamic environments.

\textbf{Stage 4: Towards Native Large Multimodal Reasoning Model (Prospect)}

While LMRMs show promise in addressing complex tasks through extended chains of thought, their language-centric architectures impose critical constraints \citep{kumar2025overthinking,pfister2025understanding}. First, their predominant focus on vision and language modalities (e.g., text, images, videos) limits their applicability in real-world settings, where diverse data types, such as audio, tactile signals, sensor streams, and temporal sequences, are deeply intertwined. Language-generated reasoning alone often struggles to support multimodal generative thinking, reflection, and control. Second, current models exhibit deficiencies in interactive, long-horizon reasoning and adaptive planning. Although they can produce extended reasoning chains in static settings, their ability to engage in real-time, iterative interaction with dynamic environments remains underdeveloped.

To address these gaps, we prospect the development of ‌\textbf{native large multimodal reasoning models (N-LMRMs)‌ as a potential paradigm shift in machine intelligence} ({\hypersetup{linkcolor=Emerald}\S\ref{sec:native}}). In contrast to conventional LMRMs, which retrofit language models with auxiliary modality processors, N-LMRMs will be natively designed to unify multimodal understanding, generation, and agentic reasoning within a fully end-to-end architecture. Real-world data types are encoded within a unified representation space, like VideoPoet \citep{videopoet2024}, BAGEL~\citep{DBLP:journals/corr/abs-2505-14683}, and Janus-Pro~\citep{chen2025janus}, while large-scale synthetic data facilitates holistic learning of reasoning and planning in the environment of any modality interaction. 
This evolution hinges on two transformative capabilities:
\textit{1) Multimodal Agentic Reasoning}:‌ N-LMRMs will embody agentic intelligence, enabling proactive, goal-driven interactions with complex environments, such as long-horizon planning---hierarchical task breakdown and memory-enhanced reasoning for coherence in extended interactions; dynamic adaptation---real-time strategy adjustment based on environmental feedback; embodied learning---closed-loop training frameworks enabling models to learn through simulated or physical interactions for better generalization. \textit{2) ‌Omni-Modal Understanding and Generative Reasoning}:‌
N-LMRMs will move beyond modality-specific encoders and decoders by utilizing a unified representational space for smooth cross-modal synthesis and analysis. This approach includes heterogeneous data fusion for the joint embedding of diverse data types, contextual multimodal generation for the coherent creation of composite outputs, and modality-agnostic inference that enables adaptable processing pipelines for the task-agnostic handling of new or any cross-modal data.



Taken together, the evolution from modular perception-driven systems to emerging native multimodal reasoners outlines a clear trajectory toward more unifying, adaptive, comprehensive high-level AI systems. In the following sections, we provide a detailed analysis of each stage, its representative models, and the emerging research directions that shape the future of multimodal reasoning.

\section{Roadmap of Multimodal Reasoning Models}
\label{roadmap}

\subsection{Stage 1 Perception Driven Modular Reasoning - Developing Task-Specific Reasoning Modules}
\label{sec:stage1}

In the early stages of multimodal reasoning, constraints such as limited multimodal data, nascent neural network architectures, and less sophisticated learning methods led to the development of models tailored to specific tasks. These models typically employed distinct modules to achieve multimodal representation, alignment, fusion, and reasoning. According to the model architectures and learning approaches, these models can be summarized as modular reasoning networks and pretrained Vision-Language Models (VLMs) based modular reasoning.

\tikzstyle{my-box}=[
    rectangle,
    draw=gray!50,
    rounded corners,
    text opacity=1,
    minimum height=2em, 
    minimum width=5em,
    inner sep=3pt,
    inner ysep=6pt, 
    align=center,
    fill opacity=0.15,
    line width=0.5pt,
]

\tikzstyle{leaf}=[my-box, minimum height=2em,
    fill=gray!5, text=black, align=left, font=\normalsize,
    inner xsep=3pt,
    inner ysep=6pt, 
    line width=0.5pt,
]

\definecolor{c1}{RGB}{102,178,255} 
\definecolor{c2}{RGB}{255,153,153} 
\definecolor{c3}{RGB}{255,204,102} 
\definecolor{c4}{RGB}{153,221,153} 
\definecolor{c5}{RGB}{204,179,255} 
\definecolor{c7}{RGB}{153,221,214} 
\definecolor{c8}{RGB}{221,160,221} 
\definecolor{c9}{RGB}{255,179,207} 

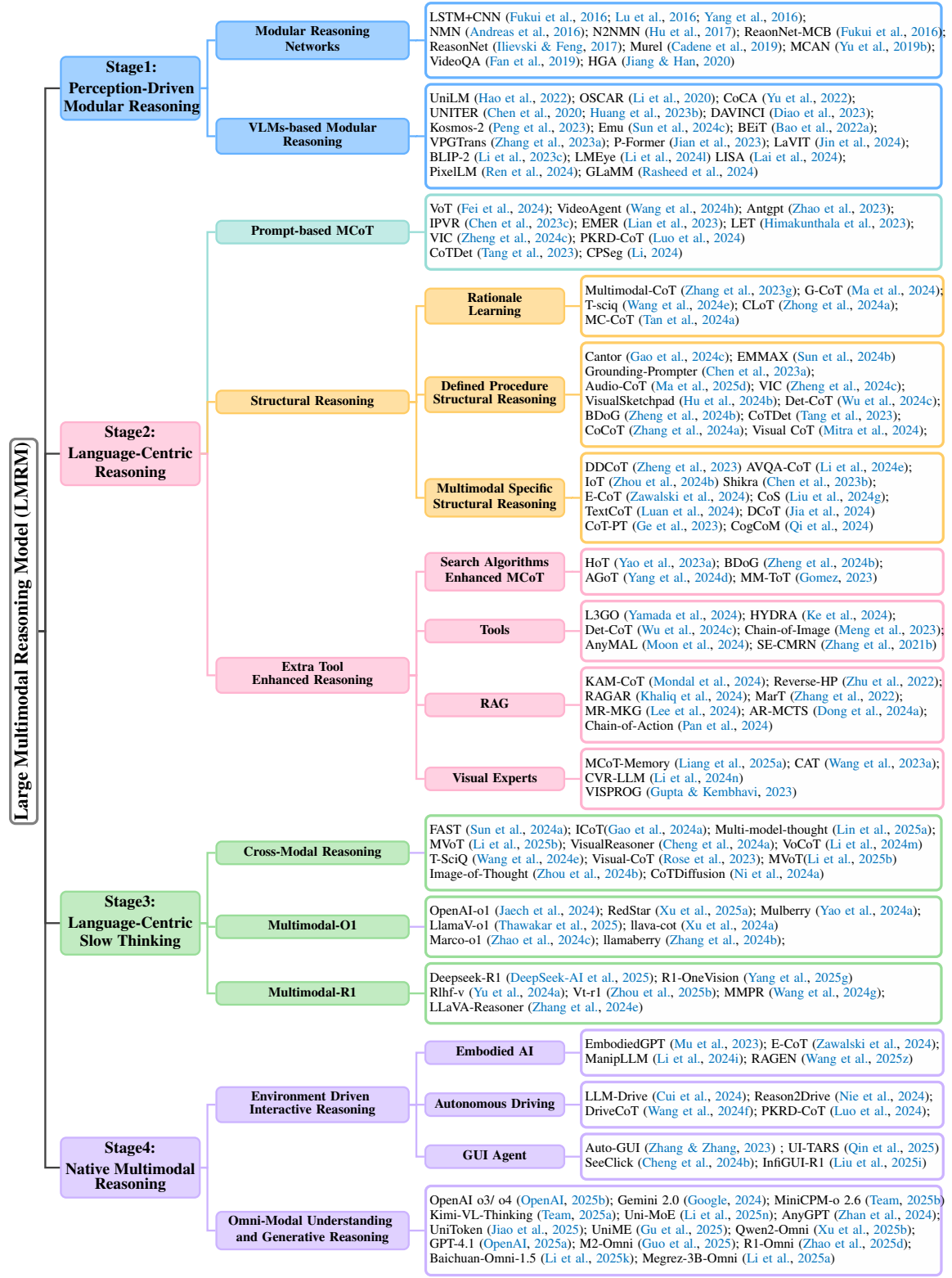
\begin{figure*}[!tp]
    \centering
    \resizebox{\textwidth}{!}{
        \begin{forest}
            forked edges,
            for tree={
                grow=east,
                reversed=true,
                anchor=base west,
                parent anchor=east,
                child anchor=west,
                base=center,
                font=\large,
                rectangle,
                draw=gray,
                rounded corners,
                align=left,
                text centered,
                minimum width=4em,
                edge+={darkgray, line width=0.5mm},
                s sep=3pt,
                inner xsep=2pt,
                inner ysep=3pt,
                line width=0.8pt,
                ver/.style={rotate=90, child anchor=north, parent anchor=south, anchor=center},
            },
            where level=1{text width=10em,font=\normalsize,}{},
            where level=2{text width=14em,font=\normalsize,}{},
            where level=3{text width=10em,font=\normalsize,}{},
            where level=4{text width=38em,font=\normalsize,}{}, 
            where level=5{text width=10em,font=\normalsize,}{},
            [
                \Large \textbf{Large Multimodal Reasoning Model (LMRM)}, ver, line width=0.7mm
                [
                    \large \shortstack{\textbf{Stage1:} \\[0.4ex] \textbf{Perception-Driven} \\ \textbf{Modular Reasoning}}, fill=c1!60, draw=c1, line width=0.5mm
                    [
                        \shortstack{\textbf{Modular Reasoning} \\ \textbf{Networks}}, 
                        fill=c1!60, draw=c1, line width=0.5mm, edge={c1}
                        [
                        LSTM+CNN \citep{fukui2016multimodal,lu2016hierarchical, DBLP:conf/cvpr/YangHGDS16}; \\ 
                        NMN \citep{andreas2016neural}; N2NMN \citep{hu2017learning}; ReaonNet-MCB \citep{fukui2016multimodal}; \\
                        ReasonNet \citep{DBLP:conf/nips/IlievskiF17}; Murel \citep{cadene2019murel}; MCAN \citep{yu2019deep}; \\
                        VideoQA \citep{DBLP:conf/cvpr/FanZZW0H19}; HGA \citep{jiang2020reasoning}
                        , leaf, text width=38em, draw=c1, line width=0.7mm, edge={c1} %
                        ]
                    ]
                    [
                        \shortstack{\textbf{VLMs-based Modular} \\ \textbf{Reasoning}}, 
                        fill=c1!60, draw=c1, line width=0.5mm, edge={c1}
                        [
                        UniLM \citep{DBLP:journals/corr/abs-2206-06336}; OSCAR \citep{DBLP:conf/eccv/Li0LZHZWH0WCG20}; CoCA \citep{DBLP:journals/tmlr/YuWVYSW22}; \\
                        UNITER \citep{DBLP:conf/eccv/ChenLYK0G0020,DBLP:conf/nips/Huang0WHSML0MPL23}; DAVINCI \citep{DBLP:conf/iclr/DiaoZZW23}; \\ 
                        Kosmos-2 \citep{DBLP:journals/corr/abs-2306-14824}; Emu \citep{DBLP:conf/iclr/SunYCZZWGL0W24}; BEiT \citep{DBLP:conf/iclr/Bao0PW22}; \\
                        VPGTrans \citep{DBLP:conf/nips/Zhang0Y00LC23}; P-Former \citep{DBLP:conf/nips/JianGV23}; LaVIT \citep{DBLP:conf/iclr/Jin0XCLTHCSMZOG24};  \\
                        BLIP-2 \citep{DBLP:conf/icml/0008LSH23}; LMEye \citep{DBLP:journals/tmm/LiHCMXZ24}
                        LISA \citep{lai2024lisa}; \\
                        PixelLM \citep{ren2024pixellm}; GLaMM \citep{DBLP:conf/cvpr/Rasheed0MS0CAX024}
                        , leaf, text width=38em, draw=c1, line width=0.7mm, edge={c1} %
                        ]
                    ]
                ]
                [   
                    \large \shortstack{\textbf{Stage2:} \\[0.4ex] \textbf{Language-Centric} \\ \textbf{Reasoning}}, fill=c9!60, draw=c9, line width=0.5mm
                    [
                        \textbf{Prompt-based MCoT}, align=center, fill=c7!60, draw=c7, line width=0.5mm, edge={c7}
                        [
                            VoT \citep{fei2024video}; VideoAgent \citep{wang2024videoagent}; Antgpt \citep{zhao2023antgpt}; \\ 
                            IPVR \citep{chen2023see}; EMER \citep{lian2023explainable}; LET \citep{himakunthala2023let}; \\ 
                            VIC \citep{zheng2024thinking}; PKRD-CoT \citep{luo2024pkrd} \\
                            CoTDet \citep{tang2023cotdet}; CPSeg \citep{li2024cpseg} 
                            ,leaf, text width=38em, draw=c7, line width=0.7mm, edge={c7} %
                        ]
                    ]
                    [
                        \textbf{Structural Reasoning}, fill=c3!60, draw=c3, line width=0.5mm, edge={c3}
                        [
                            \shortstack{\textbf{Rationale} \\ \textbf{Learning}}, fill=c3!60, draw=c3, line width=0.5mm, edge={c3}
                            [
                                Multimodal-CoT \citep{zhang2023multimodal}; G-CoT \citep{ma2024dolphins}; \\
                                T-sciq \citep{DBLP:conf/aaai/WangHHXLLS24}; CLoT \citep{zhong2024let};\\
                                MC-CoT \citep{tan2024boosting}
                            ,leaf, text width=26.5em, draw=c3, line width=0.7mm, edge={c3} %
                            ]
                        ]
                        [
                            \shortstack{\textbf{Defined Procedure} \\  \textbf{Structural Reasoning}}, fill=c3!60, draw=c3, line width=0.5mm, edge={c3}
                            [
                                Cantor \citep{gao2024cantor}; EMMAX \citep{sun2024emmax}\\
                                Grounding-Prompter \citep{chen2023grounding}; \\
                                Audio-CoT \citep{ma2025audio_cot}; VIC \citep{zheng2024thinking}; \\
                                VisualSketchpad \citep{hu2024visual}; Det-CoT \citep{wu2024dettoolchain}; \\
                                BDoG \citep{zheng2024picture}; CoTDet \citep{tang2023cotdet}; \\
                                CoCoT \citep{zhang2024cocot};
                                Visual CoT \citep{mitra2024compositional};
                            ,leaf, text width=26.5em, draw=c3, line width=0.7mm, edge={c3} %
                            ]
                        ]
                        [
                            \shortstack{\textbf{Multimodal Specific} \\ \textbf{Structural Reasoning}}, fill=c3!60, draw=c3, line width=0.5mm, edge={c3}
                            [
                                DDCoT \citep{zheng2023ddcot} AVQA-CoT \citep{li2024avqa_cot}; \\
                                IoT \citep{zhou2024image} Shikra \citep{chen2023shikra}; \\
                                E-CoT \citep{zawalski2024robotic}; CoS \citep{liu2024chain};\\
                                TextCoT \citep{luan2024textcot}; DCoT \citep{jia2024dcot} \\
                                CoT-PT \citep{ge2023chain}; CogCoM \citep{qi2024cogcom}
                            ,leaf, text width=26.5em, draw=c3, line width=0.7mm, edge={c3} %
                            ]
                        ]
                    ]
                    [
                         \shortstack{\textbf{Extra Tool} \\ \textbf{Enhanced Reasoning}}
                         , fill=c9!60, draw=c9, line width=0.5mm, edge={c9}
                        [
                            \shortstack{\textbf{Search Algorithms} \\ \textbf{Enhanced MCoT}},
                            fill=c9!60, draw=c9, line width=0.5mm, edge={c9}
                            [
                                    HoT \citep{yao2023HoT}; BDoG \citep{zheng2024picture}; \\
                                    AGoT \citep{yang2024AGoT}; MM-ToT \citep{gomez2023mmtot}
                                ,leaf, text width=26.5em, draw=c9, line width=0.7mm, edge={c9} %
                            ]
                        ]
                        [
                            \textbf{Tools}, fill=c9!60, draw=c9, line width=0.5mm, edge={c9}
                            [
                                L3GO \citep{yamada2024Co3D_Thoughts}; HYDRA \citep{ke2024hydra}; \\  Det-CoT \citep{wu2024dettoolchain};  
                                Chain-of-Image \citep{meng2023chain};\\
                                AnyMAL \citep{moon2024anymal}; SE-CMRN \citep{zhang2021explicit}
                                , leaf, text width=26.5em, draw=c9, line width=0.7mm, edge={c9} %
                            ]
                        ]
                        [
                            \textbf{RAG}, fill=c9!60, draw=c9, line width=0.5mm, edge={c9}
                            [
                                KAM-CoT \citep{mondal2024kam}; Reverse-HP \citep{zhu2022multimodal}; \\
                                RAGAR \citep{khaliq2024ragar}; MarT \citep{zhang2022multimodal}; \\
                                MR-MKG \citep{lee2024multimodal}; AR-MCTS \citep{dong2024progressive_retrieval}; \\
                                Chain-of-Action \citep{pan2024chain}
                                , leaf, text width=26.5em, draw=c9, line width=0.7mm, edge={c9} %
                            ]
                        ]
                        [
                            \textbf{Visual Experts}, fill=c9!60, draw=c9, line width=0.5mm, edge={c9}
                            [
                                MCoT-Memory \citep{liang2024memory_driven}; CAT \citep{wang2023caption}; \\
                                 CVR-LLM \citep{li2024enhancing}  \\
                                VISPROG \citep{gupta2023visual} \\
                                , leaf, text width=26.5em, draw=c9, line width=0.7mm, edge={c9} %
                            ]
                        ]
                    ]
                ] 
                [   
                    \large \shortstack{\textbf{Stage3:} \\[0.4ex] \textbf{Language-Centric} \\ \textbf{Slow Thinking}}, fill=c4!60, draw=c4, line width=0.5mm
                    [
                        \shortstack{\textbf{Cross-Modal Reasoning}}, 
                        align=center, fill=c4!60, draw=c4, line width=0.5mm, edge={c4}
                        [
                            FAST \citep{sun2024visual}; ICoT\citep{gao2024interleaved}; Multi-model-thought \citep{lin2025investigating}; \\ 
                            MVoT \citep{li2025imagine}; VisualReasoner \citep{cheng2024least}; VoCoT \citep{li2024vocot} \\
                            T-SciQ \citep{DBLP:conf/aaai/WangHHXLLS24};
                            Visual-CoT \citep{rose2023visual}; MVoT\citep{li2025imagine} \\
                            Image-of-Thought \citep{zhou2024image}; CoTDiffusion \citep{ni2024generate}
                            , leaf, text width=38em, draw=c4, line width=0.7mm, edge={c5} %
                        ]
                    ]
                    [
                        \textbf{Multimodal-O1}, fill=c4!60, draw=c4, line width=0.5mm, edge={c4}
                        [
                            OpenAI-o1 \citep{DBLP:journals/corr/abs-2412-16720}; RedStar \citep{xu2025redstar}; Mulberry \citep{yao2024mulberry};\\ LlamaV-o1  \citep{thawakar2025llamav_o1}; llava-cot \citep{xu2024llava_cot} 
                            \\
                            Marco-o1 \citep{DBLP:journals/corr/abs-2411-14405};
                            llamaberry \citep{zhang2024llamaberry}; 
                            \\
                            , leaf, text width=38em, draw=c4, line width=0.7mm, edge={c5} %
                        ]
                    ]
                    [
                        \textbf{Multimodal-R1}, fill=c4!60, draw=c4, line width=0.5mm, edge={c4}
                        [
                            Deepseek-R1 \citep{DBLP:journals/corr/abs-2501-12948}; R1-OneVision \citep{yang2025r1onevisionadvancinggeneralizedmultimodal} \\
                            Rlhf-v \citep{yu2024rlhf}; Vt-r1 \citep{zhou2025VisualThinker-R1-Zero}; MMPR \citep{DBLP:journals/corr/abs-2411-10442}; \\
                            LLaVA-Reasoner \citep{DBLP:journals/corr/abs-2410-16198}
                            , leaf, text width=38em, draw=c4, line width=0.7mm, edge={c4} 
                        ]
                    ]
                ] 
                [   
                     \large \shortstack{\textbf{Stage4:} \\[0.4ex] \textbf{Native Multimodal} \\ \textbf{Reasoning}}, fill=c5!60, draw=c5, line width=0.5mm
                    [
                        \shortstack{\textbf{Environment Driven } \\ \textbf{Interactive Reasoning}}, 
                        align=center, fill=c5!60, draw=c5, line width=0.5mm, edge={c5}
                        [
                             \textbf{Embodied AI}, 
                             align=center, fill=c5!60, draw=c5, line width=0.5mm, edge={c5}
                            [
                                EmbodiedGPT \citep{mu2023embodiedgpt};
                                E-CoT \citep{zawalski2024robotic}; \\
                                ManipLLM \citep{li2024manipllm}; RAGEN \citep{RAGEN} \\
                                ,leaf, text width=26.5em, draw=c5, line width=0.7mm, edge={c5} %
                            ]
                        ]
                        [
                             \textbf{Autonomous Driving}, 
                             align=center, fill=c5!60, draw=c5, line width=0.5mm, edge={c5}
                            [
                                LLM-Drive \citep{cui2024receive}; Reason2Drive \citep{nie2024reason2drive}; \\
                                DriveCoT \citep{wang2024drivecot}; PKRD-CoT \citep{luo2024pkrd}; 
                                ,leaf, text width=26.5em, draw=c5, line width=0.7mm, edge={c5} %
                            ]
                        ]
                        [
                             \textbf{GUI Agent}, 
                             align=center, fill=c5!60, draw=c5, line width=0.5mm, edge={c5}
                            [
                                Auto-GUI \citep{zhang2023CoAction} ; UI-TARS \citep{qin2025ui}\\
                                SeeClick \citep{cheng2024seeclick}; InfiGUI-R1 \citep{liu2025infiguir1}\\
                                ,leaf, text width=26.5em, draw=c5, line width=0.7mm, edge={c5} %
                            ]
                        ]
                    ]
                    [
                    \shortstack{\textbf{Omni-Modal Understanding } \\ \textbf{and Generative Reasoning}}, 
                        fill=c5!60, draw=c5, line width=0.5mm, edge={c5}
                        [
                        OpenAI o3/ o4~\citep{openai2025o3o4}; Gemini 2.0~\citep{Gemini2}; MiniCPM-o 2.6~\citep{team2025minicpm}\\
                        Kimi-VL-Thinking~\citep{kimiteam2025kimivltechnicalreport}; Uni-MoE~\citep{li2025uni};  
                        AnyGPT \citep{zhan2024anygpt}; \\ 
                        UniToken \citep{jiao2025unitoken};
                        UniME \citep{gu2025breakingmodalitybarrier}; Qwen2-Omni~\citep{xu2025qwen2}; \\
                        GPT-4.1~\citep{openai2025gpt41}; 
                        M2-Omni \citep{guo2025m2};
                        R1-Omni \citep{zhao2025r1omniexplainableomnimultimodalemotion}; \\
                        Baichuan-Omni-1.5 \citep{li2025baichuan};
                        Megrez-3B-Omni~\citep{li2025megrez} \\
                        , leaf, text width=38em, draw=c5, line width=0.7mm, edge={c5} %
                        ]
                    ]
                ]
            ]
        \end{forest}
    }
    \caption{Taxonomy of Large Multimodal Reasoning Models.}
    \label{fig:taxonomy}
\end{figure*}

\subsubsection{Modular Reasoning Networks\label{sec: stage1.1}}
\label{sec:Modular Reasoning Networks}
Initial approaches relied on generic CNN and LSTM backbones to derive answers from multimodal data. However, these were quickly improved by architectures that modularized reasoning based on perceptual cues. Neural Module Networks (NMN)~\citep{andreas2016neural} dynamically assembled task-specific modules to compose visual and textual features, replacing static fusion. Hierarchical Co-Attention (HieCoAtt)~\citep{lu2016hierarchical} introduced modular cross-modal attention to align question semantics with image regions hierarchically. Multimodal Compact Bilinear Pooling (MCB)~\citep{fukui2016multimodal} optimized feature interactions through efficient learnable bilinear modules. Stacked Attention Networks (SANs)~\citep{DBLP:conf/cvpr/YangHGDS16} modularized reasoning via iterative attention hops over visual features. Dynamic Memory Networks (DMN)~\citep{xiong2016dynamic} integrated memory modules for multi-episode reasoning over sequential inputs. ReasonNet~\citep{DBLP:conf/nips/IlievskiF17} decomposed reasoning into entity-relation modules for structured inference. UpDn~\citep{anderson2018bottom} introduced bottom-up and top-down attention to prioritize object-level features for reasoning (e.g., VQA-v2). MAC~\citep{hudson2018compositional} employed a memory-augmented control unit for iterative compositional reasoning. BAN~\citep{kim2018bilinear} captured high-order interactions using bilinear attention networks across modalities. Heterogeneous Memory Enhanced Multimodal Attention, HeteroMemory~\citep{DBLP:conf/cvpr/FanZZW0H19} extended modularity to video by synchronizing appearance and motion modules with temporal fusion. MuRel~\citep{cadene2019murel} modeled reasoning as a relational network over object pairs for fine-grained inference. MCAN~\citep{yu2019deep} used modular co-attention with self- and guided-attention for deep cross-modal reasoning.

These advancements illustrate how perception-driven designs - incorporating attention mechanisms, memory components, and compositional modules - facilitate fine-grained reasoning that is aligned with specific task requirements. However, the advent of Transformer~\citep{vaswani2017attention} architecture, coupled with pretraining-finetuning learning schemes, has propelled multimodal representation, alignment, and fusion. Specifically, Trasnformer-based pretrained VLMs enhance the integration of visual and textual information at the data and model interior, thus enabling perception-driven reasoning capabilities.

\subsubsection{Vision-Language Models-based Modular Reasoning}
\label{sec:vlms-basedmr}
These VLMs are trained with large-scale image-text pairs, advancing perception-driven reasoning tasks, like NLVR\(^2\)~\citep{suhr2018corpus}, TVQA~\citep{lei2018tvqa}, GQA~\citep{DBLP:conf/cvpr/HudsonM19}, OK-VQA~\citep{DBLP:conf/cvpr/MarinoRFM19}, VCR~\citep{DBLP:conf/cvpr/ZellersBFC19}, and ScienceQA~\citep{saikh2022scienceqa}. Specifically, VLMs introduced Transformer and employed large-scale image-text data to unify the process of multimodal representation, perception, fusion, and inference. Below are three kinds of pretrained VLMs-based modular reasoning:

\textbf{Dual-Encoder Contrastive Reasoning.} These models leverage dual-stream architectures with contrastive learning to dynamically align and reason over visual and textual features through cross-modal interactions. For example, ViLBERT~\citep{lu2019vilbert} uses dual-stream Transformers with cross-modal attention for dynamic feature alignment. LXMERT~\citep{tan2019lxmert} adds interaction layers between dual encoders to reason over relational embeddings. CLIP~\citep{DBLP:conf/icml/RadfordKHRGASAM21} leverages contrastive pretraining for zero-shot reasoning via aligned embeddings. ALBEF~\citep{li2021align} integrates contrastive learning with momentum distillation to reason over distilled embeddings. METER~\citep{dou2022empirical} enhances dual-stream reasoning with a modular encoder-decoder framework for robust alignment (e.g., VCR). SimVLM~\citep{wang2021simvlm} uses prefix-based pretraining to align vision and language for efficient reasoning. VLMo~\citep{bao2022vlmo} introduces a mixture-of-modality-experts framework for flexible cross-modal reasoning. CoCa~\citep{DBLP:journals/tmlr/YuWVYSW22} integrates contrastive and generative heads for versatile reasoning (e.g., NLVR\(^2\)). BLIP~\citep{li2022blip} introduce the image-text transformer module Q-former and employs vision-language pretraining with contrastive objectives to reason via bootstrapped alignment.

\begin{table}[t]
\centering
\renewcommand{\arraystretch}{1.5}
\caption{The classific works of the initial stage of perception-driven multimodal modular reasoning, where VLMs and MLLMs play a significant role in advancing the performance of multimodal reasoning tasks.}
\resizebox{\textwidth}{!}{
\begin{tabular}{l|c|c|c|c}
\toprule
\textbf{Model} & \textbf{Year} & \textbf{Architecture} & \textbf{Highlight} & \textbf{Training Method} \\
\hline
\multicolumn{5}{c}{\textbf{\textcolor{msftBlue}{Neural Modular Reasoning Networks}}}\\
\hline
\makecell[tl]{NMN~\citep{andreas2016neural}} & 2016 & Modular & Dynamically assembles task-specific modules for visual-textual reasoning. & Supervised learning \\
\hline
\makecell[tl]{HieCoAtt~\citep{lu2016hierarchical}} & 2016 & Attention-based & Aligns question semantics with image regions via hierarchical cross-modal attention. & Supervised learning \\
\hline
\makecell[tl]{MCB~\citep{fukui2016multimodal}} & 2016 & Bilinear & Optimizes cross-modal feature interactions with efficient bilinear modules. & Supervised learning \\
\hline
\makecell[tl]{SANs~\citep{DBLP:conf/cvpr/YangHGDS16}} & 2016 & Attention-based & Iteratively refines reasoning through multiple attention hops over visual features. & Supervised learning \\
\hline
\makecell[tl]{DMN~\citep{xiong2016dynamic}} & 2016 & Memory-based & Integrates memory modules for multi-episode reasoning over sequential inputs. & Supervised learning \\
\hline
ReasonNet~\citep{DBLP:conf/nips/IlievskiF17} & 2017 & Modular & Decomposes reasoning into entity-relation modules for structured inference. & Supervised learning \\
\hline
UpDn~\citep{anderson2018bottom} & 2018 & Attention-based & Combines bottom-up and top-down attention for object-level reasoning. & Supervised learning \\
\hline
MAC~\citep{hudson2018compositional} & 2018 & Memory-based & Uses a memory-augmented control unit for iterative compositional reasoning. & Supervised learning \\
\hline
BAN~\citep{kim2018bilinear} & 2018 & Bilinear & Captures high-order interactions via bilinear attention across modalities. & Supervised learning \\
\hline
\makecell[tl]{HeteroMemory~\citep{DBLP:conf/cvpr/FanZZW0H19}} & 2019 & Memory-based & Synchronizes appearance and motion modules for video-based temporal reasoning. & Supervised learning \\
\hline
MuRel~\citep{cadene2019murel} & 2019 & Relational & Models reasoning as a relational network over object pairs for fine-grained inference. & Supervised learning \\
\hline
MCAN~\citep{yu2019deep} & 2019 & Attention-based & Employs modular co-attention with self- and guided-attention for deep reasoning. & Supervised learning \\
\hline
\multicolumn{5}{c}{\textbf{\textcolor{msftBlue}{VLMs-based Modular Reasoning}}} \\
\hline
ViLBERT~\citep{lu2019vilbert} & 2019 & Dual-Encoder & Aligns visual-text features via dual-stream Transformers with cross-modal attention. & Pretraining + fine-tuning \\
\hline
LXMERT~\citep{tan2019lxmert} & 2019 & Dual-Encoder & Enhances cross-modal reasoning with dual-stream pretraining on diverse tasks. & Pretraining + fine-tuning \\
\hline
X-LXMERT~\citep{tan2019lxmert} & 2020 & Dual-Encoder & Extends dual-stream reasoning with generative cross-modal pretraining. & Pretraining + fine-tuning \\
\hline
ALBEF~\citep{li2021align} & 2021 & Dual-Encoder & Integrates contrastive learning with momentum distillation for robust reasoning. & Contrastive + generative pretraining \\
\hline
SimVLM~\citep{wang2021simvlm} & 2021 & Dual-Encoder & Uses prefix-based pretraining for flexible cross-modal reasoning. & Pretraining + fine-tuning \\
\hline
VLMo~\citep{bao2022vlmo} & 2022 & Dual-Encoder & Employs a mixture-of-modality-experts for dynamic cross-modal reasoning. & Pretraining + fine-tuning \\
\hline
METER~\citep{dou2022empirical} & 2022 & Dual-Encoder & Enhances reasoning with a modular encoder-decoder for robust alignment. & Pretraining + fine-tuning \\
\hline
BLIP~\citep{li2022blip} & 2022 & Dual-Encoder & Bootstraps alignment with contrastive learning for efficient reasoning. & Contrastive + generative pretraining \\
\hline
VisualBERT~\citep{li2019visualbert} & 2019 & Single-Transformer-Backbone & Fuses visual-text inputs in a single Transformer for joint contextual reasoning. & Pretraining + fine-tuning \\
\hline
VL-BERT~\citep{su2019vl} & 2019 & Single-Transformer-Backbone & Enhances cross-modal reasoning with unified visual-language pretraining. & Pretraining + fine-tuning \\
\hline
UNITER~\citep{DBLP:conf/eccv/ChenLYK0G0020} & 2020 & Single-Transformer-Backbone & Reasons via joint contextual encoding in a single Transformer backbone. & Pretraining + fine-tuning \\
\hline
PixelBERT~\citep{huang2020pixel} & 2020 & Single-Transformer-Backbone & Processes pixels with CNN+Transformer for fine-grained cross-modal reasoning. & Pretraining + fine-tuning \\
\hline
UniVL~\citep{luo2020univl} & 2020 & Single-Transformer-Backbone & Unifies video-language reasoning with a single Transformer for temporal tasks. & Pretraining + fine-tuning \\
\hline
Oscar~\citep{DBLP:conf/eccv/Li0LZHZWH0WCG20} & 2020 & Single-Transformer-Backbone & Anchors reasoning with object tags in a unified Transformer for semantic inference. & Pretraining + fine-tuning \\
\hline
VinVL~\citep{zhang2021vinvl} & 2021 & Single-Transformer-Backbone & Boosts reasoning with enhanced visual features in a single Transformer. & Pretraining + fine-tuning \\
\hline
ERNIE-ViL~\citep{yu2021ernie} & 2021 & Single-Transformer-Backbone & Integrates scene graph knowledge for structured visual-language reasoning. & Pretraining + fine-tuning \\
\hline
UniT~\citep{hu2021unit} & 2021 & Single-Transformer-Backbone & Streamlines multimodal tasks with a shared self-attention Transformer backbone. & Pretraining + fine-tuning \\
\hline
Flamingo~\citep{alayrac2022flamingo} & 2022 & Single-Transformer-Backbone & Prioritizes dynamic vision-text interactions via cross-attention. & Pretraining + fine-tuning \\
\hline
CoCa~\citep{DBLP:journals/tmlr/YuWVYSW22} & 2022 & Single-Transformer-Backbone & Combines contrastive and generative heads for versatile cross-modal reasoning. & Contrastive + generative pretraining \\
\hline
BEiT-3~\citep{wang2022image} & 2022 & Single-Transformer-Backbone & Unifies vision-language learning with masked data modeling. & Pretraining + fine-tuning \\
\hline
OFA~\citep{wang2022ofa} & 2022 & Single-Transformer-Backbone & Provides a unified multimodal framework for efficient cross-modal reasoning. & Pretraining + fine-tuning \\
\hline
PaLI~\citep{chen2022pali} & 2022 & Single-Transformer-Backbone & Scales reasoning with a multilingual single-Transformer framework. & Pretraining + fine-tuning \\
\hline
BLIP-2~\citep{DBLP:conf/icml/0008LSH23} & 2023 & Single-Transformer-Backbone & Uses a querying Transformer for improved cross-modal reasoning efficiency. & Pretraining + fine-tuning \\
\hline
Kosmos-1~\citep{DBLP:conf/nips/Huang0WHSML0MPL23} & 2023 & Single-Transformer-Backbone & Enables interleaved input processing for flexible multimodal understanding. & Pretraining + fine-tuning \\
\hline
Kosmos-2~\citep{DBLP:journals/corr/abs-2306-14824} & 2023 & Single-Transformer-Backbone & Enhances grounding capability for precise object localization and reasoning. & Pretraining + fine-tuning \\
\hline
CLIP-Cap~\citep{mokady2021clipcap} & 2021 & Vision-Encoder-LLM & Projects CLIP visual features into an LLM for reasoning and captioning. & Fine-tuning \\
\hline
LLaVA~\citep{DBLP:conf/nips/LiuLWL23a} & 2023 & Vision-Encoder-LLM & Tunes ViT-LLM integration for conversational multimodal reasoning. & Instruction tuning \\
\hline
MiniGPT-4~\citep{zhu2023minigpt} & 2023 & Vision-Encoder-LLM & Aligns ViT to a frozen LLM via projection for streamlined reasoning. & Fine-tuning \\
\hline
InstructBLIP~\citep{dai2023instructblip} & 2023 & Vision-Encoder-LLM & Uses instruction tuning to align ViT with LLM for multimodal reasoning. & Instruction tuning \\
\hline
Qwen-VL~\citep{bai2023qwenvlversatilevisionlanguagemodel} & 2023 & Vision-Encoder-LLM & Incorporates spatial-aware ViT for enhanced grounded reasoning. & Pretraining + fine-tuning \\
\hline
mPLUG-Owl~\citep{ye2023mplug} & 2023 & Vision-Encoder-LLM & Integrates modular visual encoder with LLM for instruction-following reasoning. & Instruction tuning \\
\hline
Otter~\citep{li2023otter} & 2023 & Vision-Encoder-LLM & Combines modular visual encoder with LLM for in-context multimodal reasoning. & Instruction tuning \\
\bottomrule
\end{tabular}}
\end{table}

\textbf{Single-Transformer-Backbone Interactive Reasoning.} This paradigm embeds visual and textual inputs in a single Transformer, enabling direct cross-modal reasoning through unified encoding method. VisualBERT~\citep{li2019visualbert}, UNITER~\citep{DBLP:conf/eccv/ChenLYK0G0020}, VL-BERT~\citep{su2019vl} fuse visual-text inputs in a single Transformer to reason via joint contextual encoding or enhanced cross-modal pretraining. PixelBERT~\citep{huang2020pixel} employs a CNN and Transformer architecture to process pixels for fine-grained reasoning (e.g., NLVR\(^2\)). UniVL~\citep{luo2020univl} unifies video-language reasoning with a single Transformer for temporal cross-modal tasks (e.g., TVQA). Oscar~\citep{DBLP:conf/eccv/Li0LZHZWH0WCG20}, VinVL~\citep{zhang2021vinvl} anchor reasoning with object tags or enhanced visual features in a unified Transformer, boosting semantic inference (e.g., VCR, GQA). ERNIE-ViL~\citep{yu2021ernie} integrates scene graph knowledge into a single Transformer, enhancing compositional reasoning through structured visual-language interactions. UniT~\citep{hu2021unit} streamlines multimodal tasks with a shared self-attention backbone for unified reasoning. PaLI~\citep{chen2022pali} scales single-Transformer reasoning with a multilingual framework for cross-lingual inference (e.g., OK-VQA). Flamingo~\citep{alayrac2022flamingo} employs cross-attention to prioritize dynamic vision-text interactions. BEiT-3~\citep{wang2022image} adopts masked data modeling to unify vision-language learning. OFA~\citep{wang2022ofa}, BLIP-2~\citep{DBLP:conf/icml/0008LSH23} introduce a unified multimodal framework or a querying Transformer to excel in cross-modal reasoning with improved efficiency (e.g., VQA-v2). Kosmos-1~\citep{DBLP:conf/nips/Huang0WHSML0MPL23}, Kosmos-2~\citep{DBLP:journals/corr/abs-2306-14824} enable interleaved input processing or grounding capability for flexible multimodal understanding and precise object localization.

\textbf{Multimodal LLMs-based Implicit Reasoning.} This approach projects visual inputs into a large language model’s text space, leveraging the contextual reasoning capabilities of large language models \citep{li2023multi} to improve the performance of multimodal reasoning. Their architecture contains pretrained visual encoders and large language models, arr. \textit{Vision-Encoder-LLM}. CLIP-Cap~\citep{mokady2021clipcap} projects CLIP visual features into an LLM for reasoning and captioning tasks. LLaVA~\citep{DBLP:conf/nips/LiuLWL23a} enables conversational reasoning by tuning ViT-LLM integration for interactive tasks or scaling for complex VQA tasks. MiniGPT-4~\citep{zhu2023minigpt}, InstructBLIP~\citep{dai2023instructblip} align a ViT to a frozen LLM via a projection layer or instruction tuning, streamlining visual-text reasoning. Qwen-VL~\citep{bai2023qwenvlversatilevisionlanguagemodel} incorporates a spatial-aware ViT, enhancing grounded reasoning for spatially complex tasks. mPLUG-Owl~\citep{ye2023mplug}, LMEye~\citep{DBLP:journals/tmm/LiHCMXZ24}, and  Otter~\citep{li2023otter} integrate a modular visual encoder with an LLM for instruction-following and in-context learning for multimodal reasoning.

While the architectural innovations of these three kinds of models have significantly advanced multimodal reasoning for tasks, their reliance on predefined feature alignments or contextual encodings often limits their ability to handle complex, multi-step reasoning scenarios requiring iterative or compositional inference. These constraints highlight the need for Multimodal Chain-of-Thought (MCoT) reasoning (Sec. \ref{sec:stage2}) in large-scale models like the development of LLMs, which can dynamically decompose tasks, integrate intermediate reasoning steps, and adaptively align perception and inference for more robust and generalizable performance across diverse multimodal challenges.

\begin{TakeawayBox}{Takeaways: Perception-Driven Modular Reasoning}
Early multimodal models primarily focused on the representation, alignment, and fusion of information. Reasoning in these models was often implicit, typically requiring separate, task-specific reasoning modules. More recently, multimodal large language models, particularly those adopting a vision encoder-language model structure, have achieved a unified multimodal reasoning architecture and demonstrated improved multi-task reasoning performance.
\end{TakeawayBox}

\subsection{Stage 2 Language-Centric Short Reasoning - System-1 Reasoning}
\label{sec:stage2}

With the advent of large-scale multimodal pretraining, MLLMs have started to demonstrate emergent reasoning capabilities. However, such inferences are often shallow, relying primarily on implicit correlations rather than explicit logical processes. MCoT has emerged as a simple yet effective approach to mitigate this limitation. By incorporating intermediate reasoning steps, MCoT improves cross-modal alignment, knowledge integration, and contextual grounding, all without the need for extensive supervision or significant architectural modifications. In this stage, we categorize existing approaches into three paradigms: prompt-based MCoT, structural reasoning with predefined patterns, and tool-augmented reasoning with lightweight external modules.

\begin{figure}[htbp]
    \centering
    \includegraphics[width=1\linewidth]{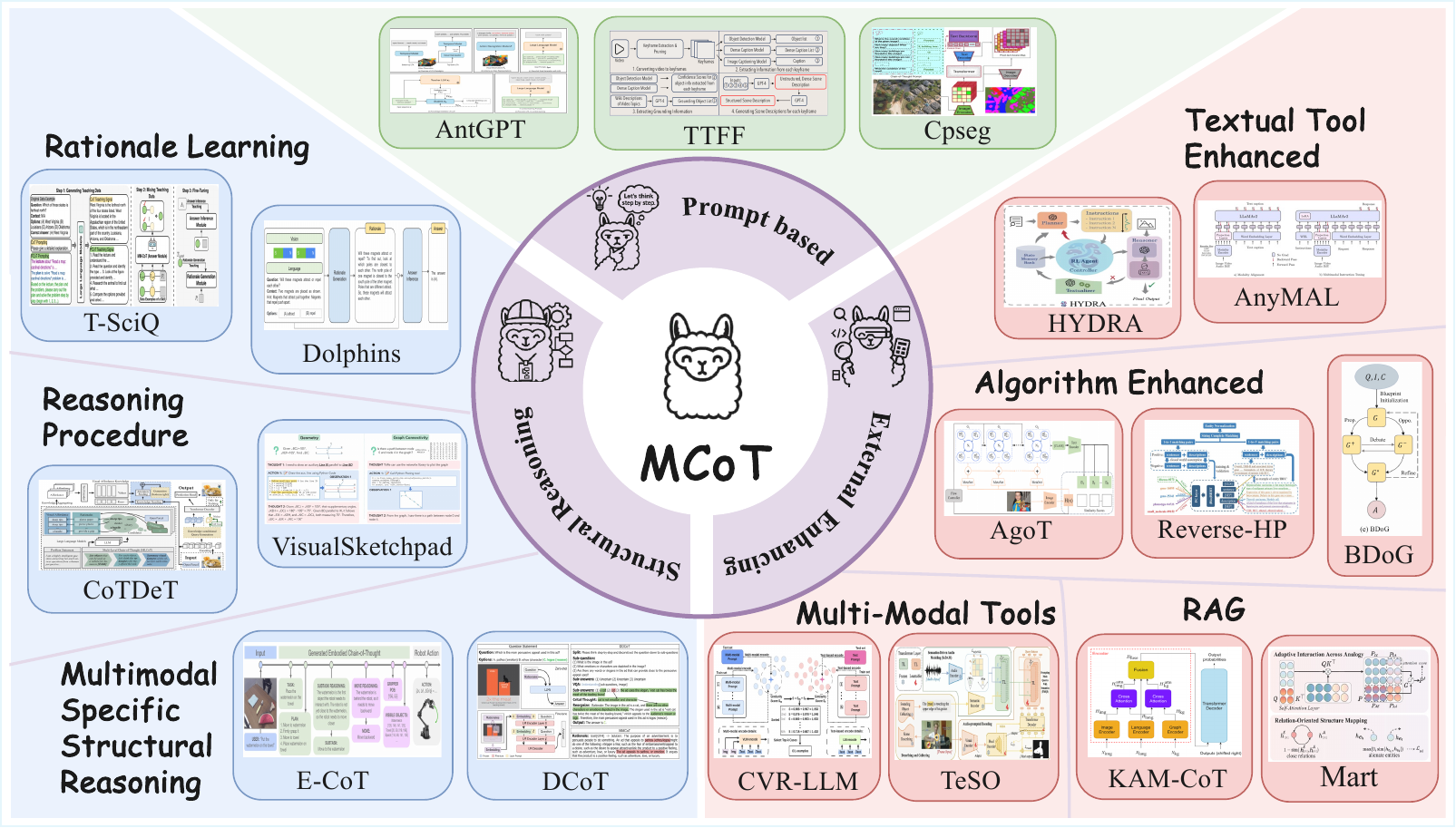}
    \vspace{-10pt}
    \caption{Taxonomy and representative methods of structural reasoning in multimodal chain-of-thought.}
    \label{fig:stage2_overview}
\end{figure}

\subsubsection{Prompt-based MCoT \label{sec:prompt} } \label{sec: stage2.1}

Prompt-based Multimodal Chain-of-Thought (MCoT) methods extend the textual CoT paradigm to multimodal contexts. These methods enable step-by-step reasoning across modalities, providing strong interpretability while requiring minimal additional training. In visual reasoning, IPVR~\citep{chen2023see} proposed a structured “see-think-confirm” prompting framework that guides LLMs through visual grounding and rationale verification. VIC~\citep{zheng2024thinking} prompts textual reasoning chains before visual input to mitigate hallucinations and improve accuracy.

For video understanding, VoT~\citep{fei2024video} leverages spatial-temporal scene graphs to prompt progressive reasoning from low-level perception to high-level interpretation. VideoAgent~\citep{wang2024videoagent} is an LLM-coordinated system that iteratively prompts key information from long videos with minimal frame usage. LET~\citep{himakunthala2023let} employs a frame-by-frame prompting strategy on the VIP dataset to guide temporal reasoning for video infilling and prediction.

In domain-specific applications, PKRD-CoT~\citep{luo2024pkrd} introduces a zero-shot prompting framework that structures autonomous driving reasoning across perception, knowledge, reasoning, and decision-making. LPE~\citep{xie2025leveraging} uses prompt-based reasoning on spoken content and emotional cues to generate empathetic responses. EMER~\citep{lian2023explainable} applies prompting in multimodal emotion recognition to integrate unimodal clues and produce interpretable predictions.

Task-oriented reasoning has also benefited from prompt-based MCoT. CoTDet~\citep{tang2023cotdet} uses multi-level prompting to extract affordance knowledge for object detection. AntGPT~\citep{zhao2023antgpt} prompts LLMs to infer human goals and temporal dynamics from video-based action sequences. CPSeg~\citep{li2024cpseg} formulates chain-of-thought prompts to align textual and pixel-level semantics for enhanced segmentation.

\subsubsection{Structural Reasoning} \label{sec: stage2.2}
\label{sec: structural_reasoning}
Unlike prompt-based MCoT methods, which induce reasoning behaviour through handcrafted exemplars or a zero-shot prompting approach, structural reasoning focuses on learning reasoning patterns through supervised training. By integrating explicit procedural structures, these methods convert loosely guided reasoning into standardized, stage-wise processes. This standardization improves scalability, reliability, and efficiency in complex multimodal tasks.
We categorize structural reasoning into three representative types: (i) \textit{rationale construction}, which learns to produce atomic reasoning steps as interpretable scaffolds; (ii) \textit{defined reasoning procedures}, which adapt structured texture reasoning schemes to multimodal settings; and (iii) \textit{modality-specific structural reasoning}, which further incorporates modality-aware constraints and designs to better align with the characteristics of visual, auditory, or embodied inputs.

\paragraph{Rationale Construction}

Effective rationale learning approaches form the foundation of structural reasoning in multimodal contexts. Recent studies continue to actively explore rationale generation from various perspectives. Multimodal-CoT~\citep{zhang2023multimodal} proposes a two-stage Multimodal-CoT framework that decouples rationale generation from answer prediction to reduce hallucinations. T-sciq~\citep{DBLP:conf/aaai/WangHHXLLS24} leverages teacher LLMs to generate rationales with varying complexity, showing rationale quality is key to reasoning accuracy. In autonomous driving, G-CoT~\citep{ma2024dolphins} designs Dolphins explicitly linking rationales to visual and historical driving signals for more grounded reasoning. MC-CoT~\citep{tan2024boosting} uses a self-consistency strategy to select the most accurate rationale among multiple candidates, boosting smaller models’ performance. CLoT~\citep{zhong2024let} promotes non-linear, explorative rationale construction via Leap-of-Thought to support creative reasoning.

\paragraph{Defined Reasoning Procedure}

In the realm of enhancing the interpretability of text reasoning processes, numerous studies have proposed structured reasoning stages. Cantor~\citep{gao2024cantor}, for instance, differentiates between perception and decision-making stages. In the perception stage, low-level attributes such as objects, colours, and shapes are extracted from images or textual descriptions, followed by the decision-making stage that integrates these features for problem-solving. TextCoT~\citep{luan2024textcot} adopts a three-phase process. The Image Overview stage generates a global description, the Coarse Localization stage pinpoints the answer region using the grounding ability of LMMs, and the Fine-grained Observation stage combines global and local details for accurate answers. Similarly, Grounding-Prompter~\citep{chen2023grounding} conducts global understanding, noise evaluation, partition understanding, and prediction. It gradually merges global and local semantics, resists noise, and improves the perception of temporal boundaries. Audio-CoT~\citep{ma2025audio_cot} utilizes three chain-of-thought reasoning paradigms. Manual-CoT depends on handcrafted examples for reasoning guidance, Zero-Shot-CoT achieves zero-shot reasoning with simple prompts, and Desp-CoT facilitates reasoning by generating audio descriptions. VIC~\citep{zheng2024thinking} breaks tasks into text-based sub-steps before integrating visual inputs to form final rationales. Visual Sketchpad~\citep{hu2024visual} organizes rationales into thought, action and observation phases during the sketching process. Det-CoT~\citep{wu2024dettoolchain} formalizes VQA reasoning as a combination of subtasks and reviews.  BDoG~\citep{zheng2024picture} utilizes a dedicated debate and summarization pipeline with unique agents. CoTDet~\citep{tang2023cotdet} achieves object detection via human-like procedure of listing, analyzing and summarizing. CoCoT~\citep{zhang2024cocot} systematically compares input similarities and differences. SegPref~\citep{wang2024can} localizes sounding objects accurately in the visual space through global understanding, sounding object filtering, and noise removal.  EMMAX~\citep{sun2024emmax} combines grounded planning approaches with predictive movement techniques.

\paragraph{Multimodal Specific Structural Reasoning}

Recent research has introduced modality-specific reasoning structures tailored to the unique challenges of multimodal inputs, particularly in vision-language tasks. A prominent line of work focuses on region-based grounding, where spatial localization is used to guide structured reasoning. For instance, CoS~\citep{liu2024chain} and TextCoT~\citep{luan2024textcot} adopt a two-stage pipeline that first identifies regions of interest conditioned on the input question, followed by localized inspection to enable multi-granular reasoning without resolution loss. DCoT~\citep{jia2024dcot} extends this paradigm by introducing a dual-guidance mechanism that combines bounding-box grounding with the retrieval of semantically similar examples, jointly enhancing fine-grained and context-aware reasoning. Beyond spatial grounding, CoT-PT~\citep{ge2023chain} integrates visual and textual embeddings through prompt tuning and gradually refines visual concept representations via coarse-to-fine abstraction.

Another class of approaches focuses on text-guided semantic enrichment. Shikra~\citep{chen2023shikra} and TextCoT~\citep{luan2024textcot} leverage image captions as high-level semantic cues to guide spatial attention and object grounding. This strategy reduces dependence on external detection modules and facilitates more interpretable referential reasoning. Inspired by classical CoT frameworks, DDCoT~\citep{zheng2023ddcot} and AVQA-CoT~\citep{li2024avqa_cot} decompose complex visual or audio-visual queries into sequential sub-questions, enabling compositional reasoning and improved multi-hop inference across modalities.

Finally, E-CoT~\citep{zawalski2024robotic} extends structured reasoning to embodied scenarios by interleaving task rephrasing, planning, and low-level action execution. This highlights the necessity of reasoning chains that span both semantic and sensorimotor levels in vision-language-action models.

\begin{table}[htbp]
\centering
\renewcommand{\arraystretch}{1.5}
\caption{The Structural Reasoning, which transforms loosely guided reasoning into standardized, step-by-step processes by integrating explicit procedural structures into the model, enhancing scalability, reliability, and efficiency in complex multimodal tasks.}
\tiny
\begin{tabular}{
  >{\raggedright\arraybackslash}p{1.2cm}|
  >{\raggedright\arraybackslash}p{0.6cm}|
  >{\raggedright\arraybackslash}p{1.0cm}|
  >{\raggedright\arraybackslash}p{2.6cm}|
  >{\raggedright\arraybackslash}p{1.6cm}|
  >{\raggedright\arraybackslash}p{5.6cm}
}
\toprule
\textbf{Name} & \textbf{Modality} & \textbf{Task} & \textbf{Reasoning Structure} & \textbf{Datasets} & \textbf{Highlight} \\
\midrule
Cantor~\citeyearpar{gao2024cantor}  & T,I & VQA & Perception, Decision & - & Decouples perception and reasoning via feature extraction and CoT-style integration.  \\ \hline
TextCoT\citeyearpar{luan2024textcot} & T,I & VQA & Caption, Localization, Precise observation & - & first summarizes visual context, then generates CoT-based responses. \\ \hline
Grounding-Prompter\citeyearpar{chen2023grounding} & T,V,A & Temporal Sentence Grounding & Denoising & VidChapters-7M & Grounding-Prompter performs global parsing, denoising, partitioning before reasoning.  \\ \hline
Audio-CoT\citeyearpar{ma2025audio_cot} & T,A & AQA & Manual-CoT, Zero-Shot-CoT, Desp-CoT & - & Enhances visual reasoning by utilizing three chain-of-thought paradigms.  \\ \hline
VIC\citeyearpar{zheng2024thinking} & I,T & VQA & Thinking before looking & - & Breaks tasks into text-based sub-steps before integrating visual inputs to form final rationales  \\ \hline
Visual Sketchpad\citeyearpar{hu2024visual} & I,T & VQA, math QA & Sketch-based reasoning paradigm & - & Organizes rationales into Thought-Action-Observation phases.  \\ \hline
Det-CoT\citeyearpar{wu2024dettoolchain} & I,T & VQA & Subtask decomposition, Execution, and Verification & - & Formalizes VQA reasoning as a combination of subtasks and reviews. \\ \hline
BDoG\citeyearpar{zheng2024picture} & I,T & VQA & Entity update, Relation update, Graph pruning & - & Utilizes a dedicated debate and summarization pipeline with unique agents.  \\ \hline
CoTDet\citeyearpar{tang2023cotdet} & I,T & object detection & Object listing, Affordance analysis, Visual feature summarization & COCO-Tasks & Achieves object detection via human-like procedure of listing, analyzing and summarizing.  \\ \hline
CoCoT\citeyearpar{zhang2024cocot} & I,T & VQA & Contrastive prompting strategy  & - & Systematically contrasts input similarities and differences.  \\ \hline
SegPref\citeyearpar{wang2024can} & T,A,V & Temporal Sentence Grounding & Visual summary, Sound filtering, Denoising & Youtube-8M, Semantic-ADE20K & Localizes sounding objects accurately in the visual space through global understanding, sounding object filtering, and noise removal.  \\ \hline
EMMAX\citeyearpar{sun2024emmax} & I,T & Robotic task & Grounded CoT reasoning, Look-ahead spatial reasoning & Dataset based on BridgeV2 & Combines grounded planning approaches with predictive movement techniques. \\ 
\midrule 
DDCoT \citeyearpar{zheng2023ddcot}       & T,I & VQA & Question Deconstruct,Rationale & ScienceQA & Maintains a critical attitude by identifying reasoning and recognition responsibilities through the combined effect of negative-space design and visual deconstruction.\\
\hline
AVQA-CoT \citeyearpar{li2024avqa_cot}   & T,A,V & AVQA & Question Deconstruct , Question Selection , Rationale & MUSIC-AVQA & Decomposes complex questions into multiple simpler sub-questions and leverages LLMs to select relevant sub-questions for audio-visual question answering. \\
\hline
CoT-PT \citeyearpar{ge2023chain}     & T,I & Image Classification, Image-Text Retrieval , VQA & Coarse-to-Fine Image Concept Representation & ImageNet  & First to successfully adapt CoT for prompt tuning by combining visual and textual embeddings in the vision domain.\\
\hline
IoT~\citeyearpar{zhou2024image}      & T,I & VQA & Visual Action Selection , Execution , Rationale , Summary , Self-Refine & - & Enhances visual reasoning by integrating visual and textual rationales through a model-driven multimodal reasoning chain.\\
\hline
Shikra ~\citeyearpar{chen2023shikra}      & T,I & VQA , PointQA  & Caption  , Object Grounding & ScienceQA & Maintains a critical attitude by identifying reasoning and recognition responsibilities through the combined effect of negative-space design and visual deconstruction.\\
\hline
E-CoT ~\citeyearpar{zawalski2024robotic}      & T,I,A &  Policies' Generalization & Task Rephrase , Planning , Task Deconstruct , Object Grounding  & Bidgedata v2 &Integrates semantic planning with low-level perceptual and motor reasoning, advancing task formulations in embodied intelligence.\\
\hline
CoS ~\citeyearpar{liu2024chain}      & T,I & VQA & Object Grounding , Rationale & Llava665K & Guides the model to identify and focus on key image regions relevant to a question, enabling multi-granularity understanding without compromising resolution.\\
\hline
TextCoT ~\citeyearpar{luan2024textcot}      & T,I & VQA & Caption , Object Grounding , Image Zoom & Llava665K , SharedGPT4V & Enables accurate and interpretable multimodal question answering through staged processing: overview, coarse localization, and fine-grained observation.\\
\hline
DCoT ~\citeyearpar{jia2024dcot}       & T,I & VQA & Object Grounding , Fine-Grained Image Generation , Similar Example Retrieve , Rationale & - & Uses a dual-guidance mechanism by combining bounding box cues to focus attention on relevant image regions and retrieving the most suitable examples from a curated demonstration cluster as contextual support.\\
\bottomrule
\end{tabular}
\end{table}

\begin{TakeawayBox}{Takeaways: Structural Reasoning}
Structural reasoning methods define standardized reasoning workflows by integrating modular sub-tasks such as question deconstruct, visual grounding, caption generation, summary, phases, and image procession. These approaches enhance interpretability and consistency by organizing generation task into explicit stages. Recent trends also incorporate modality-aware designs to better align reasoning with visual, auditory, or embodied inputs.
\end{TakeawayBox}

\subsubsection{Externally Augmented Reasoning} \label{sec: stage2.3}
\label{sec: stage_2_3}
Externally augmented reasoning introduces advanced algorithms, auxiliary tools, or expert modules to compensate for limitations in the model's inherent reasoning capacity. These components are integrated at inference time or coupled during training, enabling more flexible, scalable, and task-specialized reasoning workflows. By decoupling core reasoning steps from the base model, such methods support long-horizon planning, precise grounding, and access to dynamic or domain-specific information.
We group externally augmented methods into four categories: (i) \textit{search algorithm-enhanced MCoT}, which navigates reasoning spaces via various search algorithm; (ii) \textit{tool-based augmentation}, which leverages external language tools or systems to guide reasoning execution; (iii) \textit{retrieval-augmented reasoning}, which incorporates relevant multimodal knowledge from external sources into the reasoning path; and (iv) \textit{multimodal enhancing}, which incorporate specialized multimodal modules to support perception-driven reasoning.

\begin{table}[htbp]
\centering
\renewcommand{\arraystretch}{1.5}
\caption{Externally Augmented Reasoning, which enhances a model's reasoning by incorporating external resources like algorithms, tools, or expert modules to overcome its inherent limitations.}
\tiny
\begin{tabular}{
  >{\raggedright\arraybackslash\sloppy}p{1.8cm}|
  >{\raggedright\arraybackslash\sloppy}p{0.8cm}|
  >{\raggedright\arraybackslash\sloppy}p{1.8cm}|
  >{\raggedright\arraybackslash\sloppy}p{1.6cm}|
  >{\raggedright\arraybackslash\sloppy}p{1.8cm}|
  >{\raggedright\arraybackslash\sloppy}p{4.6cm}
}
\toprule
\textbf{Name} & \textbf{Modality} & \textbf{Task} & \textbf{Enhancement Type} & \textbf{External Source} & \textbf{Highlight} \\
\midrule
MM-ToT ~\citeyearpar{gomez2023mmtot} & T,I & Image Generation & Search Algorithm & DFS,BFS & Applies DFS and BFS to select optimal outputs. \\
\hline
HoT ~\citeyearpar{yao2023HoT} & T,I & VQA & Search Algorithm & multi-hop random walks on graph & Generates linked thoughts from multimodal data in a hyperedge. \\
\hline
AGoT ~\citeyearpar{yang2024AGoT} & T,I & Text-Image Retrieval, VQA & Search Algorithm &  prompt aggregation and prompt flow operations  & Builds a graph to aggregate multi-faceted reasoning with visuals. \\
\hline
BDoG ~\citeyearpar{zheng2024picture} & T,I & VQA & Search Algorithm & Graph Condensation: Entity update, Relation update, Graph pruning  & Effective three-agent debate forms thought graph for multimodal queries. \\ 
\midrule 
L3GO ~\citeyearpar{yamada2024Co3D_Thoughts} & T,I & 3D Object Generation \& Composition & Tools & Blender, ControlNet & Iterative part-based 3D construction through LLM reasoning in a simulation environment. \\
\hline
HYDRA ~\citeyearpar{ke2024hydra} & T,I & Knowledge-QA, Visual Grounding & Tools & RL agent controller, Visual Foundation Models & RL agent controls multi-stage visual reasoning through dynamic instruction selection. \\
\hline
Det-CoT ~\citeyearpar{wu2024dettoolchain} & T,I & object detection & Tools & Visual Processing Prompts & Visual prompts guide MLLM attention for structured detection reasoning. \\
\hline
Chain-of-Image ~\citeyearpar{meng2023chain} & T,I & Geometric, chess \& commonsense reasoning & Tools & Chain of Images prompting & Generates intermediate images during reasoning for visual pattern recognition. \\
\hline
AnyMAL ~\citeyearpar{moon2024anymal} & T, I, A, V & Cross-modal reasoning, multimodal QA & Tools & Pre-trained alignment module & Efficient integration of diverse modalities; strong reasoning via LLaMA-2 backend. \\
\hline
SE-CMRN ~\citeyearpar{zhang2021explicit} & T,I & Visual Commonsense Reasoning & Tools & Syntactic Graph Convolutional Network & Enhances language-guided visual reasoning via syntactic GCN in a dual-branch network. \\
\midrule 
RAGAR ~\citeyearpar{khaliq2024ragar} & T,I & Political Fact-Checking & RAG & DuckDuckGo \& SerpAPI & Integrates MLLMs with retrieval-augmented reasoning to verify facts using text and image evidence. \\
\hline
Chain-of-action ~\citeyearpar{pan2024chain} & T,I & Info retrieval & RAG & Google Search, ChromaDB & Decomposes questions into reasoning chains with configurable retrieval actions to resolve conflicts between knowledge sources. \\
\hline
KAM-CoT ~\citeyearpar{mondal2024kam} & T,I, KG & Educational science reasoning & RAG & ConceptNet knowledge graph & Enhances reasoning by retrieving structured knowledge from graphs and integrating it through two-stage training. \\
\hline
AR-MCTS ~\citeyearpar{dong2024progressive_retrieval} & T,I & Multi-step reasoning & RAG & Contriever, CLIP dual-stream & Step-wise retrieval with Monte Carlo Tree Search for verified reasoning. \\
\hline
MR-MKG ~\citeyearpar{lee2024multimodal} & T, I & General multimodal reasoning & RAG & RGAT & Enhances multimodal reasoning by integrating information from multimodal knowledge graphs. \\
\hline
 Reverse-HP ~\citeyearpar{zhu2022multimodal} & T, I & Disease-related reasoning & RAG & reverse hyperplane projection & Utilizes KG embeddings to enhance reasoning for specific diseases with multimodal data. \\
\hline
MarT ~\citeyearpar{zhang2022multimodal} & T, I & Analogical reasoning & RAG & Structure-guided relation transfer & Uses structure mapping theory and relation-oriented transfer for analogical reasoning with KG. \\
\midrule 
MCoT-Memory~\citeyearpar{liang2024memory_driven} & T,I & VQA & Multimodal Information Enhancing &  LLAVA & Memory framework and scene graph construction for effective long-horizon task planning  \\ 
\hline
MGCoT~\citeyearpar{yao2023beyond} & T,I & VQA & Multimodal Embedding Enhancing & ViT-large encoder 
& Precise visual feature extraction aiding multimodal reasoning  \\ 
\hline
CCoT~\citeyearpar{mitra2024compositional}  & T,I & VQA & Multimodal Perception Enhancing & Scene Graphs & Utilization of the generated scene graph as an intermediate reasoning step.  \\ 
\hline
CVR-LLM~\citeyearpar{li2024enhancing} & T,I & VQA & Multimodal Embedding Enhancing & BLIP2flant5 \& BLIP2 multi-embedding & Precise context-aware image descriptions through iterative self-refinement and effective text-multimodal factors integrations \\ 
\hline
CAT~\citeyearpar{wang2023caption}  & T,I & Image Captioning & Multimodal Perception Enhancing & SAM & Promising pre-trained image caption generators, SAM, and instruction-tuned large language models integration\\ 
\bottomrule
\end{tabular}
\end{table}

\paragraph{Search Algorithm Enhanced MCoT}
Search strategy-driven MCoT approaches empower models to navigate and optimise reasoning trajectories throughout the reasoning process dynamically. MM-ToT~\citep{gomez2023mmtot}, for instance, leverages GPT-4 and Stable Diffusion, employing depth-first search (DFS) and breadth-first search (BFS) algorithms to identify the most optimal multimodal outputs according to a 0.0–1.0 metric scale. HoT~\citep{yao2023HoT} creates interconnected thoughts from multimodal inputs and packages them into a single hyperedge. Unlike this, Aggregation Graph-of-Thought (AGoT)~\citep{yang2024AGoT} builds a reasoning aggregation graph, which integrates diverse reasoning elements at every step and subsequently incorporates visual data. Blueprint Debate on Graph (BDoG)~\citep{zheng2024picture} takes a distinctive route, discarding search algorithms and instead utilizing three agents—an affirmative debater, a negative debater, and a moderator. These agents engage in iterative debates to address multimodal questions, with the moderator ultimately synthesizing a final answer, thus implicitly constructing a graph-of-thought that explores and aggregates a wide range of thoughts. Overall, compared to prompt-based methods that rely on linear, example-driven inference, search strategy-oriented MCoT variants enable models to explore multiple reasoning pathways, thereby significantly enhancing adaptability and the depth of problem-solving.

\paragraph{Textural Tools}

To enhance the reasoning capabilities of multimodal Chain-of-Thought (MCoT) frameworks, some works incorporate external textual-enhancing tools that guide, structure, or refine the overall reasoning process through language. L3GO~\citep{yamada2024Co3D_Thoughts} employs GPT-4 with Chain-of-Thought prompting to produce explicit textual reasoning steps, which guide 3D mesh construction in a Blender environment, aided by ControlNet for visual grounding. HYDRA~\citep{ke2024hydra} and Det-CoT~\citep{wu2024dettoolchain} leverage large language models not only as planners, but also as dynamic instruction generators, error diagnosers, and reasoning controllers. These models interact with visual foundation models (e.g., BLIP2, LLaVA) and reinforcement learning agents, while using textual prompts and feedback to iteratively improve visual understanding and decision-making. Both systems integrate a State Memory Bank to maintain dialogue history or prior instructions, enabling incremental CoT reasoning via textual modulation. Chain-of-Image~\citep{meng2023chain} introduces SyMLLM, which generates intermediate images from language descriptions, turning complex problems into visual reasoning tasks—yet still grounded in language-based control. Similarly, AnyMAL~\citep{moon2024anymal} unifies diverse modalities into a textual space for cross-modal reasoning, while SE-CMRN~\citep{zhang2021explicit} utilizes syntactic cues via GCNs to improve performance in visual commonsense reasoning.


\paragraph{RAG}
Several approaches enhance multimodal reasoning through retrieval mechanisms, e.g., solving online questions~\citep{chen2024adaptive}. RAGAR~\citep{khaliq2024ragar} proposed CoRAG and ToRAG to support political fact-checking through retrieval of multimodal evidence. Chain-of-Action~\citep{pan2024chain} retrieves information from heterogeneous sources through configurable reasoning chains. KAM-CoT~\citep{mondal2024kam} incorporates Knowledge Graphs as external knowledge sources to augment multimodal reasoning. AR-MCTS~\citep{dong2024progressive_retrieval} integrates dynamic step-wise retrieval with Monte Carlo Tree Search, enabling MLLMs to access relevant knowledge at each reasoning step and automatically generate high-quality reasoning annotations. Knowledge graph integration has further expanded multimodal reasoning capabilities through diverse approaches: MR-MKG~\citep{lee2024multimodal} enhances general multimodal reasoning by retrieving relevant triples from MMKGs via RGAT, Reverse-HP~\citep{zhu2022multimodal} enables disease-related reasoning using reverse hyperplane projection on SDKG-11, and MarT~\citep{zhang2022multimodal} employs structure mapping theory for multimodal analogical reasoning through relation-oriented transfer between entities in MarKG.


\paragraph{Multimodal Tools}
Using visual experts is another effective way to enhance the capabilities of models for multimodal reasoning. MCoT-Memory~\citep{liang2024memory_driven} improves long-horizon planning by incorporating memory retrieval and scene graph updates, retaining high-confidence experiences for robust decision-making. MGCoT~\citep{yao2023beyond} uses the ViT-large encoder (for multimodal tasks) to extract visual features, the Stanford CoreNLP system for coreference resolution, and the OpenIE system to extract thought unit nodes, thus enabling efficient GoT reasoning. CCoT~\citep{mitra2024compositional} enhances the compositional visual understanding and multimodal reasoning capabilities of LMMs through two key steps: scene graph generation and response generation. It utilizes the generated scene graph as an intermediate reasoning step. CVR-LLM~\citep{li2024enhancing} includes two key components: CaID generates context-aware image descriptions through iterative self-refinement, and CVR-ICL innovatively integrates text and multimodal factors to select context examples, enhancing the performance of LLMs in complex visual reasoning tasks. CAT~\citep{wang2023caption} integrates pre-trained image caption generators, SAM, and instruction-tuned large language models. Through visual controls and language controls, it realizes user-centered image description. VISPROG ~\citep{gupta2023visual} iterates alternately through three steps: initial generation, feedback, and refinement. It utilizes a suitable language model and three prompts and based on few-shot prompting, guides the model to generate feedback and refine the output until the stopping condition is met.

\begin{TakeawayBox}{Takeaways: Externally Augmented Reasoning}
Externally augmented reasoning introduces auxiliary modules (such as search algorithms, tool agents, retrieval systems, and specialized multimodal processors) to assist or offload parts of the reasoning process. These methods enable more controllable, scalable, and task-adaptive reasoning by decoupling planning, grounding, or perception tasks from the backbone model, often enhancing long-horizon reasoning and domain specialization.
\end{TakeawayBox}

\subsection{Stage 3 Language-Centric Long Reasoning - System-2 Thinking and Planning}
\label{sec:stage3}

While structural reasoning introduces predefined patterns to guide MLLMs toward more systematic reasoning, it remains constrained by shallow reasoning depth and limited adaptability. To handle more complex multimodal tasks, recent work aims to develop System-2-style reasoning~\citep{kahneman2011thinking}. Unlike fast and reactive strategies, this form of reasoning is deliberate, compositional, and guided by explicit planning. By extending reasoning chains, grounding them in multimodal inputs, and training with supervised or reinforcement signals, these models begin to exhibit long-horizon reasoning and adaptive problem decomposition.

\begin{figure}[t]
    \centering
    \includegraphics[width=1\linewidth]{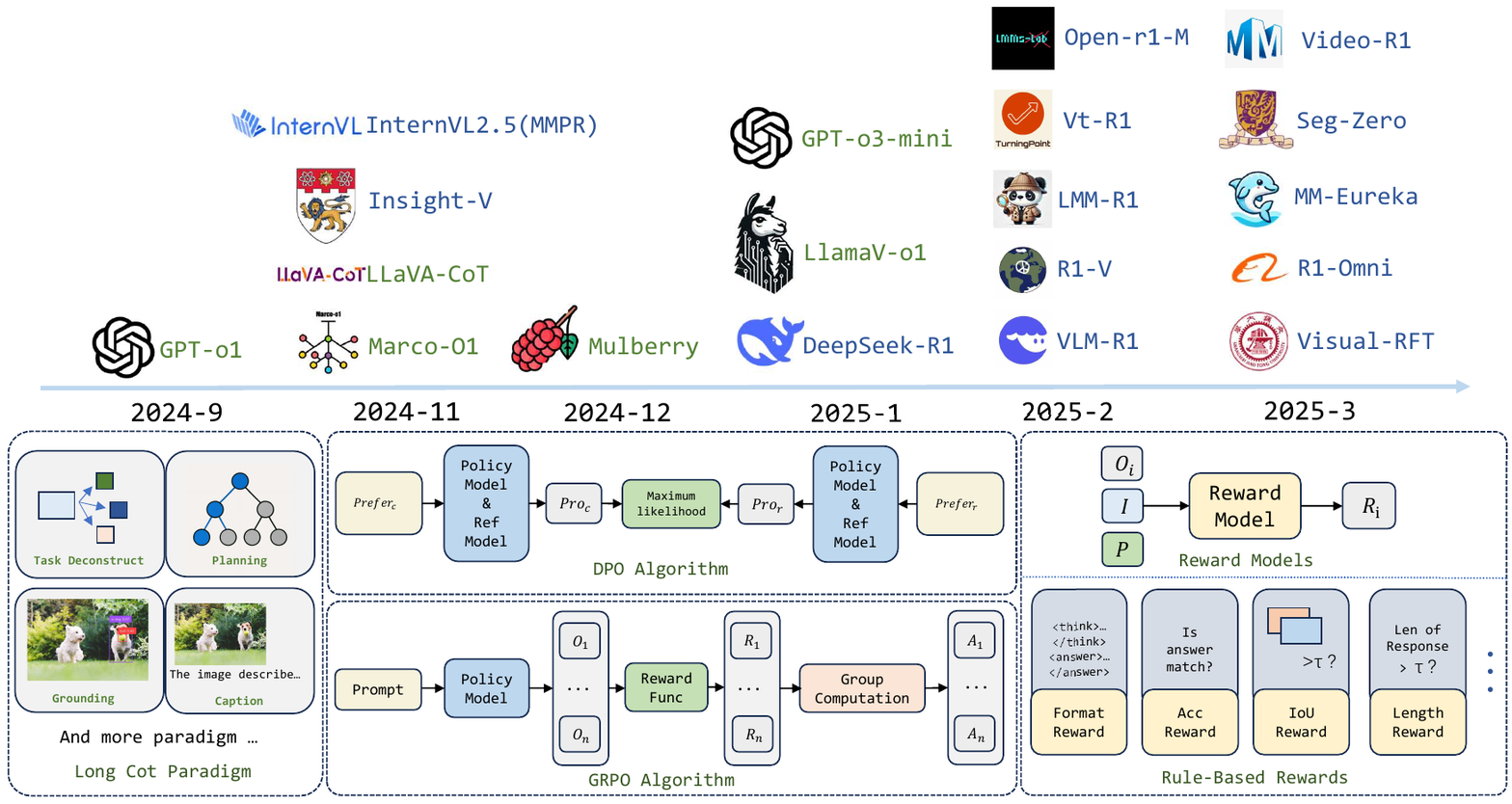}
    \vspace{-50pt}
    \caption{Timeline (top) and core components (bottom) of recent multimodal O1-like and R1-like models. The top part illustrates the chronological emergence of representative models. The bottom part summarizes key components including structured reasoning paradigms, reinforcement learning algorithms (e.g., DPO and GRPO), and the design of rule-based reward models.}
    \label{fig:stage3_overview}
\end{figure}

\begin{table}[htbp]
\centering
\renewcommand{\arraystretch}{1.5}
\caption{Approaches enhancing Cross-Modal Reasoning, which refers to the ability to integrate and reason across multiple modalities, such as text,
images, videos.}
\tiny
\begin{tabular}{
  >{\raggedright\arraybackslash}p{1.8cm}|
  >{\raggedright\arraybackslash}p{0.8cm}|
  >{\raggedright\arraybackslash}p{2.8cm}|
  >{\raggedright\arraybackslash}p{1.6cm}|
  >{\raggedright\arraybackslash}p{5.0cm}
}
\toprule
\textbf{Name} & \textbf{Modality} & \textbf{Cross-Modal Reasoning} & \textbf{Task} & \textbf{Highlight} \\
\midrule
IdealGPT~\citeyearpar{DBLP:conf/emnlp/YouSW0WACC23} & T, I & Answer sub-questions about image via gpt & VQA, Text Entailment & Using gpt to iteratively decompose and solve visual reasoning tasks \\
\hline
AssistGPT~\citeyearpar{gao2023assistgpt} & T, I, V & Plan, Execute, Inspect via External Tools (gpt4, OCR, Grounding, et al.) & VQA, Causal Reasoning  & Using an interleaved code and language reasoning approach to handle complex multimodal tasks \\
\hline
ProViQ~\citeyearpar{choudhury2023zero} & T, V & Generate and execute Python programs for the video & Video VQA & Using procedural programs to solve visual subtasks in videos \\
\hline
MM-REACT~\citeyearpar{yang2023mm} & T, I, V & Use CV tools for sub-taskss about image & VQA, Video VQA & Vision experts combined with GPT for multimodal reasoning and action \\
\hline
VisualReasoner~\citeyearpar{cheng2024least} & T, I & Synthesize multi-step reasoning (Using exteral CV tools) data & GQA, VQA & Proposing a least-to-most visual reasoning paradigm and a data synthesis approach for training\\
\hline
Multi-model-thought~\citeyearpar{lin2025investigating} & T, I & External Tools (Visual Sketchpad) & Geometry, Math, VQA & Investigating inference-time scaling for multi-modal thought across diverse tasks\\
\hline
FaST~\citeyearpar{sun2024visual} & T, I & System switch adapter for visual reasoning & VQA & Integrating fast and slow thinking mechanisms into visual agents \\
\hline
ICoT~\citeyearpar{gao2024interleaved} & T, I & Generate interleaved visual-textual reasoning via ADS & VQA & Using visual patches as reasoning carriers to improve LMMs' fine-grained reasoning \\
\hline
Image-of-Thought~\citeyearpar{zhou2024image} & T, I & Extract visual rationales step-by-step via IoT prompting & VQA & Using visual rationales to enhance LLMs' reasoning accuracy and interpretability \\
\hline
CoTDiffusion~\citeyearpar{ni2024generate} & T, I & External Algorithms & Robotics & Generating subgoal images before action to enhance reasoning in long-horizon robot manipulation tasks \\
\hline
T-SciQ~\citeyearpar{DBLP:conf/aaai/WangHHXLLS24} & T, I & Model-Intrinsic Capabilities & ScienceQA & Using LLM-generated reasoning signals to teach multimodal reasoning for complex science QA \\
\hline
Visual-CoT~\citeyearpar{rose2023visual} & T, I & Model-Intrinsic Capabilities & VQA, DocQA, ChartQA & Using visual-text pairs as reasoning carriers to bridge logical gaps in sequential data \\
\hline
VoCoT~\citeyearpar{li2024vocot} & T, I & Model-Intrinsic Capabilities & VQA & Using visually-grounded object-centric reasoning paths for multi-step reasoning \\
\hline
MVoT~\citeyearpar{li2025imagine} & T, I & Model-Intrinsic Capabilities & Spatial Reasoning & Using multimodal reasoning with image visualizations to enhance complex spatial reasoning in LMMs \\

\bottomrule
\end{tabular}
\end{table}

\subsubsection{Cross-Modal Reasoning} 
\label{sec: stage3.1}
Cross-modal reasoning refers to the ability to integrate and reason across multiple modalities, such as text, images, and videos. Recent advancements in cross-modal reasoning emphasize augmenting multimodal information beyond textual inputs. These methods utilize either model-intrinsic capabilities or external tools and algorithms to achieve this augmentation. These methods aim to enhance reasoning accuracy and robustness by dynamically incorporating complementary information from diverse modalities.


\paragraph{External Tools}
Beyond the use of external tools for multimodal understanding described in \S\ref{sec: stage2.3}, recent approaches increasingly explore tool integration to directly facilitate multimodal reasoning. VisProg~\citep{gupta2023visual} and ProViQ~\citep{choudhury2023zero} leverage program generation and procedural execution to enable cross-modal reasoning, dynamically generating executable code or logic paths to solve complex tasks such as video question answering, multi-step visual reasoning, and geometric problem-solving. 
In parallel, methods such as AssistGPT~\citep{gao2023assistgpt}, MM-ReAct~\citep{yang2023mm}, and Multi-Modal-Thought~\citep{lin2025investigating} adopt modular integration frameworks (e.g., PEIL, vision expert prompting) to coordinate tool use based on reasoning progression. These systems enable interpretable and adaptive reasoning by calling different tools dynamically during task execution.
VisualReasoner~\citep{cheng2024least} further introduces a data synthesis strategy to generate multi-step reasoning traces, which are then used to train a plug-and-play visual reasoning module applicable to a variety of vision-language backbones.
Collectively, these efforts extend the landscape of multimodal reasoning by combining program induction, dynamic tool orchestration, and data-driven reasoning supervision.

\paragraph{External Algorithms} 
Besides external tools, algorithmic methods have also been explored to enhance cross-modal reasoning through explicit cognitive process modelling. 
FAST~\citep{sun2024visual} and ICoT~\citep{gao2024interleaved} leverage cognitive processes akin to human thinking. Specifically, FAST employs a system switch adapter to dynamically alternate between fast and slow thinking modes, whereas ICoT utilizes Attention-driven Selection (ADS) to interleave visual and textual reasoning steps. Meanwhile, Image-of-Thought~\citep{zhou2024image} and CoTDiffusion~\citep{ni2024generate} focus on generating visual rationales. Image-of-Thought extracts visual information step-by-step, whereas CoTDiffusion creates visual subgoal plans, extending algorithmic augmentation into robotics domains.

\paragraph{Model-Intrinsic Capabilities} 
In contrast to approaches relying on external augmentation, several methods exploit intrinsic model capabilities to achieve cross-modal reasoning. These approaches rely on the inherent capabilities of LMMs to generate or infer multimodal information without relying on external tools.
T-SciQ~\citep{DBLP:conf/aaai/WangHHXLLS24}, Visual-CoT~\citep{rose2023visual} and VoCoT~\citep{li2024vocot} demonstrated how fine-tuning LMMs on carefully designed CoT datasets (e.g., VoCoT-Instruct80K) could enable single-step multimodal reasoning in charts, documents, and geometry problems. MVoT~\citep{li2025imagine} represents an early effort, where a self-contained architecture iteratively refines visual-textual representations for embodied reasoning tasks. 

\begin{TakeawayBox}{Takeaways: Cross-Modal Reasoning}
Cross-modal reasoning methods enhance multimodal inference by integrating visual, auditory, and programmatic cues across modalities. Representative strategies include leveraging external tools, algorithmic control for interleaving modality-specific steps, and model-intrinsic fusion of multimodal representations, enabling more grounded, interpretable, and robust reasoning in open-ended tasks.
\end{TakeawayBox}

\begin{table}[t]
\centering
\renewcommand{\arraystretch}{1.1}
\caption{Approaches of Multimodal-o1. It mainly relies on a multi-stage, structured reasoning path to solve problems.}
\tiny
\resizebox{\textwidth}{!}{
\begin{tabular}{
  >{\raggedright\arraybackslash}p{1.6cm}|
  >{\raggedright\arraybackslash}p{2.0cm}|
  >{\raggedright\arraybackslash}p{2.2cm}|
  >{\raggedright\arraybackslash}p{0.8cm}|
  >{\raggedright\arraybackslash}p{2.3cm}|
  >{\raggedright\arraybackslash}p{1.6cm}|
  >{\raggedright\arraybackslash}p{3.6cm}
}
\toprule
\textbf{Name} & \textbf{Backbone} & \textbf{Dataset} & \textbf{Modality} & \textbf{Reasoning Paradigm} & \textbf{Task Type} & \textbf{Highlight} \\
\midrule
Macro-O1~\citeyearpar{DBLP:journals/corr/abs-2411-14405} & Qwen2-7B-Instruct & Open-O1 CoT + Marco-o1 CoT + Marco-o1 Instruction & T & MCTS-guided Thinking & Math, Translate & MCTS for solution expansion and reasoning action strategy \\
\hline
llamaberry~\citeyearpar{zhang2024llamaberry} & LLaMA-3.1-8B & PRM800K + OpenMathInstruct-1 & T & MCTS-guided Thinking & Math & SR-MCTS for search and PPRM for evaluation \\
\hline
LLaVA-CoT~\citeyearpar{xu2024llava_cot} & Llama-3.2V-11B-cot & LLaVA-CoT-100k & T, I & Summary, Caption, Thinking & Science, General & Introduce LLaVA-CoT-100k and scalable beam search \\
\hline
LlamaV-o1~\citeyearpar{thawakar2025llamav_o1} & Llama-3.2V-11B-cot & LLaVA-CoT-100k + PixMo & T, I & Summary, Caption, Thinking & Science, General & Introduce VCR-Bench and outperforms \\
\hline
Mulberry~\citeyearpar{yao2024mulberry} & Llama-3.2V-11B-cot, LLaVA-Next-8B, Qwen2-VL-7B & Mulberry-260K & T, I & Caption, Rationales, Thinking & Math, General & Introduce Mulberry-260k and CoMCTS for collective learning \\
\hline
RedStar-Geo~\citeyearpar{xu2025redstar} & InternVL2-8B & GeoQA & T, I & Long-Thinking & Math & Competitive with minimal Long-CoT data \\
\hline
RBF++~\citeyearpar{chen2025rbfquantifyingoptimizingreasoning} & LLaMA3-8B-Instruct & GSM8K + SVAMP + MATH & T & SR-MCTS + PPRM & Math & Proposes SR-MCTS for structured search and PPRM for evaluating reasoning boundaries \\
\bottomrule
\end{tabular}}
\end{table}

\subsubsection{Multimodal-O1}
\label{sec: stage3.2}

With the rise of OpenAI o1, which sparked widespread interest in large reasoning models, open-source reproductions such as Marco-o1~\citep{DBLP:journals/corr/abs-2411-14405} and llamaberry~\citep{zhang2024llamaberry} utilizing CoT fine-tuning began to emerge. Compared to traditional CoT approaches, CoT fine-tuning enhances the model's reasoning capabilities on open-ended questions by incorporating mechanisms for self-reflection and error correction. LLaVA-CoT~\citep{DBLP:journals/corr/abs-2411-10440}, LlamaV-o1~\citep{thawakar2025llamav_o1},  RedStar~\citep{xu2025redstar} and Mulberry~\citep{yao2024mulberry} extend the reasoning paradigm to the multimodal domain. In contrast to the two-stage reasoning paradigm of 'Thinking -> Answer' in text domains, these works expand the reasoning process to a four-stage approach that includes Summary (Rationale), Caption, Thinking and Answer.

Building on CoT fine-tuning and testing-time scaling with various reasoning strategies is also an important method to enhance reasoning capabilities. Best-of-N sampling generates multiple responses for a given prompt, expanding the search space to identify better solutions. Beam Search, on the other hand, does not generate a complete response in one pass but instead selects the most promising intermediate outputs at each step using scoring. LLaVA-CoT~\citep{DBLP:journals/corr/abs-2411-10440} and LlamaV-o1~\citep{thawakar2025llamav_o1} apply this method to strengthen reasoning abilities. Monte Carlo Tree Search (MCTS) allows for parallel exploration of multiple solution paths, enabling more comprehensive exploration and higher-quality solutions compared to Beam Search. Marco-o1~\citep{DBLP:journals/corr/abs-2411-14405}, llamaberry~\citep{zhang2024llamaberry}, and Mulberry~\citep{yao2024mulberry} have successfully integrated this approach into the generation process of reasoning models.

\begin{TakeawayBox}{Takeaways: Multimodal-O1}
Multimodal-O1 models extend System-1 reasoning by deepening CoT workflows through multi-stage generation structures, long-horizon reasoning, and structured supervision. Enhanced by fine-tuning on rationale-rich data and supported by planning algorithms such as Beam Search or MCTS, these models achieve more coherent, interpretable, and scalable multimodal reasoning.
\end{TakeawayBox}

\begin{table}[htbp]
\centering
\renewcommand{\arraystretch}{1.1}
\caption{Approaches of Multimodal-R1. It mainly employs reinforcement learning approaches to improve the general reasoning capability of large multimodal models.}
\tiny
\begin{tabular}{
  >{\raggedright\arraybackslash}p{2.0cm}|
  >{\raggedright\arraybackslash}p{1.6cm}|
  >{\raggedright\arraybackslash}p{1.8cm}|
  >{\raggedright\arraybackslash}p{1.5cm}|
  >{\raggedright\arraybackslash}p{0.8cm}|
  >{\raggedright\arraybackslash}p{1.0cm}|
  >{\raggedright\arraybackslash}p{1.0cm}|
  >{\raggedright\arraybackslash}p{0.6cm} |
  >{\raggedright\arraybackslash}p{0.9cm}
}
\toprule
\textbf{Apporach} & \textbf{Backbone} & \textbf{Dataset} & \textbf{RL Algorithm} & \textbf{Modality} & \textbf{Task Type} & \textbf{RL Framework} & \textbf{Cold Start} & \textbf{Rule-base/RM} \\
\midrule
RLHF-V~\citeyearpar{yu2024rlhf} & LLaVA-13B & RLHF-V-Dataset (1.4k) & DPO & T, I & VQA & Muffin & - & (unknown) \\
\hline
InternVL2.5~\citeyearpar{DBLP:journals/corr/abs-2411-10442} & InternVL & MMPR (3M) & MPO (DPO) & T, I & VQA & - & - & (unknown) \\
\hline
Insight-V~\citeyearpar{dong2024insight} & LLaMA3-LLaVA-Next & - & DPO & T, I & VQA & trl & - & (unknown) \\
\hline
LLaVA-Reasoner-DPO~\citeyearpar{DBLP:journals/corr/abs-2410-16198} & LLaMA3-LLaVA-Next & ShareGPT4o-reasoning-dpo (6.6k) & DPO & T, I & VQA & trl & - & (unknown) \\
\hline
VLM-R1~\citeyearpar{shen2025vlmr1} & Qwen2.5-VL & coco , LISA , Refcoco & GRPO & T, I & Grounding ,Math , Open-Vocabulary Detection & trl & No & Rule-base \\
\hline
R1-V~\citeyearpar{chen2025r1v} & Qwen2-VL & CLEVR  , GEOQA & GRPO & T, I & Counting , Math & trl & No & Rule-base \\
\hline
MM-EUREKA~\citeyearpar{meng2025mmeureka} & InternVL2.5 & K12 , MMPR & RLOO & T, I & Math & OpenRLHF & Yes & Rule-base \\
\hline
MM-EUREKA-Qwen~\citeyearpar{meng2025mmeureka} & Qwen2.5-VL & K12 , MMPR & GRPO & T, I & Math & OpenRLHF & No & Rule-base \\
\hline
Video-R1~\citeyearpar{feng2025video} & Qwen2.5-VL & Video-R1 (260K) & GRPO & T, I, V & Video VQA & trl & Yes & Rule-base \\
\hline
LMM-R1~\citeyearpar{peng2025lmmr1} & Qwen2.5-VL & VerMulti & PPO & T, I & Math & OpenRLHF & No & RM \\
\hline
Vision-R1~\citeyearpar{huang2025vision} & Qwen2.5-VL & LLaVA-CoT , Mulberry & GRPO & T, I & Math & - & Yes & Rule-base \\
\hline
Visual-RFT~\citeyearpar{liu2025visual} & Qwen2-VL & coco , LISA , ... & GRPO & T, I & Detection , Classification & trl & No & Rule-base \\
\hline
R1-OneVision~\citeyearpar{yang2025r1onevisionadvancinggeneralizedmultimodal} & Qwen2.5-VL & R1-Onevision-Dataset & GRPO & T, I & Math , Science , General , Doc & - & Yes & Rule-base \\
\hline
Seg-Zero~\citeyearpar{liu2025segzero} & Qwen2.5-VL , SAM2 & RefCOCOg , ReasonSeg & GRPO & T, I & Grounding & verl & No & Rule-base \\
\hline
VisualThinker-R1-Zero~\citeyearpar{zhou2025VisualThinker-R1-Zero} & Qwen2-VL & SAT dataset & GRPO & T, I & Spatial Reasoning & trl & No & Rule-base \\
\hline
STAR-R1~\citeyearpar{li2025starr1spatialtransformationreasoning} & Qwen2.5-VL-7B & TRANCE (13.5k) & GRPO & T, I & Spatial Reasoning & vLLM & No & Rule-base \\
\hline
R1-Omni~\citeyearpar{zhao2025r1omniexplainableomnimultimodalemotion} & HumanOmni & MAFW , DFEW & GRPO & T, I, A, V & emotion recognition & trl & Yes & Rule-base \\
\hline
OThink-MR1~\citeyearpar{liu2025othink} & Qwen2.5-VL & CLEVR , GEOQA & GRPO & T, I & Counting , Math & - & No & Rule-base \\
\hline
Multimodal-Open-R1~\citeyearpar{multimodal-open-r1} & Qwen2-VL & multimodal-open-r1-8k-verified (based on Math360K and Geo170K) & GRPO & T,I & Math & trl & No & Rule-base \\
\hline
Reason-RFT \citeyearpar{tan2025reasonrftreinforcementfinetuningvisual} & Qwen2.5-VL & CLEVR-Math, Super-CLEVR, GeoMath, Geometry3K, TRANCE & GRPO & T, I & Counting, Structure Perception, Spatial Transformation & trl & No & Rule-base \\
\hline
VL-Rethinker \citeyearpar{wang2025vlrethinkerincentivizingselfreflectionvisionlanguage} & Qwen2.5-VL & MathVista, MathVerse, MathVision, MMMU-Pro, EMMA, MEGA & GRPO+SSR & T, I & Mathematical, Scientific, Real-world Reasoning & trl & No & Rule-base \\
\hline
Curr-ReFT~\citeyearpar{deng2025boostinggeneralizationreasoningvision} & Qwen2.5-VL & RefCOCOg , Math360K , Geo170K & GRPO & T,I & Detection , Classification , Math & Curr-RL & No & RM \\
\hline
Open-R1-Video~\citeyearpar{wang-2025-open-r1-video} & Qwen2-VL & open-r1-video-4k & GRPO & T, I, V & Video VQA & trl & No & Rule-base \\
\hline
VisRL~\citeyearpar{chen2025visrl} & Qwen2.5-VL & VisCoT & DPO & T,I & VQA & trl & Yes & RM \\
\hline
R1-VL~\citeyearpar{zhang2025r1vl} & Qwen2-VL & Mulberry-260k & StepGRPO & T,I & Math , ChartQA & not release & No & Rule-base \\
\hline
SARI~\citeyearpar{wen2025sari} & Qwen2-Audio-7B-Instruct & MMAU (32k) & GRPO & A, T & MCQA & not release & No & Rule-based\\
\hline
SpaceR~\citeyearpar{ouyang2025spacerreinforcingmllmsvideo} & Qwen2.5-VL-7B & SpaceR-151k& SG-RLVR(GRPO) & T,I,V & Video Spatial Reasoning & - & Yes & Rule-base\\
\bottomrule
\end{tabular}
\end{table}

\begin{table}[htbp]
\centering
\renewcommand{\arraystretch}{1.1}
\caption{Approaches of Multimodal-R1 on image understanding and generation. It mainly employs reinforcement learning approaches to improve the generative reasoning of large multimodal models.}
\tiny
\begin{tabular}{
  >{\raggedright\arraybackslash}p{2.0cm}|
  >{\raggedright\arraybackslash}p{1.6cm}|
  >{\raggedright\arraybackslash}p{1.8cm}|
  >{\raggedright\arraybackslash}p{1.5cm}|
  >{\raggedright\arraybackslash}p{0.8cm}|
  >{\raggedright\arraybackslash}p{1.0cm}|
  >{\raggedright\arraybackslash}p{1.0cm}|
  >{\raggedright\arraybackslash}p{0.6cm} |
  >{\raggedright\arraybackslash}p{0.9cm}
}
\toprule
\textbf{Apporach} & \textbf{Backbone} & \textbf{Dataset} & \textbf{RL Algorithm} & \textbf{Modality} & \textbf{Task Type} & \textbf{RL Framework} & \textbf{Cold Start} & \textbf{Rule-base/RM} \\
\midrule
Cold ViCrit~\citeyearpar{wang2025vicrit} & Qwen2.5-VL-3B/72B-Instruct & PixMo-Cap & GRPO & T, I & Hallucination Detection & not release & No & Rule-base \\
\hline
Vision Matters~\citeyearpar{li2025vision} & Qwen2.5-VL-Instruct & Geometry3K, TQA, GeoQA, Math8K, M3CoT & GRPO + DPO & T, I & Math & MS-Swift, EasyR1 & No & RM \\
\hline
ViGaL~\citeyearpar{xie2025play} & Qwen2.5-VL-7B-Instruct & Sampled from game: Snake(36K), Rotation(36K) & RLOOT & T, I & Visual Games & OpenRLHF & No & Rule-base \\
\hline
RAP~\citeyearpar{li2025truth} & Qwen2.5-VL-3B, Qwen2.5-VL-7B & MM-Eureka & GRPO, RLOOT & T, I & Data Selection & EasyR1 & No & Not metion \\
\hline
RACRO~\citeyearpar{gou2025perceptual} & Qwen2.5-VL(3B, 7B, 32B) & ViRL39K & CROT & T, I & change reasoner without re-alignment & verl & No & combine \\
\hline
Revisual-R1~\citeyearpar{chen2025advancing} & Qwen2.5-VL-7B-Instruct & GRAMMAR & GRPO & T, I & Math & EasyR1 & Yes & Rule-base \\
\hline
Rex-Thinker~\citeyearpar{jiang2025rexthinkergroundedobjectreferring} & Qwen2.5-VL-7B & HumanRef-CoT & GRPO & T, I & Object Referring (REC) & verl & Yes & RM \\
\hline
ControlThinker~\citeyearpar{han2025controlthinkerunveilinglatentsemantics} & ControlARC & COCOStuff, MultiGen-20M & GRPO & T, I & Image Editing & not release & Yes & RM \\
\hline
SynthRL~\citeyearpar{wu2025synthrlscalingvisualreasoning} & Qwen2.5-VL-7B-Instruct & MMK12, A-MMK12 & GRPO & T, I & Math & verl & No & RM \\
\hline
SRPO~\citeyearpar{wan2025srpoenhancingmultimodalllm} & Qwen-2.5-VL-7B, Qwen-2.5-VL-32B & Mulberry dataset (260K), MathV360K, LLaVA-CoT (100K) , ScienceQA , etc. & GRPO & T, I & Math & verl & Yes & RM \\
\hline
ReasonGen-R1~\citeyearpar{zhang2025reasongenr1cotautoregressiveimage} & Janus-Pro-7B & LAION-5B & GRPO & T, I & Text to Image Generation & verl & Yes & RM \\
\hline
MoDoMoDo~\citeyearpar{liang2025modomodo} & Qwen2-VL-2B-Instruct & COCO, LISA, GeoQAV, SAT, ScienceQA & GRPO & T, I & General Visual Reasoning & trl & No & RM \\
\hline
DINO-R1~\citeyearpar{pan2025dino} & MM-Grounding-DINO & Objects365 & GRPO & T, I & Object Detection & not release & Yes & RM \\
\hline
VisualSphinx~\citeyearpar{feng2025visualsphinx} & Qwen2.5-VL-7B & VISUALSPHINX & GRPO & T, I & visual logic puzzle, math & verl & No & Rule-base \\
\hline
PixelThink~\citeyearpar{wang2025pixelthinkefficientchainofpixelreasoning} & Qwen2.5-VL-7B, SAM2-Large & RefCOCOg & GRPO & T, I & Segmentation & verl & No & Rule-base \\
\hline
ViGoRL~\citeyearpar{sarch2025grounded} & Qwen2.5-VL-3B, Qwen2.5-VL-7B & SAT-2, OS-ATLAS, ICAL, Segment Anything & GRPO & T, I & spatial reasoning, web grounding, web action, etc. & verl & Yes & Rule-base \\
\hline
Jigsaw-R1~\citeyearpar{wang2025jigsaw} & Qwen2.5-VL-7B, Qwen2.5-VL-3B, Qwen2-VL-2B, InternVL2.5-2B & COCO, CV-Bench, MMVP, SAT, Super-CLEVR & GRPO & T, I & jigsaw puzzles & trl & No & Rule-base \\
\hline
UniRL~\citeyearpar{mao2025unirl} & Show-o, Janus & COCO, GPT4o-Generated & GRPO & T, I & Image Understanding \& Generation & not release & Yes & Rule-base \\
\hline
cadrille~\citeyearpar{kolodiazhnyi2025cadrille} & Qwen2-VL-2B & DeepCAD & DPO, GRPO & T, I & Computer-Aided Design & not release & Yes & Rule-base \\
\hline
MM-UPT~\citeyearpar{wei2025unsupervised} & Qwen2.5-VL-7B & Geo3K, GeoQA, MMR1 & GRPO & T, I & Math & verl & No & Rule-base \\
\hline
RL-with-Cold-Start~\citeyearpar{wei2025unsupervised} & Qwen2.5-VL-3B, Qwen2.5-VL-7B & Geometry3K, GeoQA, GeoQA-Plus, Geos, AI2D, etc. & GRPO & T, I & Multimodal Reasoning, Math & verl & Yes & Rule-base \\
\hline
VRAG-RL~\citeyearpar{wang2025vrag} & Qwen2.5-VL-3B, Qwen2.5-VL-7B & ViDoSeek, SlideVQA, MMLongBench & GRPO & T, I & Visually Rich Information Understanding & verl & Yes & RM + Rule-base \\
\hline
MLRM-Halu~\citeyearpar{liu2025thinkingseeingassessingamplified} & Qwen2.5-VL(3B,7B) & MMMU, MMVP, MMBench, MMStar, etc. & GRPO & T, I & reasoning, perception & not release & Yes & Rule-base \\
\hline
Active-O3~\citeyearpar{zhu2025activeo3empoweringmultimodallarge} & Qwen2.5-VL-7B & SODA, LVIS & GRPO & T, I & active perception & not release & Yes & RM \\
\hline
RLRF~\citeyearpar{rodriguez2025renderingawarereinforcementlearningvector} & Qwen2.5-VL(3B,72B), Qwen3-8B & SVG-Stack & GRPO & T, I & Inverse rendering & not release & Yes & RM \\
\hline
VisTA~\citeyearpar{huang2025visualtoolagentvistareinforcementlearning} & Qwen2.5-VL-7B & ChartQA, Geometry3K & GRPO & T, I & Visual Reasoning, Tool Selection & OpenR1 & Yes & RM+Rule-base \\
\bottomrule
\end{tabular}
\end{table}

\begin{table}[htbp]
\centering
\renewcommand{\arraystretch}{1.1}
\caption{Approaches of Multimodal-R1 on image-text reasoning and image generation. It mainly employs reinforcement learning approaches to improve the reasoning capability of large multimodal models.}
\tiny
\begin{tabular}{
  >{\raggedright\arraybackslash}p{2.0cm}|
  >{\raggedright\arraybackslash}p{1.6cm}|
  >{\raggedright\arraybackslash}p{1.8cm}|
  >{\raggedright\arraybackslash}p{1.5cm}|
  >{\raggedright\arraybackslash}p{0.8cm}|
  >{\raggedright\arraybackslash}p{1.0cm}|
  >{\raggedright\arraybackslash}p{1.0cm}|
  >{\raggedright\arraybackslash}p{0.6cm} |
  >{\raggedright\arraybackslash}p{0.9cm}
}
\toprule
\textbf{Apporach} & \textbf{Backbone} & \textbf{Dataset} & \textbf{RL Algorithm} & \textbf{Modality} & \textbf{Task Type} & \textbf{RL Framework} & \textbf{Cold Start} & \textbf{Rule-base/RM} \\
\midrule
SATORI-R1~\citeyearpar{shen2025satorir1incentivizingmultimodalreasoning} & Qwen2.5-VL-Instruct-3B & Text-Total, ICDAR2013, ICDAR2015, etc. & GRPO & T, I & task-critical regions, answer accuracy & not release & No & RM \\
\hline
URSA~\citeyearpar{luo2025ursaunderstandingverifyingchainofthought} & Qwen2.5 Math-Instruct, SAM-B+SigLIP-L & DualMath-1.1M & GRPO & T, I & data reasoning, reward hacking & URSA & No & RM \\
\hline
v1~\citeyearpar{chung2025dontlookoncemultimodal} & Qwen2-VL(7B,72B), Qwen2.5-VL(7B,72B) & v1g & NoT & I & retrieve regions & None & No & No \\
\hline
GRE Suite~\citeyearpar{wang2025gresuitegeolocalizationinference} & Qwen2.5VL(3B,7B,32B) & Im2GPS3k, GWS15k & GRPO & T, I & reasoning location & LLaMA-Factory & Yes & RM+Rule-base \\
\hline
V-Triune~\citeyearpar{ma2025rlallvisualtriple} & Qwen2.5-VL-7B/32B-Instruct & mm\_math, geometry3k, mmk12, PuzzleVQA, etc. & GRPO & T, I & intensive perception & verl & Yes & RM \\
\hline
RePrompt~\citeyearpar{wu2025repromptreasoningaugmentedrepromptingtexttoimage} & Qwen2.5 7B & GenEva & GRPO & T, I & image generation & trl & Yes & RM \\
\hline
GoT-R1~\citeyearpar{duan2025gotr1unleashingreasoningcapability} & Qwen2.5VL-7B & JourneyDB-GoT, FLUX-GoT & GRPO & T, I & semantic-spatial reasoning & not release & No & RM \\
\hline
SophiaVL-R1~\citeyearpar{fan2025sophiavlr1reinforcingmllmsreasoning} & Qwen2.5-VL-7B-Instruct & SophiaVL-R1-130k & GRPO & T, I & reasoning-specific, general understanding & VeRL & No & RM+Rule-base \\
\hline
R1-ShareVL~\citeyearpar{yao2025r1sharevlincentivizingreasoningcapability} & Qwen2.5-VL-7B/32B & MM-Eureka & GRPO & T, I & General Visual Reasoning & EasyR1 & No & Rule-base \\
\hline
VLM-R\textsuperscript{3}~\citeyearpar{jiang2025vlmr3regionrecognitionreasoning} & Qwen2.5-VL-7B & VLIR & GRPO & T, I & Region Recognition and Reasoning & DeepSpeed & Yes & Rule-base \\
\hline
TON~\citeyearpar{wang2025thinknotselectivereasoning} & Qwen-2.5-VL-Instruct-3B/7B & CLEVR, Super-CLEVR, GeoQA, AITZ & GRPO & T, I & counting, navigation, math reasoning & vLLM & Yes & Rule-base \\
\hline
Pixel Reasoner~\citeyearpar{su2025pixelreasonerincentivizingpixelspace} & Qwen2.5-VL-7B & SA1B, FineWeb and STARQA & GRPO & T, I & pixel-space reasoning & OpenRLHF & No & Rule-base \\
\hline
GRIT~\citeyearpar{fan2025gritteachingmllmsthink} & Qwen2.5-VL-3B, InternVL-3-2B & VSR, TallyQA, GQA, MME, MathVista, etc. & GRPO & T, I & visual grounding, multi-step reasoning & Deepspeed Zero2 & No & Rule-base+RM \\
\hline
VARD~\citeyearpar{dai2025vardefficientdensefinetuning} & SD 1.4 & SCOPe, Pick-a-Pic, ImageRewardDB & DDPO w/o KL & I & image generation & not release & No & RM \\
\hline
Chain-of-Focus~\citeyearpar{zhang2025chainoffocusadaptivevisualsearch} & Qwen2.5-VL-7B & MM-CoF, SA\_1B, TextVQA, m3cot, V\*, POPE & GRPO & T, I & visual search and reasoning & not release & Yes & Rule-base \\
\hline
Visionary-R1~\citeyearpar{xia2025visionaryr1mitigatingshortcutsvisual} & Qwen2.5-VL-3B & A-OKVQA, ChartQA, AI2D, ScienceQA, etc. & GRPO & T, I & VQA & not release & No & Rule-base \\
\hline
VisualQuality-R1~\citeyearpar{wu2025visualqualityr1reasoninginducedimagequality} & Qwen2.5-VL-7B & KADID-10K, SPAQ & GRPO & T, I & image quality scoring & not release & No & Rule-base \\
\hline
DeepEyes~\citeyearpar{zheng2025deepeyesincentivizingthinkingimages} & Qwen2.5-VL-7B & V* training set, ArxivQA, ThinkLite-VL & GRPO & T, I & Multimodal Reasoning & verl & No & Rule-base \\
\hline
Visual-ARFT~\citeyearpar{liu2025visualagenticreinforcementfinetuning} & Qwen2.5-VL(3B,7B) & MAT-Search, MAT-Coding, 2WikiMultihopQA, etc. & GRPO & T, I & Multimodal Agentic Reasoning & not release & No & Rule-base \\
\hline
UniVG-R1~\citeyearpar{bai2025univgr1reasoningguideduniversal} & Qwen2-VL-2B 7B & MGrounding-630k, RefCOCO/+/g, etc. & GRPO & T, V, I & Visual Grounding (Multi-image, Complex) & Open-R1 & Yes & RM+Rule-base \\
\hline
G1~\citeyearpar{chen2025g1bootstrappingperceptionreasoning} & Qwen2.5-VL-7B & 128 parallel games, 500 training steps & GRPO & T, I & Interactive Game Decision-Making & EasyR1 & Yes & Rule-base \\
\hline
VisionReasoner~\citeyearpar{liu2025visionreasonerunifiedvisualperception} & Qwen2-VL & COCO, RefCOCO(+/g), ReasonSeg, etc. & GRPO & T, I & detection, segmentation, counting & not release & No & Rule-base \\
\hline
VPRL~\citeyearpar{xu2025visualplanningletsthink} & LVM-3B & FrozenLake, Maze, MiniBehavior & GRPO & I, V & Visual Spatial Planning & not release & Yes & Rule-base \\
\hline
GuardReasoner-VL~\citeyearpar{liu2025guardreasonervlsafeguardingvlmsreinforced} & Qwen2.5-VL Instruct 3B/7B & GuardReasoner-VLTrain & GRPO & T, I & Moderation (Harmfulness Detection) & EasyR1 & Yes & Rule-base \\

\bottomrule
\end{tabular}
\end{table}

\begin{table}[htbp]
\centering
\renewcommand{\arraystretch}{1.1}
\caption{Approaches of Multimodal-R1 on mathematic, code, and multimodal grounding. It mainly employs reinforcement learning approaches to improve the reasoning capability of large multimodal models.}
\tiny
\begin{tabular}{
  >{\raggedright\arraybackslash}p{2.0cm}|
  >{\raggedright\arraybackslash}p{1.6cm}|
  >{\raggedright\arraybackslash}p{1.8cm}|
  >{\raggedright\arraybackslash}p{1.5cm}|
  >{\raggedright\arraybackslash}p{0.8cm}|
  >{\raggedright\arraybackslash}p{1.0cm}|
  >{\raggedright\arraybackslash}p{1.0cm}|
  >{\raggedright\arraybackslash}p{0.6cm}|
  >{\raggedright\arraybackslash}p{0.9cm}
}
\toprule
\textbf{Apporach} & \textbf{Backbone} & \textbf{Dataset} & \textbf{RL Algorithm} & \textbf{Modality} & \textbf{Task Type} & \textbf{RL Framework} & \textbf{Cold Start} & \textbf{Rule-base/RM} \\
\midrule
OpenThinkIMG~\citeyearpar{su2025openthinkimglearningthinkimages} & Qwen2-VL-2B-Instruct & CHARTGEMMA & GRPO & T, I & Chart Reasoning & Open-R1 & Yes & Rule-base \\
\hline
DanceGRPO~\citeyearpar{xue2025dancegrpounleashinggrpovisual} & Stable Diffusion, HunyuanVideo, FLUX, etc. & curated prompt dataset, VidProM & GRPO & T, I & T2V, I2V, T2I Generation & fastvideo & No & RM \\
\hline
Flow-GRPO~\citeyearpar{liu2025flow} & SD3.5-M & GenEval, OCR, from pickscore & GRPO & T, I & Composition Image Gen, Text Rendering, etc. & no release & No & RM, Rule-base \\
\hline
X-Reasoner~\citeyearpar{liu2025xreasonergeneralizablereasoningmodalities} & Qwen2.5-VL-7B-Instruct & OpenThoughts, Orz-math, MedQA & GRPO & T, I & Generalization across domains & no release & No & Rule-base \\
\hline
T2I-R1~\citeyearpar{jiang2025t2ir1reinforcingimagegeneration} & Janus-Pro-7B & T2I-CompBench & GRPO & T, I & Text-to-Image Generation & Open-R1 & No & RM \\
\hline
RLHF-V~\citeyearpar{yu2024rlhf} & LLaVA-13B & RLHF-V-Dataset (1.4k) & DPO & T,  I & VQA & Muffin & - & (unknown) \\
\hline
RM-R1 ~\citeyearpar{chen2025rmr1rewardmodelingreasoning} & Qwen-Instruct (7B/14B/32B),  DeepSeek-Distilled-Qwen (7B/14B/32B)  & Skywork-Reward-Preference,  Code-Preference-Pairs,  Math-DPO-10K  & GRPO & T & Reward Modeling & verl & Yes & RM \\
\hline
AutoThink~\citeyearpar{tu2025learningthinkshapingadaptive} & DeepSeek-R1-Distill-Qwen-1.5B & MATH,  Minerva,  Olympiad,  AIME24,  AMC23 & GRPO & T & Mathematical Reasoning & verl & No & RM \\
\hline
WEBAGENT-R1~\citeyearpar{wei2025webagentr1trainingwebagents} & Qwen2.5-3B/Llama3.1-8B & WebArena-Lite & M-GRPO & T & web tasks & no release & Yes & RM \\
\hline
R1-Code-Interpreter~\citeyearpar{chen2025r1codeinterpretertrainingllmsreason} & Qwen-2.5-(3B, 7B, 14B) & SymBench, BIG-Bench-Hard, Reasoning-Gym & GRPO & T & planning  & verl & Yes & RM \\
\hline
Mixed-R1~\citeyearpar{xu2025mixedr1unifiedrewardperspective} & Qwen2.5-VL-(3B, 7B) & Mixed-45K & GRPO & T,  I,  V & reasoning & no release & Yes & RM + Rule-base \\
\hline
Router-R1 ~\citeyearpar{zhang2025routerr1teachingllmsmultiround} & Qwen2.5-3B-Instruct ,  LLaMA-3.2-3B-Instruct & Natural Questions,  TriviaQA,   PopQA; HotpotQA,  2WikiMultiHopQA,  Musique,  Bamboogle & PPO & T & Multi-hop Question Answering & verl & Yes & RM + Rule-base \\
\hline
FinLMM-R1~\citeyearpar{lan2025finlmmr1enhancingfinancialreasoning} & Qwen2.5-VL-3B  & FinData & GRPO & T, I & Reasoning & TAR-LMM  & No & RM \\
\hline
GVM-RAFT~\citeyearpar{yao2025optimizingchainofthoughtreasonersgradient} & Qwen2.5-Math-1.5B and Qwen2.5-Math-7B & Numina-Math & Dynamic RAFT & T & Math & verl & No & Rule-base \\
\hline
LoVeC~\citeyearpar{zhang2025reinforcementlearningbetterverbalized} & Llama-3-8B-Instruct and Gemma-2-9B-It & WildHallu, Bios, PopQA & GRPO, DPO, and ORPO & T & long-form generation & TRL/vLLM & No & Rule-base+RM \\
\hline
Critique-GRPO~\citeyearpar{zhang2025critiquegrpoadvancingllmreasoning} & Qwen2.5-7B-Base and Qwen3-8B-Base & OpenR1-Math-220k & GRPO & T & mathematical,  STEM,  and general reasoning & verl & Yes & RM \\
\hline
ReCode~\citeyearpar{wu2025recodeupdatingcodeapi} & Qwen-2.5-Coder-7B-Instruct and DeepSeekv1.5-Coder-7B-Instruct & construct own training dataset & GRPO, DAPO & T & code generation & not release & No  & Rule-base \\
\hline
ShapeLLM-omni~\citeyearpar{ye2025shapellm} & Qwen-2.5-VL-Instruct-7B & 3D-Alpaca & Not explicitly mentioned (Uses autoregressive models) & T, I, 3D & 3D Generation,  3D Understanding,  3D Editing & not release & No  & Rule-based \\
\hline
Omni-Perception~\citeyearpar{wang2025omniperceptionomnidirectionalcollisionavoidance} & PD-RiskNet &  Simulated with LiDAR simulation toolkit & PPO  & LiDAR + Proprioception & Locomotion + Collision Avoidance & not release & No & RM \\
\hline
Omni TM-AE~\citeyearpar{kadhim2025omnitmaescalableinterpretable} & TM-AE & One Billion Word Benchmark, IMDb, News20 & Tsetlin Automaton Logic & T & Similarity,  Classification, Clustering & no release & No  & RM \\
\hline
DeepVideo-R1~\citeyearpar{park2025deepvideor1videoreinforcementfinetuning} & Qwen2.5-VL-Instruct & SEED-Bench-R1-Train + NExTGQA  & GRPO & T,  I,  V & Video VQA  & not release & No & Rule-base \\
\hline
ReFoCUS~\citeyearpar{lee2025refocusreinforcementguidedframeoptimization} & LLaVA-OV / InternVL & ReFoCUS-962K & GRPO & T,  I,  V & Video VQA  & not release & No & RM \\

\bottomrule
\end{tabular}
\end{table}

\begin{table}[htbp]
\centering
\renewcommand{\arraystretch}{1.1}
\caption{Approaches of Multimodal-R1 on video, medical and audio. It mainly employs reinforcement learning approaches to improve the reasoning capability of large multimodal models on a specific modality.}
\tiny
\begin{tabular}{
  >{\raggedright\arraybackslash}p{2.0cm}|
  >{\raggedright\arraybackslash}p{1.6cm}|
  >{\raggedright\arraybackslash}p{1.8cm}|
  >{\raggedright\arraybackslash}p{1.5cm}|
  >{\raggedright\arraybackslash}p{0.8cm}|
  >{\raggedright\arraybackslash}p{1.0cm}|
  >{\raggedright\arraybackslash}p{1.0cm}|
  >{\raggedright\arraybackslash}p{0.6cm}|
  >{\raggedright\arraybackslash}p{0.9cm}
}
\toprule
\textbf{Apporach} & \textbf{Backbone} & \textbf{Dataset} & \textbf{RL Algorithm} & \textbf{Modality} & \textbf{Task Type} & \textbf{RL Framework} & \textbf{Cold Start} & \textbf{Rule-base/RM} \\
\midrule
TW-GRPO~\citeyearpar{dang2025reinforcingvideoreasoningfocused} & Qwen2.5-VL-Instruct & CLEVRER dataset & GRPO & T,  I,  V & Video VQA  & trl & No & Rule-base \\
\hline
Spatial-MLLM~\citeyearpar{wu2025spatialmllmboostingmllmcapabilities} & Qwen2.5-VL-Instruct & Spatial-MLLM-120k & GRPO & T,  I,  V & Spatial & not release & Yes & Rule-base \\
\hline
VAU-R1~\citeyearpar{zhu2025vaur1advancingvideoanomaly} & Qwen2.5-VL-Instruct & VAU-Bench-Train & GRPO & T,  I,  V & Video Grounding & trl & No & Rule-base \\
\hline
MUSEG~\citeyearpar{luo2025musegreinforcingvideotemporal} & Qwen2.5-VL-Instruct & E.T. Instruct 164k + CharadesSTA & GRPO & T,  I,  V & Video VQA + Video Grounding & trl & No & Rule-base \\
\hline
VerIPO~\citeyearpar{li2025veripocultivatinglongreasoning} & Qwen2.5-VL-Instruct & DAPO-Math + ViRL39K + VQA-Video-24K & GRPO + DPO & T,  I,  V & Video VQA + Spatial & OpenRLHF & No & Rule-base \\
\hline
SpaceR~\citeyearpar{ouyang2025spacerreinforcingmllmsvideo} & Qwen2.5-VL-Instruct & SpaceR-151k & GRPO & T,  I,  V & Spatial + VideoVQA & trl & No & Rule-base \\
\hline
TinyLLaVA-Video-R1~\citeyearpar{zhang2025tinyllavavideor1smallerlmmsvideo} & Qwen2.5-VL-Instruct & NextQA & GRPO & T,  I,  V & Video VQA & trl & Yes & Rule-base \\
\hline
VideoChat-R1~\citeyearpar{li2025videochatr1enhancingspatiotemporalperception} & Qwen2.5-VL-Instruct & Charade - STA + NExTGQA + FIBER-1k + VidTAB & GRPO & T,  I,  V & Video Grounding + Video VQA & trl & No & Rule-base \\
\hline
R1-Zero-VSI~\citeyearpar{liao2025improvedvisualspatialreasoningr1zerolike} & Qwen2-VL-Instruct & VSI-100k & GRPO & T,  I,  V & Spatial & not release & No & Rule-base \\
\hline
GRPO-CARE~\citeyearpar{chen2025exploringeffectreinforcementlearning} & Qwen2.5-VL-Instruct & SEED-Bench-R1-Train & GRPO & T,  I,  V & Video VQA + Spatial & trl & No & Rule-base \\
\hline
Time-R1~\citeyearpar{wang2025timer1posttraininglargevision} & Qwen2.5-VL-Instruct & Temporal-Spatial Video Datasets & GRPO & T,  I,  V & Video Grounding & trl & Yes & Rule-base \\
\hline
Lingshu~\citeyearpar{xu2025lingshu} & Qwen2.5-VL-Instruct & 3.75M open-source medical samples and 1.30M
synthetic medical samples / MedEvalKit & GRPO & T,  I & multimodal QA, medical report generation & not release & Yes & Rule-base \\
\hline
Patho-R1~\citeyearpar{zhang2025patho} & OpenAI-CLIP/Qwen2.5VL & PubMed, Quilt, PathGen & GRPO+DAPO & T,  I & Open-ended/Close-ended VQA & verl & Yes & Rule-base \\
\hline
ChestX-Reasoner~\citeyearpar{fan2025chestx} & Qwen2VL-7B & Medical Datasets & GRPO & T,  I & single/binary disease diagnosis & verl & Yes & Rule-base \\
\hline
Med-R1~\citeyearpar{lai2025med} & Qwen2-VL-2B-Instruct & OmniMedVQA & GRPO & T,  I & medical VQA & not release & Yes & Rule-base \\
\hline
MedVLM-R1~\citeyearpar{pan2025medvlm} & Qwen2-VL-2B &  HuatuoGPT-Vision  & GRPO & T,  I & Radiological VQA & not release & Yes & Rule-base \\
\hline
SARI~\citeyearpar{wen2025sari} & Qwen2-Audio-7B-Instruct/ Qwen2.5-Omni & AudioSet, MusicBench, Meld, AVQA & GRPO & T, A & Audio QA & trl & No & Rule-base \\
\hline
R1-AQA~\citeyearpar{li2025reinforcement} & Qwen2-Audio-7B-Instruct &  AVQA & GRPO & T, A & Audio QA & trl & Yes & Rule-base \\
\hline
Audio-Reasoner~\citeyearpar{xie2025audio} & Qwen2-Audio-7B-Instruct &  AVQA & GRPO & T, A & Audio QA & not release & Yes & Rule-base \\
\hline
Omni-R1~\citeyearpar{rouditchenko2025omni} & Qwen2.5-Omni & AVQA & GRPO & T, A & Audio QA & not release & No & Rule-base \\
\hline
WavReward~\citeyearpar{ji2025wavreward} & Qwen2.5-Omni-7B-Think & ChatReward-30K & PPO & T, A & end-to-end dialogue  & not release & No & Rule-base \\
\hline
SoundMind~\citeyearpar{diao2025soundmind} & Qwen2.5-Omni-7B & Audio Logical Reasoning (ALR) & REINFORCE++ & T, A & Audio text bimodal reasoning & verl & No & Rule-base \\
\hline
AudSemThinker~\citeyearpar{wijngaard2025audsemthinker} & Qwen2.5-Omni-7B & AUDSEM & GRPO & T, A & semantic audio reasoning & trl & No & Rule-base \\
\hline
AV-Reasoner~\citeyearpar{lu2025avreasonerimprovingbenchmarkingcluegrounded} & Ola-Omni7B & AVQA, Music AVQA, AVE, UnAV, LLP, AVSS-ARIG, DVD-Counting, RepCount & GRPO & T, I, V, A & Counting, Video VQA, Reasoning & trl & Yes & Rule-base \\
\hline
Omni-R1 (ZJU)~\citeyearpar{zhong2025omnir1reinforcementlearningomnimodal} & Qwen2.5-Omni-7B & RefAVS, ReVOS, MeViS, refCOCOg & GRPO & T, V, A & Multimodal Segmentation (AVS, VOS) & trl & Yes & Rule-base \\
\hline
Omni-R1 (MIT)~\citeyearpar{rouditchenko2025omni} & Qwen2.5-Omni-7B & AVQA-GPT, VGGS-GPT & GRPO & T, A & Audio QA & not release & no & RM \\
\hline
EchoInk-R1~\citeyearpar{xing2025echoinkr1exploringaudiovisualreasoning} & Qwen2.5-Omni-7B & AVQA-R1-6K & GRPO & T, I, A & Audio VQA & trl & no & Rule-base \\
\hline
Skywork-VL Reward~\citeyearpar{wang2025skyworkvlrewardeffectivereward} & Qwen2.5-VL-7B-Instruct & LLaVA-Critic-113k, Skywork-Reward-Preference-80Kv0.2, RLAIF-V-Dataset & MPO & T, I & VQA, Math, Science, Reasoning & not release & no & Rule-base \\

\bottomrule
\end{tabular}
\end{table}

\begin{table}[htbp]
\centering
\renewcommand{\arraystretch}{1.1}
\caption{Approaches of Multimodal-R1 on GUI (agentic) and reward models. It mainly employs reinforcement learning approaches to improve the interactive reasoning capability of large multimodal models.}
\tiny
\begin{tabular}{
  >{\raggedright\arraybackslash}p{2.0cm}|
  >{\raggedright\arraybackslash}p{1.6cm}|
  >{\raggedright\arraybackslash}p{1.8cm}|
  >{\raggedright\arraybackslash}p{1.5cm}|
  >{\raggedright\arraybackslash}p{0.8cm}|
  >{\raggedright\arraybackslash}p{1.0cm}|
  >{\raggedright\arraybackslash}p{1.0cm}|
  >{\raggedright\arraybackslash}p{0.6cm}|
  >{\raggedright\arraybackslash}p{0.9cm}
}
\toprule
\textbf{Apporach} & \textbf{Backbone} & \textbf{Dataset} & \textbf{RL Algorithm} & \textbf{Modality} & \textbf{Task Type} & \textbf{RL Framework} & \textbf{Cold Start} & \textbf{Rule-base/RM} \\
\midrule
UnifiedReward-Think~\citeyearpar{wang2025unifiedmultimodalchainofthoughtreward} & UnifiedReward & HPD(25.6K), EvalMuse(3K), OpenAI-4o\_t2i\_human\_preference (6.7K), VideoDPO (10K), Text2Video Human Preference (5.7K), ShareGPTVideo-DPO (17K) & GRPO & T, I, V & Video/Image Understanding, Reward Assessment & trl & yes & Rule-base \\
\hline
R1-Reward~\citeyearpar{zhang2025r1rewardtrainingmultimodalreward} & QwenVL-2.5-7B-Instruct & RLAIF-V, VL-Feedback, POVID, WildVision-Battle & StableReinforce (Reinforce++ variant) & T, I, V & Video/Image Understanding, Reward Assessment & OpenRLHF & yes & Rule-base \\
\hline
ComfyUI-R1~\citeyearpar{xu2025comfyui} & Qwen2.5-Coder-7B-Instruct & no release & GRPO & T, I, V & workflow generation & verl & yes & Rule-base \\
\hline
GUI-Critic-R1~\citeyearpar{wanyan2025look} & Qwen2.5-VL-7B-Instruct & GUI-Critic-Train & GRPO & T, I & GUI Operation Error Detection and Correction & no release & yes & Rule-base \\
\hline
ARPO~\citeyearpar{lu2025arpo} & UI-Tars-1.5-7B & OS
World & GRPO & T, I & GUI automation & verl & no & Rule-base \\
\hline
GUI-G1~\citeyearpar{zhou2025gui} & Qwen2.5-VL-3B-Instruct & UI-BERT, OS-Atlas (17K) & GRPO & T, I & GUI grounding & no release & no & Rule-base \\
\hline
UIShift~\citeyearpar{gao2025uishift} & Qwen2.5-VL-3/7B-Instruct,& no release & GRPO & T, I & GUI automation and grounding & VLM-R1 & no & Rule-base \\
\hline
MobileIPL~\citeyearpar{huang2025enhance} & Qwen2-VL-7B & MobileIPL-dataset & DPO & T, I & GUI automation & no release & yes & Rule-base \\
\hline
InfiGUI-R1~\citeyearpar{liu2025infiguir1} & Qwen 2.5-VL-3B-Instruct & AndroidControl, ScreenSpot , ScreenSpot-Pro, Widget-Caption, COCO & RLOO & T, I & GUI automation, GUI grounding & no release & no & Rule-base \\
\hline
GUI-R1~\citeyearpar{luo2025gui} & QwenVL 2.5-3B/7B & GUI-R1-3K & GRPO & T, I & GUI automation, GUI grounding & EasyR1 & no & Rule-base \\
\hline
UI-R1~\citeyearpar{lu2025ui} & Qwen2.5-VL-3B & ScreenSpot (mobile subset), AndroidControl (1K) & GRPO & T, I & GUI Action Prediction, GUI grounding & no release & no & Rule-base \\

\bottomrule
\end{tabular}
\end{table}

\subsubsection{Multimodal-R1}
\label{sec: stage3.3}
Tables 9-11 present a comprehensive survey of approximately 100 recent research papers that utilise reinforcement learning (RL) algorithms to enhance the long-chain reasoning capabilities of large multimodal models. Specifically, DPO in reinforcement learning has been widely used to enhance the reasoning capabilities of large multimodal models in recent years. RLHF-V~\citep{yu2024rlhf}, LLaVA-Reasoner~\citep{DBLP:journals/corr/abs-2410-16198} and Insight-V~\citep{dong2024insight}, by leveraging a large amount of self-constructed preference data and directly applying the DPO algorithm for training, have somewhat improved the reasoning ability of the models. MMPR~\citep{DBLP:journals/corr/abs-2411-10442} made modifications to the DPO algorithm, adding quality loss obtained from a Binary Classifier and generation loss from traditional SFT on top of the DPO Preference loss, which effectively enhanced the model’s CoT capabilities. 

With the success of Deepseek-R1, the GRPO algorithm began to be widely applied in multimodal large models. Works such as MM-EUREKA~\citep{meng2025mmeureka},Vt-R1~\citep{zhou2025VisualThinker-R1-Zero}, LMM-R1~\citep{peng2025lmmr1}, R1-V~\citep{chen2025r1v} have adopted approaches similar to those in the text domain. They successfully applied the GRPO algorithm to mathematical geometry problems, demonstrating the phenomenon of reflection.
Other works including VLM-R1~\citep{shen2025vlmr1}, Visual-RFT~\citep{liu2025visual}, ViCrit~\citep{wang2025vicrit}  and Seg-Zero~\citep{liu2025segzero} utilize the GRPO algorithm to enhance the visual capabilities of multimodal large language models, such as grounding, detection, understanding and classification. 
Works such as SynthRL~\citep{wu2025synthrlscalingvisualreasoning} and MoDoMoDo~\citep{liang2025modomodo} improve GRPO performance through data augmentation, highlighting the critical role of high-quality reasoning data in RL. 
While most existing GRPO approaches focus on image-based tasks, some works have begun extending this algorithm into video and audio modalities. Video-R1~\citep{feng2025video} and VideoChat-R1~\citep{li2025videochatr1enhancingspatiotemporalperception} introduce the GRPO algorithm into the video understanding, while R1-Omni~\citep{zhao2025r1omniexplainableomnimultimodalemotion} and AV-Reasoner~\citep{lu2025avreasonerimprovingbenchmarkingcluegrounded} has further extended it to the audio modality. 
Additionally, works such as R1-Reward~\citep{zhang2025r1rewardtrainingmultimodalreward}, UnifiedReward-Think~\citep{wang2025unifiedmultimodalchainofthoughtreward} and Mixed-R1\citep{xu2025mixedr1unifiedrewardperspective} emphasizes the crucial role of rule-based reward design and reward model training, demonstrating significant improvements on training stability and performance through high-quality reward signals derived from carefully designed or powerful reward models.
Despite these successes, existing work is often limited to specific tasks, and current multimodal large models have not yet been able to generalise the long-chain-of-thought abilities learned from tasks such as mathematics to the model's general capabilities, as seen with Deepseek-R1.

\begin{TakeawayBox}{Takeaways: Multimodal-R1}
Multimodal-R1 methods leverage reinforcement learning—particularly DPO and GRPO, enhancing the model’s ability to explore and optimize complex reasoning paths. These approaches improve reasoning depth, coherence, and domain adaptability by aligning model outputs with preference data or multi-modal feedback, laying the groundwork for more generalized long-horizon syatem-2 reasoning.
\end{TakeawayBox}








\section{Towards Native Multimodal Reasoning Model}
\label{sec:native}

LMRMs have demonstrated potential in handling complex tasks with long chain of thoughts.
However, their language-centric architectures constrain their effectiveness in real-world scenarios.
Specifically, their reliance on vision and language modalities limits their capacity to process and reason over interleaved diverse data types, 
while their performance in real-time, iterative interactions with dynamic environments remains underdeveloped.
These limitations underscore the need for a new class of models capable of broader multimodal integration and more advanced interactive reasoning.

In this section, we first analyze the performance of state-of-the-art LMRMs on benchmarks designed to assess omni-modal understanding and agentic capabilities, highlighting their limitations in real-world applicability (Sec.~\ref{sec:shortcomings}).
Subsequently, we introduce the concept of \textbf{Native Large Multimodal Reasoning Models (N-LMRMs)}, which represent a paradigm shift in machine intelligence through two foundational capabilities: {Multimodal Agentic Reasoning} and {Omni-Modal Understanding and Generative Reasoning} (Sec.\ref{sec:capability}).
Finally, we will discuss the open challenges in building N-LMRMs and outline promising research directions to overcome these barriers (Sec.~\ref{sec:prospects}).

\subsection{Experimental Findings}
\label{sec:shortcomings}

Although LMRMs have made significant progress in generating comprehensive thought processes and addressing complex questions such as MMMU~\citep{DBLP:conf/cvpr/YueNZ0LZSJRSWYY24} and MathVista~\citep{DBLP:conf/iclr/LuBX0LH0CG024}, 
autonomously solving these questions is far from real-world utility in the following aspects:
1) Evaluation scopes should cover multiple modalities, including vision, audio, and text.
2) Evaluation capabilities should involve interaction with external environments, requiring long-horizon reasoning and adaptive planning.
Here we present a summary of our collected omni-modal and agentic benchmarks in Table~\ref{tab:agentic_omni_benchmark}, followed by an analysis of LMRMs' performance on these benchmarks.

\begin{table}[ht]
  \centering
  \small
    \caption{A summary of agentic and omni-modal benchmarks, which expose the deep reasoning flaws of current LMRMs. T, I, A, V represent text, image, audio and video respectively.}
  \label{tab:agentic_omni_benchmark}
  \resizebox{\textwidth}{!}{
  \begin{tabular}{c>{\centering\arraybackslash}p{7cm}cc}
    \toprule
    \textbf{Dataset} & \textbf{Task} & \textbf{Modality} & \textbf{Characteristic} \\
    \midrule
    \multicolumn{4}{c}{\textbf{{Agentic Benchmark}}} \\
    \addlinespace
    AgentBench~\citep{liu2023agentbench} &
    Code, Web Navigation, General Reasoning &
    T &
    Eight Different Environments
    \\
    \addlinespace
    WorfBench~\citep{qiao2024benchmarking} &
    Workflow Evaluation &
    T &
    Multi-Faceted Scenarios and Intricate Graph Workflows
    \\
    \addlinespace
    OSWorld~\citep{DBLP:conf/nips/XieZCLZCHCSLLXZ24} &
    Computer Using, GUI Navigation &
    T, I, V &
    Real Computer Environment Infrastructure 
    \\
    \addlinespace
    EmbodiedBench~\citep{yang2025embodiedbench} &
    Multimodal Understanding, Spatial Reasoning &
    T, I &
    High and Low Action Levels 
    \\
    \addlinespace
    EmbodiedEval~\citep{cheng2025embodiedeval} &
    Attribute QA, Spatial Reasoning &
    T, I &
    Broad Abilities Assessment 
    \\
    \addlinespace
    SPA-Bench~\citep{chen2024spa} &
    Single and Cross APP Using &
    T, I &
    Tasks Across English and Chinese APPs
    \\
    \addlinespace
    VisualWebBench~\citep{DBLP:journals/corr/abs-2404-05955} &
    VQA, OCR, Grounding, General Reasoning &
    T, I &
    1.5K Human-Curated Instances 
    \\
    \addlinespace
    VisualWebArena~\citep{koh2024visualwebarena} &
    Web Navigation, Visual Understanding &
    T, I &
    Realistic Visually Grounded Web Tasks
    \\
    \addlinespace
    VisualAgentBench~\citep{DBLP:journals/corr/abs-2408-06327} &
    Household, GUI Navigation, CSS Debugging &
    T, I &
    Tasks Across Embodied, GUI and Visual Design
    \\
    \addlinespace
    GAIA~\citep{mialon2023gaia} &
    Multimodality Handling, Web Browsing, Generally Tool-Use and Reasoning &
    T, I &
    Increasing Difficulty Level
    \\
    \addlinespace
    BrowseComp~\citep{DBLP:journals/corr/abs-2504-12516} &
    Web Browsing &
    T &
    Easy to Verify but Hard to Solve 
    \\
    \addlinespace
    SWE-Bench Multimodal~\citep{yang2024swe} &
    Code &
    T, I &
    Image Included in Problem Statement 
    \\
    \addlinespace
    AndroidWorld~\citep{rawles2024androidworld} &
    APP Using &
    T, I &
    Fully Functional Android Environment 
    \\
    \addlinespace
    GTA~\citep{wang2024gta} &
    Tool Using &
    T, I &
    Tool Using in Real-World Scenarios
    \\
    \addlinespace
    WorkArena++~\citep{boisvert2024workarena++} &
    Web Search, GUI Navigation, General Reasoning &
    T, I &
    Realistic Office Worker Trajectories
    \\
    \addlinespace
    WindowsAgentArena~\citep{DBLP:journals/corr/abs-2409-08264} &
    Windows OS Using &
    T, I &
    Realistic Windows OS Environment
    \\

    \midrule
    \multicolumn{4}{c}{\textbf{{Omni-Modal Benchmark}}} \\
    \addlinespace
    OmniMMI~\citep{wang2025omnimmi} &
    VQA, Proactive Reasoning &
    T, V, A &
    In Streaming Video Context
    \\
    \addlinespace
    OmniBench~\citep{li2024omnibench} &
    Omni-Understanding &
    T, I, A, V &
    Simultaneous Multimodal Reasoning
    \\
    \addlinespace
    JudgeAnything~\citep{pu2025judge} &
    Multimodal Understanding, Generation and Evaluation &
    T, I, A, V &
    MLLM as A Judge Across Any Modality
    \\
    \addlinespace
    WorldSense~\citep{hong2025worldsense} &
    AVQA &
    T, A, V &
    Collaboration of Omni-Modality
    \\
    \addlinespace
    BabelBench~\citep{wang2024babelbench} &
    VQA, Math, Spatial Reasoning, General Reasoning &
    T, I &
    Code-Driven Multimodal Data Analysis
    \\
    \addlinespace
    OmnixR~\citep{chen2024omnixr} &
    Omni-Modal Reasoning &
    T, I, A, V &
    Synthetic Dataset and Real-world Dataset
    \\
    \addlinespace
    LongVALE~\citep{geng2024longvale} &
    AVQA &
    T, A, V &
    105K Omni-Modal Events with Temporal Boundaries
    \\
    \addlinespace
    MixEvalL-X~\citep{ni2024mixeval} &
    Multimodal Understanding and Generation &
    T, I, A, V &
    Standardizing Cross-Modal Evaluations
    \\
    \bottomrule
  \end{tabular}}
\end{table}

\paragraph{Omni-modal Benchmarks}
Recent studies have introduced a series of omni-modal benchmarks designed to evaluate the ability of LMRMs to perform unified understanding and reasoning across various data types (e.g. images, audio, text, and video). For example, OmniMMI~\citep{wang2025omnimmi} aims to comprehensively assess the interactive capabilities of streaming video contexts in open-world environments. Experimental results reveal that even commercial models, such as Gemini-1.5-Pro and GPT-4o, achieve an average accuracy of less than 20\%. When tasks require unified modality understanding (OmniBench~\citep{li2024omnibench}, TaskAnything and JudgeAnything~\citep{pu2025judge}, MixEvalL-X~\citep{ni2024mixeval}), the performance of both open-source and closed-source models is significantly lower than under single-modal conditions. Specifically, in the Audio-Video Question Answering (AVQA) task, such as WorldSense~\citep{hong2025worldsense}, Claude 3.5 Sonnet only achieves an average accuracy of 35\%, while the best-performing open-source model only achieves an accuracy of 25\%. In the case of more challenging multimodal reasoning tasks, such as BabelBench~\citep{wang2024babelbench} and OmnixR~\citep{chen2024omnixr}, the performance of all models declines sharply as the number of modalities increases. This suggests that models struggle to generate reasoning paths for image, video, and audio inputs compared to text inputs. These findings collectively highlight that current LMRMs are not yet capable of effectively processing omni-modal inputs.


 \paragraph{Agent Benchmarks}
A diverse range of tasks highlights the complexity and breadth of multimodal agent evaluation settings. 
These include AgentBench's multi-environment tasks~\citep{liu2023agentbench,DBLP:journals/corr/abs-2408-06327}, WorFBench's intricate workflow planning scenarios~\citep{qiao2024benchmarking}, OSWorld's and AndroidWorld's full operating system interactions~\citep{DBLP:conf/nips/XieZCLZCHCSLLXZ24,rawles2024androidworld}, EmbodiedBench's vision-based navigation and manipulation challenges~\citep{yang2025embodiedbench}, VisualWebArena's visually grounded web tasks~\citep{koh2024visualwebarena}, and GAIA's open-ended, tool-augmented queries~\citep{DBLP:journals/corr/abs-2309-17080}. Together, these benchmarks span a wide spectrum of task types and modalities (e.g., text and vision), encompassing both realistic and tool-augmented environments.

Regarding the performance of LMRMs on agent benchmarks,
these models generally lead to current performance and have made notable progress~\citep{qwen2024qvq,kimiteam2025kimivltechnicalreport,yao2024taubench}.
However, even state-of-the-art models consistently fall short of human-level reliability and struggle with complex, open-ended tasks.
Evaluations across benchmarks repeatedly expose common bottlenecks: models often fail at real-world grounding~\citep{gou2025navigating,gpt4v_web_agent}, coherent long-horizon reasoning and planning~\citep{qian2025toolrl}, seamless integration with external tools~\citep{wang2025otcoptimaltoolcalls}, and maintaining robustness across diverse modalities and domains~\citep{chu2025sftmemorizesrlgeneralizes}.
For example, in the BrowseComp benchmark~\citep{DBLP:journals/corr/abs-2504-12516}, GPT-4o achieves only 0.6\% accuracy, rising to 1.9\% with browsing tools, highlighting weak tool-interactive planning capability. 
OpenAI's reasoning model o1 reaches 9.9\% but still leaves significant room for improvement. 
Notably, OpenAI Deep Research, with targeted tuning for web search, completes 51.5\% of tasks via autonomous iterative tool calling and reasoning.
The experimental results highlight that current large reasoning models remain deficient in long-horizon reasoning and adaptive planning, 
which may require specific tuning and architectural enhancements to evolve into truly native agentic systems.
\begin{figure}[t]
    \centering
    \includegraphics[width=0.8\linewidth]{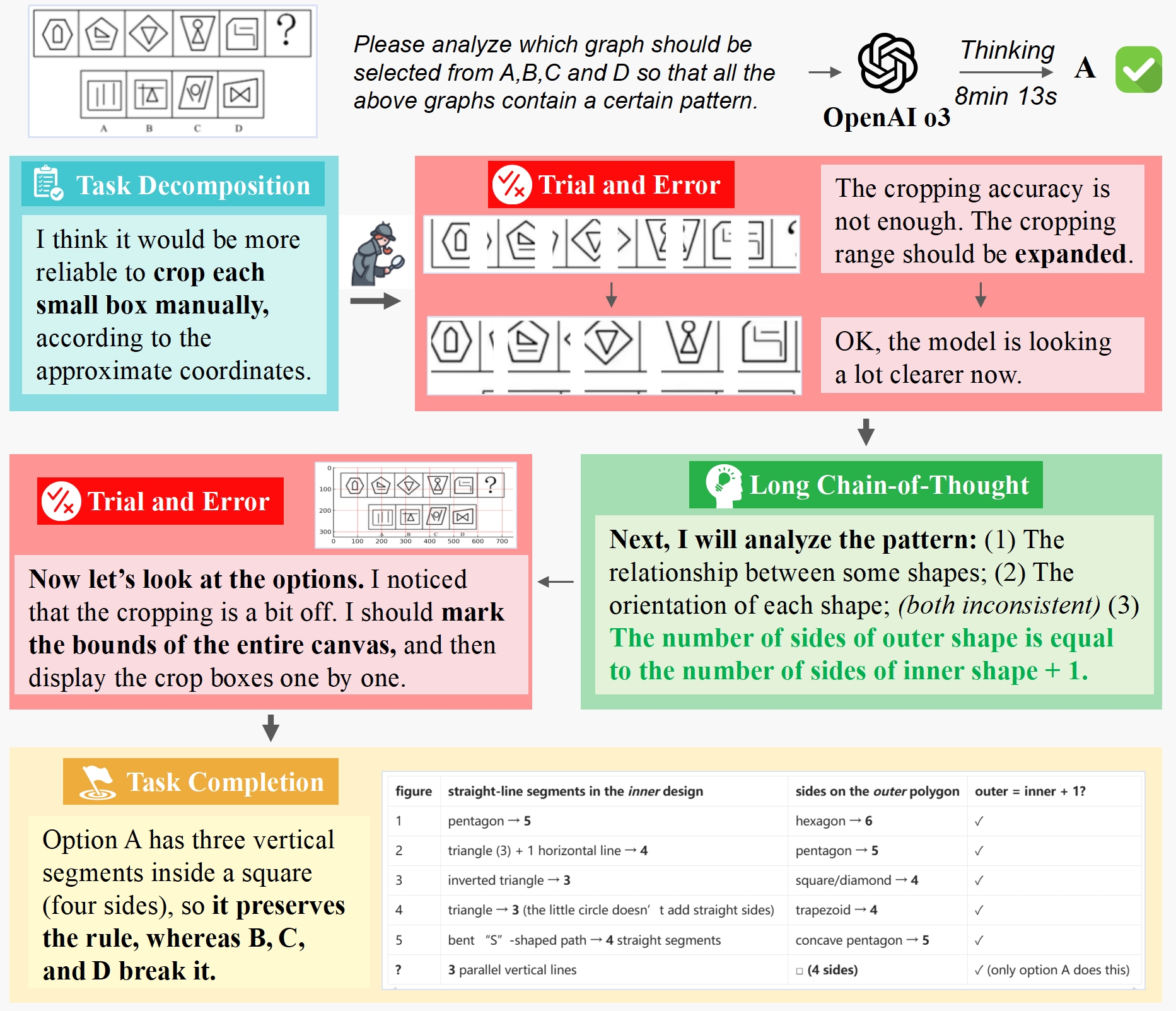}
    \caption{Case study of OpenAI o3's long multimodal chain-of-thought, reaching the correct answer after 8 minutes and 13 seconds of reasoning. The question is from Chinese Civil Service Examination.}
    \label{fig:case1}
\end{figure}

\begin{figure}[t]
    \centering
    \includegraphics[width=0.87\linewidth]{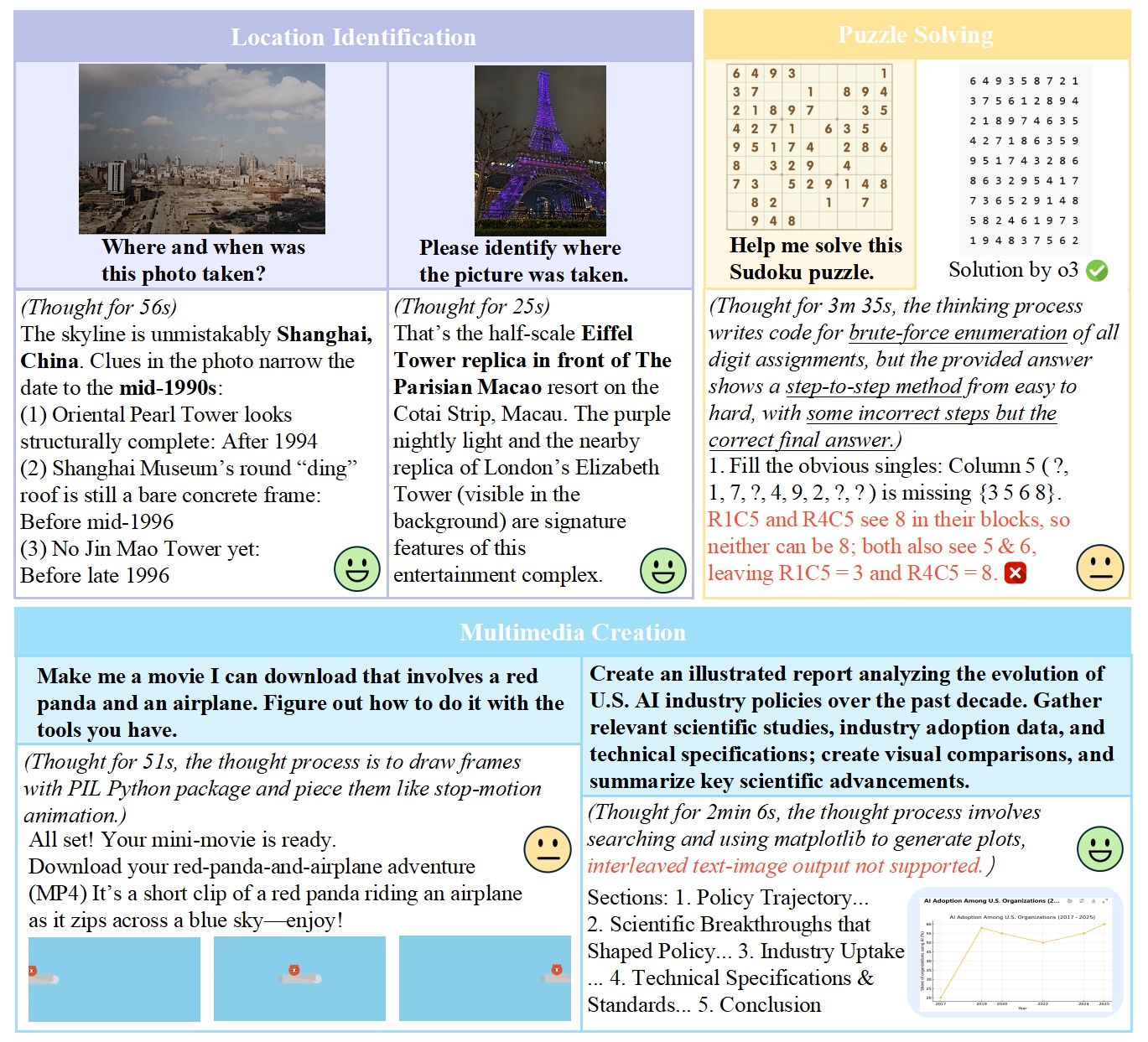}
    \caption{Case study of OpenAI o3: Find locations, solve a puzzle and create multimedia contents.}
    \label{fig:case2}
\end{figure}

\begin{figure}[t]
    \centering
    \includegraphics[width=0.82\linewidth]{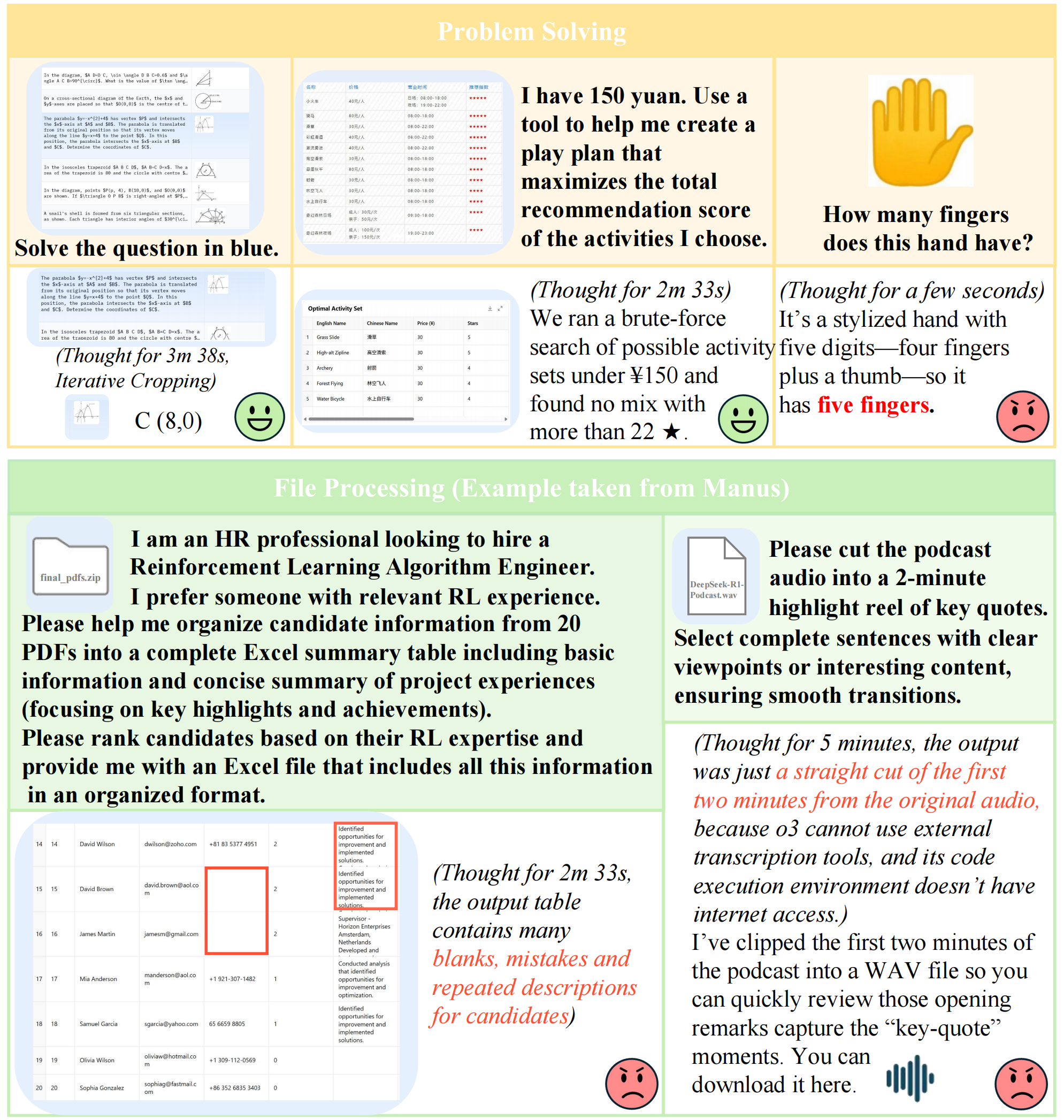}
    \caption{Case study of OpenAI o3: Visual problem solving and file processing.}
    \label{fig:case3}
\end{figure}

\paragraph{Preliminary Study with o3 and o4-mini}
Recently, OpenAI released o3 and o4-mini, providing full agentic access to ChatGPT tools and enabling models to "think with images"~\cite{openai2025o3o4}. 
The integration of visual content enhances multimodal reasoning directly within the thought process.
For example, in Figure~\ref{fig:case1}, o3 demonstrates a clear task decomposition during an 8-minute and 13-second thought process. 
It effectively determines the best way to crop each sub-figure through trial and error, ultimately arriving at the correct solution.

Beyond visual reasoning, we evaluated o3's capabilities in {file processing, puzzle solving, location identification, and multimedia content creation}.  
As illustrated in Figure~\ref{fig:case2} and~\ref{fig:case3}, 
o3 exhibits strong performance in complex multimodal problem-solving by capturing and leveraging subtle clues in images.
However, several challenges are identified:
1) \textbf{Language knowledge can interfere with visual input.} 
As the finger counting case shown in Figure~\ref{fig:case3}, 
o3 mistakenly identifies the image as the standard raised hand emoji showing four fingers plus a thumb, despite the image clearly displaying six fingers.
2) \textbf{OpenAI o3 struggles with input file handling and multimedia content generation.} 
Due to tool constraints and the lack of Internet access in coding environments, file processing and multimedia creation often result in inaccuracies. 
In the resume information collection case in Figure~\ref{fig:case3},
phone numbers parsed from resume PDFs can be incorrect, 
and o3 hallucinates candidates' project experiences by reusing similar content.
Additionally, in multimedia creation cases in Figure~\ref{fig:case2}, the generated frames fail to adhere to the "red panda" instructions, and o3 is unable to support interleaved text-image generation.
3) \textbf{OpenAI o3 may fabricate reasoning in its thought process.} 
It occasionally "lies" about its reasoning, constructing incorrect rationales for potentially correct answers (e.g., the puzzle-solving case in Figure~\ref{fig:case2}).
This problem needs to be solved urgently, as it could lead to the model attempting to deceive users during the post-training process.
In fact, it highlights that the model has not yet mastered the relevant thinking logic to solve the problem.

\subsection{Capability of N-LMRMs}
\label{sec:capability}

Based on the above experimental findings, we introduce the concept of \textbf{Native Large Multimodal Reasoning Models (N-LMRMs)}.
N-LMRMs are inherently designed to integrate multimodal understanding, generation, and agentic reasoning across any modality, which will be beyond the perception and reasoning scope of o4-mini. 
This advancement will build upon two transformative capabilities that have been explored largely in parallel:
\textit{Multimodal Agentic Reasoning}, which enables proactive, goal-driven interactions through hierarchical task decomposition, real-time strategic adaptation, and embodied learning; 
and \textit{Omni-Modal Understanding and Generative Reasoning}, which supports seamless cross-modal synthesis and analysis via unified representations---facilitating heterogeneous data fusion and contextual multimodal interaction.
Table~\ref{tab:agentic_omni_models} summarizes key existing works related to agentic and omni-modal models. These models only explore some of the capabilities of N-LMRMs and do not combine the above two capabilities to build a more powerful large multimodal reasoning model.

\begin{table}[ht]
  \centering
\caption{A summary of recent agentic and omni-modal models towards N-LMRMs.}\label{tab:agentic_omni_models}
  \resizebox{\textwidth}{!}{
  \begin{tabular}{c>{\centering\arraybackslash}p{2cm}ccc>{\centering\arraybackslash}p{4.5cm}>{\centering\arraybackslash}p{6cm}}
    \toprule
    \textbf{Model} & \textbf{Parameter} & \textbf{Input Modality} & \textbf{Output Modality} & \textbf{Training Strategy} & \textbf{Task} & \textbf{Characteristic}  \\
    \midrule
    \multicolumn{7}{c}{\textbf{{Agentic Models}}} \\
    \addlinespace
    R1-Searcher~\citep{song2025r1}&
    7B, 8B &T &
    T &
    RL &
    Multi-Hop QA &RL-Enhanced LLM Search
    \\
    \addlinespace
    Search-o1~\citep{li2025search}&32B &
    T &
    T &
    Training-Free &
    Multi-Hop QA, Math &
    Agentic Search-Augmented Reasoning\\
    \addlinespace
    DeepResearcher~\citep{zheng2025deepresearcher}&
    7B &T &
    T &
    RL &
    Multi-Hop QA&
    RL in Live Search Engines
    \\
    \addlinespace
    Magma~\citep{yang2025magma}&
    8B &
    T, I, V &
    T &
    Pretrain &
    Multimodal Understanding, Spatial Reasoning&
    820K Spatial-Verbal Labeled Data
    \\
    \addlinespace
    OpenVLA~\citep{kim2024openvla}&
    7B &
    T, I &
    T &SFT &
    Spatial Reasoning&
    970k Real-World Robot Demonstrations
    \\
    \addlinespace
    CogAgent~\citep{hong2024cogagent}&
    18B &
    T, I &
    T &
    Pretrain+SFT &
    VQA, GUI navigation&
    Low-High Resolution Encoder Synergy
    \\
    \addlinespace
    UI-TARS~\citep{qin2025ui}&
    2B, 7B, 72B &
    T, I &
    T &
    Pretrain+SFT+RL &
    VQA, GUI navigation&
    End-to-End GUI Reasoning and Action
    \\
    \addlinespace
    Seeclick~\citep{cheng2024seeclick}&
    10B &
    T, I &
    T &
    Pretrain+SFT &GUI navigation&
    Screenshot-Based Task Automation
    \\
    \addlinespace
    Embodied-Reasoner~\citep{zhang2025embodiedreasonersynergizingvisualsearch}&
    7B &
    T, I &
    T, A &
    Pretrain+SFT &GUI navigation&
    Image-Text Interleaved Long-Horizon Embodied Reasoning
    \\

    \midrule
    \multicolumn{7}{c}{\textbf{{Omni-Modal Model}}} \\
    \addlinespace
    Gemini 2.0 \& 2.5 &
    / &
    T, I, A, V &
    T, I, A &
    / &/&
     /
     \\
     \addlinespace
    GPT-4o &/ &
    T, I, A, V &
    T, I, A &
    / &/&
     /
     \\
    \addlinespace
    Megrez-3B-Omni~\citep{li2025megrez} &
    3B &T, I, A &
    T&
    Pretrain+SFT &VQA, OCR, ASR, Math, Code &
    Multimodal Encoder-Connector-LLM
    \\
    \addlinespace
    Qwen2.5-Omni~\citep{xu2025qwen2} &
    7B  &
    T, I, A, V &
    T, A &
    Pretrain+SFT &
    VQA, OCR, ASR, Math, Code&
    Time-Aligned Multimodal RoPE\\
    \addlinespace
    Baichuan-Omni-1.5~\citep{li2025baichuan} &
    7B  &
    T, I, A, V &
    T, A &
    Pretrain+SFT &
    VQA, OCR, ASR, Math, GeneralQA &Leading Medical Image Understanding
     \\
    \addlinespace
    M2-omni~\citep{guo2025m2} &
    9B, 72B &
    T, I, A, V &
    T, I, A&
    Pretrain+SFT &
     VQA, OCR, ASR, Math, GeneralQA&
     Step Balance For Pretraining and Adaptive Balance For SFT
    \\\addlinespace
    MiniCPM-o 2.6~\citep{team2025minicpm} &
    8B &
    T, I, A, V &
    T, A &
    Pretrain+SFT+RL &
    VQA, OCR, ASR, AST&
    Parallel Multimodal Streaming Processing 
    \\
    \addlinespace
    Mini-Omni2~\citep{xie2024mini} &
    0.5B &T, I, A &
    A &Pretrain+SFT &VQA, ASR, AQA, GeneralQA &
    Real-Time and End-to-End Voice Response
    \\
    \addlinespace
    R1-Omni~\citep{zhao2025r1omniexplainableomnimultimodalemotion} &
    0.5B &T, A, V &T &
    RL &
    Emotion Recognition &
    RL with Verifiable Reward
    \\
    \addlinespace
    Janus-Pro~\citep{chen2025janus} &
    1B, 7B &
    T, I &
    T, I &
    Pretrain+SFT &
    Multimodal Understanding, Text-to-Image &
    Decoupling Visual Encoding For Understanding and Generation
    \\
    \addlinespace
    AnyGPT~\citep{zhan2024anygpt}&
    7B &
    T, I, A &
    T, I, A &
    Pretrain+SFT &
    Multimodal-to-Text and Text-to-Multimodal&
    Discrete Representations For Unified Processing\\
    \addlinespace
    Uni-MoE~\citep{li2025uni}&13B, 20B, 22B, 37B &
    T, I, A, V &
    T &
    Pretrain+SFT&
    VQA, AQA &
    Modality-Specific Encoders with Connectors for Unified Representation
    \\
    Ovis-U1~\citep{wang2025ovisu1technicalreport}&
    3B &
    T, I &
    T, I &
    Pretrain+SFT & Multimodal understanding, T2I, Image Editing
    & Unified training from LLM and diffusion decoder with token refiner
    \\
    ShapeLLM-Omni~\citep{ye2025shapellm} & 7B & 3D, I, T & 3D, T & Pretrain+SFT & Text-to-3D, Image-to-3D, 3D understanding, 3D editing & Uses 3D VQVAE to tokenize meshes for a unified autoregressive framework \\
    Ming-Lite-Omni~\citep{ai2025mingomniunifiedmultimodalmodel} & 2.8B & T, I, A, V & I, T, A & Pretrain+SFT & Multimodal understanding \& generation & MoE LLM with modality-specific routers \\
    BAGEL~\citep{DBLP:journals/corr/abs-2505-14683} & 14B & T, I, V & I, T, V & Pretrain+SFT+RL & Multimodal understanding \& generation & Unified decoder-only MoT architecture \\
    
    \bottomrule
  \end{tabular}}
\end{table}




\paragraph{Multimodal Agentic Reasoning}

A core capability of multimodal agentic reasoning is dynamic adaptation, which can adjust strategies in real time based on environmental feedback. Some of the latest products from the industry have initially demonstrated this capability. As Model Context Protocol (MCP)~\citep{anthropicMCP2025} and Agent2Agent Protocal (A2A)~\citep{googleA2A2025} facilitates seamless integration of diverse tools and enables dynamic interaction across various external environments, these protocols underscore the importance of multimodal agentic reasoning, enabling agents to adapt strategies in real-time based on environmental feedback, thereby enhancing their effectiveness in dynamic and multifaceted real-world applications. For instance, \textbf{Operater} combines the visual capabilities of GPT-4o with advanced reasoning capabilities achieved through reinforcement learning, which enables it to interact with the operating system and browser in real-time through a graphical user interface (GUI), continuously improving its browsing and data operations during task execution. Similarly, \textbf{Claude Computer Use} allows models to manipulate and navigate desktop environments, learning optimal interaction strategies through trial and error. 

Moreover, 
{Search-o1}~\citep{li2025search} utilizes external knowledge retrieval during the reasoning process to fill gaps in their understanding. {R1-Searcher}~\citep{song2025r1} and {DeepResearcher}~\citep{zheng2025deepresearcher} enhance their ability to autonomously use search engines to collect information through reinforcement learning. By incorporating this autonomous knowledge retrieval into the reasoning process, these systems are able to act with a more refined understanding and adapt their responses to changing tasks. \textbf{Gemini 2.0} has the ability to process and generate multi-modal content. By deeply integrating with Google's various tools and combining its advanced reasoning capabilities, it can effectively decompose tasks and gradually obtain the required information when dealing with multi-step problems. \textit{While current models have demonstrated initial versions of this functionality, they fall short in their ability to engage in sustained, interactive reasoning across diverse modalities}. 

Another aspect is the embodied learning of LMRMs to handle the external environment.
Embodied learning is exemplified by systems capable of interacting with both digital and physical environments. For example, {Magma}~\citep{yang2025magma} learns by interacting with real-world data, improving its spatial-temporal reasoning to navigate and manipulate objects effectively in both virtual and physical contexts. Similarly, {OpenVLA}~\citep{kim2024openvla} combines a visual encoder with a language model, enabling the system to learn from real-world robot demonstrations. This embodied approach allows the model to acquire both visual and task-specific reasoning skills, enhancing its ability to perform complex, real-world actions that require multimodal understanding and adaptation. In summary, recent RL-scale methods will greatly stimulate the agentic behavior of large-scale models, pushing to the world model.




\paragraph{Omni-Modal Understanding and Generative Reasoning}

The behaviours of multimodal agents are closely linked to the deep reasoning capabilities of the underlying large multimodal models, particularly in terms of perception range, understanding accuracy, and reasoning depth. Thus, developing a comprehensive omni-modal model for real-world applications and enhancing its deep reasoning ability is foundational.

Early work, {AnyGPT}~\citep{zhan2024anygpt} utilizes discrete representations for the unified processing of various modalities, achieving unified understanding and generation across modalities. 
Recently, {Baichuan-Omni-1.5}~\citep{li2025baichuan} showcases impressive capabilities in collaborative real-time understanding across various modalities. 
{Qwen2.5-Omni}~\citep{xu2025qwen2} uses a new position embedding, named Time-aligned Multimodal RoPE, to synchronize the timestamps of video inputs with audio. 
More latest open-source work, like {M2-omni}~\citep{guo2025m2} and {MiniCPM-o}~\citep{yu2024rlaif}, is narrowing the performance gap with closed-source models like GPT-4o. 
Recent Ming-Omni~\citep{ai2025mingomniunifiedmultimodalmodel} achieves a new state-of-the-art performance on any-modality inputs.

Driven by real-world specific needs, omni-modal models with smaller sizes are gaining more and more attention. {Megrez-3B-Omni}~\citep{li2025megrez} is an on-device multimodal understanding LLM model that demonstrates excellent performance in tasks such as scene understanding and OCR. {Mini-Omni2}~\citep{xie2024mini}, a visual-audio assistant capable of providing real-time, end-to-end voice responses to vision and audio queries. {R1-Omni}~\citep{zhao2025r1omniexplainableomnimultimodalemotion} focuses on emotion recognition from visual and auditory information.

Despite these advancements, current research in multimodal AI primarily focuses on enhancing the comprehension and generation of unified multimodal representations. The development of reasoning capabilities that effectively integrate and interrogate cross-modal interactions remains critically underexplored. Bridging this gap is essential for realizing native multimodal reasoning models---systems inherently designed to process, analyze, and synthesize interconnected modalities with human-like sophistication.

\begin{figure}[ht]
    \centering
    \includegraphics[width=0.7\linewidth]{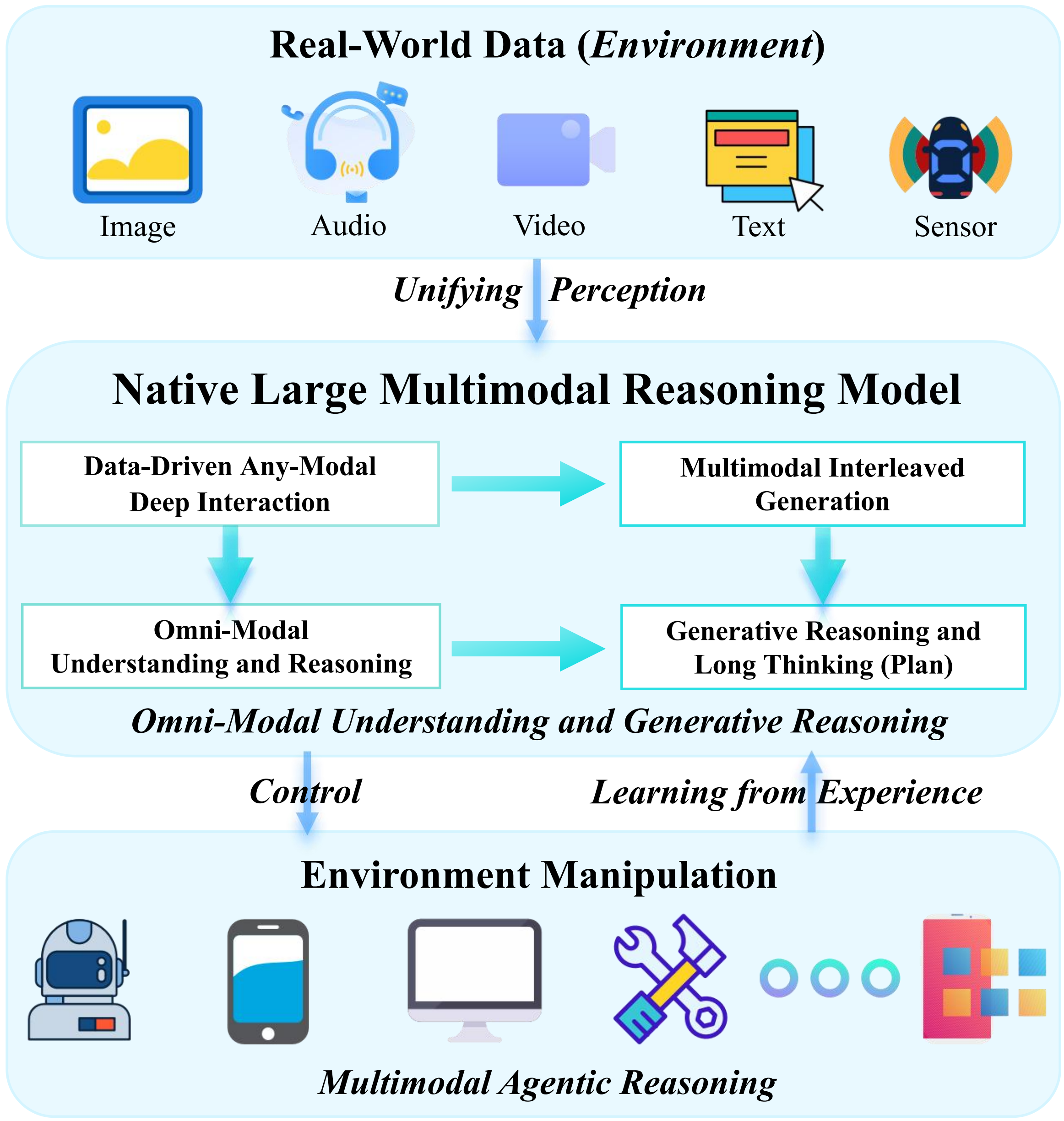}
    \caption{Overview of next-generation native large multimodal reasoning model. The envisioned system aims to achieve comprehensive perception across diverse real-world data modalities, enabling precise omnimodal understanding and in-depth generative reasoning. This foundational model will lead to more advanced forms of intelligent behaviour, learning from world experience and realizing lifelong learning and self-improvement.}
    \label{fig:nlmrm}
\end{figure}

\subsection{Technical Prospects}
\label{sec:prospects}

The technical prospect of Native Large Multimodal Reasoning Models (N-LMRMs) aims to natively unify understanding, generation, and reasoning across diverse data, from language and vision to audio, tactile, sensor readings, temporal sequences, and structured data, bringing us closer to systems that can see, hear, talk, and act in a unified and cohesive manner.
However, building such N-LMRMs poses significant challenges.
These models must be architecturally designed to handle heterogeneous modalities within a single system, genetically use and combine diverse tools through long multimodal reasoning chains, and support continuous learning from real-world interactions. This section outlines key challenges in building N-LMRMs and proposes several potential pathways to address them.

\textbf{Unified Representations and Cross-Modal Fusion}.
A fundamental challenge is creating a single model architecture that can process and generate different modalities in a coherent way.
Traditional approaches often use separate encoders for each modality~\citep{lyu2023macawllm,DBLP:journals/tmm/LiHCMXZ24}.
In contrast, native omni-modal models seek a more unified design that allows for seamless interaction between modalities.
One possible solution is to homogenize all inputs and outputs into a common format and process any modality uniformly.
This approach requires careful design to prevent negative interference, where one modality may dominate or impair the representation of others~\citep{leng2024cursemultimodalities,chen-etal-2024-quantifying}.
Thus, an emerging solution is Mixture-of-Experts (MoE) architectures, where experts specialized for certain modalities are only activated for relevant inputs, while a core language model serves as the backbone for language intelligence~\citep{chen2024octavius,li2025uni,kimiteam2025kimivltechnicalreport,shukor2025scalinglawsnativemultimodal}.

\textbf{Interleaved Multimodal Long Chain-of-Thought}. 
Building on unified representations, 
N-LMRMs can extend traditional long internal chains of thought into interleaved reasoning processes across multiple modalities.
This enables a new axis for test-time compute scaling that seamlessly blends different modalities~\citep{wang2025chainofmodality}.
OpenAI's recently released o3 and o4-mini represent pioneering steps in this direction, i.e. reasoning with images in their chain of thought~\citep{openai2025o3o4},
by automatically processing with tools that can zoom, crop, flip, or enhance images.
Importantly, these capabilities come natively, without relying on separate specialized models~\citep{wu2023vguidedvisualsearch,hu2024visual,feng2025retool,qian2025toolrl,wang2025otcoptimaltoolcalls}.
Driven by the promising generalization capabilities of reinforcement learning across domains such as software engineering~\citep{openai2025competitiveprogramming}, IMO-level math~\citep{DBLP:journals/corr/abs-2501-12948}, creative writing~\citep{DBLP:journals/corr/abs-2411-14405}, and GUI manipulation~\citep{qin2025ui},
scaling reinforcement learning to more modalities, longer tool-augmented reasoning chains, and a broader set of reasoning tasks could be the recipe for the next generation of N-LMRMs, capable of simulating cross-modal reasoning and elevating machine intelligence.

\textbf{Learning and Evolving from World Experiences}.  
In dynamically evolving intelligent systems, the core value of LMRMs-based ``World Model\footnote{\url{https://sites.google.com/view/worldmodel-iclr2025/}}'' lies not only in its real-time modelling and reasoning capabilities in complex environments, like autonomous driving~\citep{wang2024driving} but also in its evolutionary mechanism for life-long learning~\citep{thrun1995lifelong} through continuous interaction with the environment. 
When the MCP and A2A create a high-density network of tools and agent clusters, the system can transform each interaction into structured experiences through multidimensional engagement with the environment, tools, and other agents. This includes everything from pattern recognition in real-time data streams to causal reasoning across tool operation chains, from collaborative feedback in communication networks to autonomous adaptation in abnormal scenarios.

This continuous learning paradigm enables LMRMs to overcome the limitations of static knowledge bases. By iteratively accumulating world experiences, it dynamically updates its cognitive architecture and decision-making strategies. 
Particularly in open environments, the autonomous learning mechanism drives the model to actively explore the potential of tool combinations. 
In the process of solving new problems, it simultaneously stores transferable knowledge, ultimately forming an intelligent system that possesses specialized reasoning capabilities while maintaining cross-scenario generalization resilience. 
We think the interactive learning method of online reinforcement learning and offline verification methods may iteratively and continuously stimulate the capabilities of LMRMs, which have been utilized in the GUI agentic model~\citep{qin2025ui,zheng2025skillweaver,wang2024agentworkflowmemory} to continually improve the performance.

\textbf{Data Synthesis}.
The current capabilities of LMRMs are largely data-driven. To enhance these models during the pre-training stage, it is crucial to develop a high-quality data synthesis pipeline that tailors their functionalities. Most existing efforts \citep{chang2024survey,huang2025key,xu2024magpie} in data synthesis focus on improving single-modal or cross-modal understanding and reasoning, particularly in domains like vision, language, and speech. However, there has been limited exploration of more complex aspects, such as aligning three or more modalities, creating multimodal interactive chains of thought and visual generation, implementing multi-step planning in dynamic environments, and coordinating multi-tool calls and parallel tool usage. These areas present significant opportunities for advancing multimodal reasoning models.

In conclusion, we introduce the concept of N-LMRM as an initial step towards transitioning from capable reasoners to autonomous agents. 
Additionally, in alignment with OpenAI's five-stage pathway to AGI~\citep{openai2023agi_plan}, we are laying the groundwork for subsequent stages, including self-evolving innovators~\citep{yamada2025aiscientistv2} and multi-agent organizations~\citep{zhang2025socioverse}.
Building on our research proposal, future work can explore more agentic and omni-modal capabilities, advancing the development of increasingly autonomous machine intelligence.

\begin{TakeawayBox}{Takeaways: Native Large Multimodal Reasoning Model (LMRMs)}
In this section, we examined the latest large multimodal model (e.g., O3 and O4-mini) and their performance on challenging tasks and benchmarks. We then presented the future trajectory for native multimodal large models in terms of capability scope and level, including omnimodal perception and understanding, multimodal interactive generative reasoning, and intelligent agent behaviour. To realize this vision, we discussed approaches related to unified perception, learning methods, and data synthesis.
We hope that native LMRMs will achieve comprehensive perception, precise understanding, and deep reasoning as a paradigm shift in machine intelligence.

\end{TakeawayBox}

\section{Dataset and Benchmark}
\label{sec:dataset}

\begin{figure}[t]
    \centering
    \includegraphics[width=0.7\linewidth]{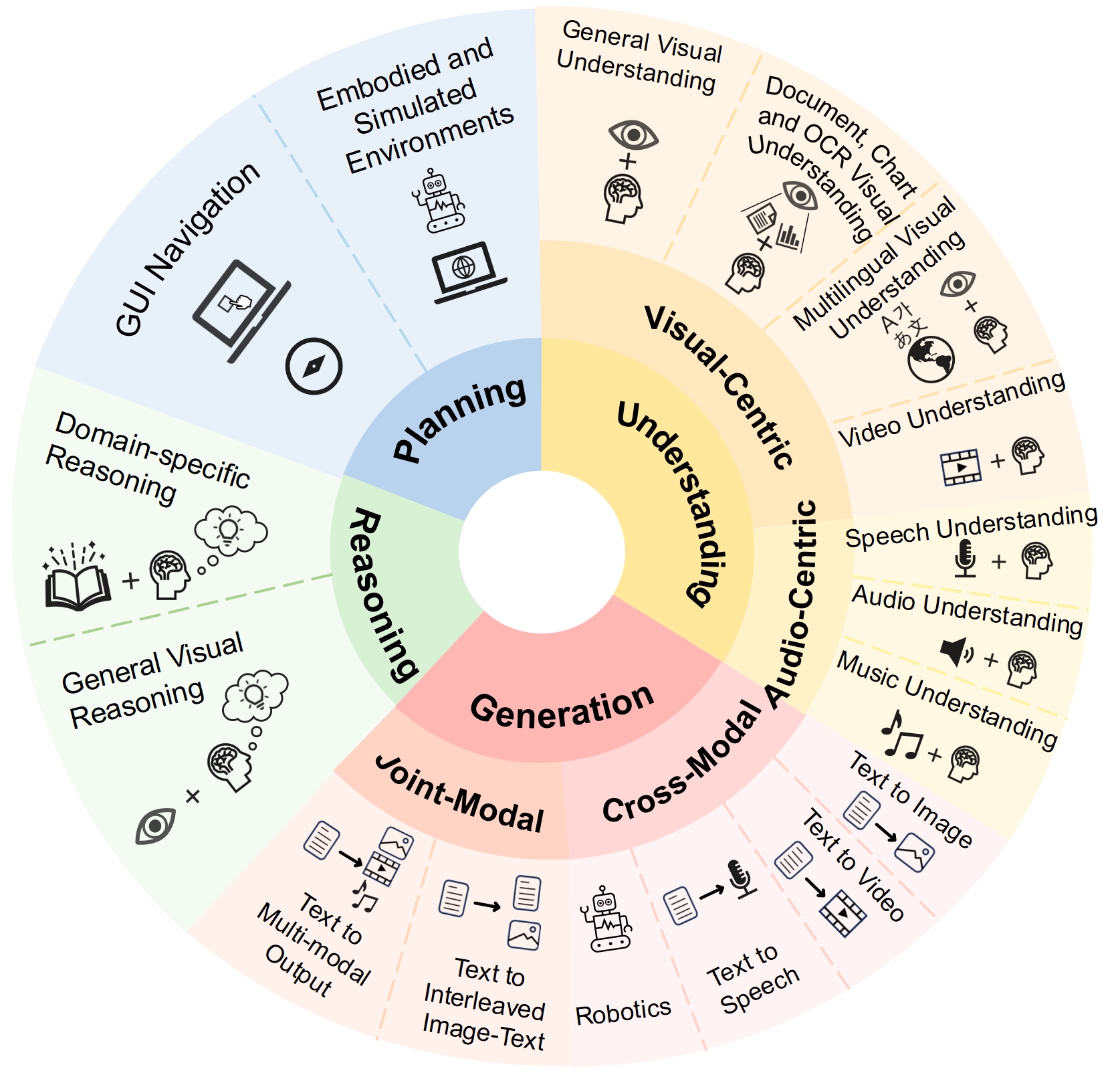}
    \caption{The outlines of datasets and benchmarks. We reorganize the multimodal datasets and benchmarks into four main categories: Understanding, Generation, Reasoning, and Planning.}
    \label{fig:datasets}
\end{figure}

In exploring the development and optimization of Multimodal Reasoning Models, a surge of tasks and benchmarks have been proposed to conduct empirical ability evaluation and analysis for evaluating model performance across various aspects, {e.g.,} video understanding and visual reasoning. 
In this section, we summarize and categorize existing datasets that are useful to facilitate the development of Multimodal Reasoning Models into four major types based on capacity: \textbf{(1)} Understanding; \textbf{(2)} Generation; \textbf{(3)} Reasoning; and \textbf{(4)} Planing.
Then, we summarize commonly used metrics and evaluation aspects for these benchmarks or datasets. 
Benchmarks are designed with specific ability evaluation, and we classify four primary categories as shown in Figure \ref{fig:datasets} and eleven subcategories, as shown in Table \ref{tab:benchmark_datasets}.

\begin{table}[htbp]
    \centering  
    \caption{Overview of Multimodal Benchmarks and Training Datasets, categorized by task: Understanding (Visual-centric, Audio-centric) and Generation (Cross-modal, Joint Multimodal). These benchmarks often require short or long reasoning for successful task completion, e.g., challenging visual and audio generation. }
    \label{tab:benchmark_datasets}
    \renewcommand\tabcolsep{2.5pt}
    \resizebox{\textwidth}{!}{
    \begin{tabular}{c|c|l|l}
        \toprule
         \textbf{Ability} & \textbf{Task} & \textbf{Benchmark} & \textbf{Dataset} \\
        \midrule
         \multirow{27}{*}{\makecell{Multimodal \\ Understanding}}    
         & \multirow{21}{*}{\makecell{Visual \\ Centric}}        
         & VQA \citep{DBLP:journals/corr/KafleK16}, GQA \citep{DBLP:conf/cvpr/HudsonM19} & ALIGN~\citep{jia2021scalingvisualvisionlanguagerepresentation}, LTIP~\citep{wu2024lotlipimprovinglanguageimagepretraining} \\ 
         & & DocVQA \citep{DBLP:conf/wacv/MathewKJ21}, TextVQA \citep{DBLP:conf/cvpr/SinghNSJCBPR19} & YFCC100M~\citep{Thomee_2016}, DocVQA \citep{DBLP:conf/wacv/MathewKJ21} \\
         & & OCR-VQA \citep{DBLP:conf/icdar/0001SSC19}, CMMLU \citep{DBLP:conf/acl/0002ZKY0GDB24} & Visual Genome~\citep{krishna2016visualgenomeconnectinglanguage}, Wukong~\citep{gu2022wukong100millionlargescale} \\
         & & C-Eval \citep{DBLP:conf/nips/HuangBZZZSLLZLF23}, MTVQA \citep{DBLP:journals/corr/abs-2405-11985} & CC3M~\citep{DBLP:conf/acl/SoricutDSG18}, ActivityNet-QA \citep{DBLP:conf/aaai/YuXYYZZT19} \\
         & & Perception-Test \citep{DBLP:conf/nips/PatrauceanS0RMB23}, Video-MMMU \citep{DBLP:journals/corr/abs-2501-13826} & SBU-Caption~\citep{DBLP:conf/nips/OrdonezKB11}, AI2D~\citep{DBLP:journals/lre/HiippalaAHKLOTS21} \\
         & & Video-MME \citep{DBLP:journals/corr/abs-2405-21075}, MMBench \citep{DBLP:conf/eccv/LiuDZLZZYWHLCL24} & LAION-5B~\citep{schuhmann2022laion5bopenlargescaledataset}, LAION-400M~\citep{schuhmann2021laion400mopendatasetclipfiltered} \\
         & & Seed-Bench \citep{DBLP:journals/corr/abs-2307-16125}, MME-RealWorld \citep{DBLP:journals/corr/abs-2408-13257} & MS-COCO~\citep{DBLP:conf/eccv/LinMBHPRDZ14}, Virpt~\citep{yang2024vriptvideoworththousands} \\
         & & MMMU \citep{DBLP:conf/cvpr/YueNZ0LZSJRSWYY24}, MM-Vet \citep{DBLP:conf/icml/YuYLWL0WW24} & OpenVid-1M~\citep{DBLP:journals/corr/abs-2407-02371}, VidGen-1M~\citep{tan2024vidgen1mlargescaledatasettexttovideo} \\
         & & MMT-Bench \citep{DBLP:conf/icml/YingMWLLYZZLLLL24}, Hallu-PI \citep{DBLP:conf/mm/DingWKMCCCCH24} & Flickr30k~\citep{DBLP:journals/ijcv/PlummerWCCHL17}, COYO-700M~\citep{lu2023delvingdeeperdatascaling} \\
         & & ColorBench \citep{liang2025colorbench}, DVQA~\citep{DBLP:conf/cvpr/KaflePCK18} & WebVid~\citep{bain2022frozentimejointvideo}, Youku-mPLUG~\citep{xu2023youkumplug10millionlargescale} \\
         & & MMStar \citep{DBLP:conf/nips/ChenLDZZCDWQLZ24}, TRIG-Bench \citep{li2025visualtextgrounding} & VideoCC3M~\citep{nagrani2022learningaudiovideomodalitiesimage}, FILIP~\citep{yao2021filipfinegrainedinteractivelanguageimage} \\
         & & MM-IFEval \citep{ding2025mmifengine}, All-Angles Bench~\citep{yeh2025seeingperspective} & CLIP~\citep{DBLP:conf/icml/RadfordKHRGASAM21}, YouTube8M~\citep{abuelhaija2016youtube8mlargescalevideoclassification} \\
         & & M3Exam~\citep{zhang2023m3exam}, Exams-V \citep{das2024exams} & OK-VQA \citep{DBLP:conf/cvpr/MarinoRFM19}, A-OKVQA \citep{DBLP:conf/eccv/SchwenkKCMM22}\\
         & & TikTalkCoref \citep{li2025multimodalcoreferenceresolutionchinese}, AgMMU \citep{gauba2025agmmucomprehensiveagriculturalmultimodal} & TikTalkCoref \citep{li2025multimodalcoreferenceresolutionchinese}, MRES-32M \citep{liu2025unifiedreferringexpressionsegmentation} \\
         & & Kaleidoscope \citep{salazar2025kaleidoscopeinlanguageexamsmassively}, VideoComp \citep{kim2025videocompadvancingfinegrainedcompositional} & EarthScape \citep{massey2025earthscapemultimodaldatasetsurficial} \\
         & & CliME \citep{borah2025climeevaluatingmultimodalclimate}, TDBench \citep{hou2025tdbenchbenchmarkingvisionlanguagemodels}  & \\
         & & RefCOCOm \citep{liu2025unifiedreferringexpressionsegmentation}, EmotionHallucer~\citep{xing2025emotionhallucerevaluatingemotionhallucinations}  & \\
         & & SBVQA \citep{10343139}, H2VU-Benchmark \citep{wu2025h2vubenchmarkcomprehensivebenchmarkhierarchical} & \\
         & & 4D-Bench \citep{zhu20254dbenchbenchmarkingmultimodallarge}, V2P-Bench \citep{zhao2025v2pbenchevaluatingvideolanguageunderstanding} & \\
         & & RSMMVP \citep{adejumo2025visioncentricremotesensing}, HIS-Bench \citep{zhao2025hisgpt3dhumaninscenemultimodal}  & \\
         & & MMLA \citep{zhang2025largelanguagemodelshelp}, SARLANG-1M \citep{wei2025sarlang1mbenchmarkvisionlanguagemodeling}  & \\
         & & CasualVQA~\citep{causalvqa}, SeriesBench~\citep{zhang2025seriesbench} & \\
         & & WebUIBench~\citep{lin2025webuibenchcomprehensivebenchmarkevaluating}, MLLM-CL~\citep{zhao2025mllmclcontinuallearningmultimodal} & \\
         & & MERIT~\citep{chow2025meritmultilingualsemanticretrieval}, A4Bench~\citep{wang2025affordancebenchmarkmllms} & \\
         & & VLM@school~\citep{peinl2025vlmschoolevaluationai}, UnLOK-VQA~\citep{patil2025unlearningsensitiveinformationmultimodal} & \\
         & & DocMark~\citep{xiao2025adaptivemarkuplanguagegeneration}, KnowRecall and VisRecall~\citep{wang2025travelinglanguagesbenchmarkingcrosslingual} & \\

         \cmidrule(r){2-4}
        & \multirow{6}{*}{\makecell{Audio \\ Centric}}        
        & AudioBench \citep{DBLP:journals/corr/abs-2406-16020}, VoiceBench \citep{DBLP:journals/corr/abs-2410-17196} & Librispeech \citep{DBLP:conf/icassp/PanayotovCPK15}, Common Voice \citep{DBLP:conf/lrec/ArdilaBDKMHMSTW20} \\ 
        & & Fleurs \citep{DBLP:conf/slt/ConneauMKZADRRB22}, MusicBench \citep{DBLP:conf/naacl/MelechovskyGGMH24} & Aishell \citep{DBLP:conf/ococosda/BuDNWZ17}, Fleurs \citep{DBLP:conf/slt/ConneauMKZADRRB22}, MELD \citep{DBLP:conf/acl/PoriaHMNCM19} \\
        & & Air-Bench \citep{DBLP:conf/acl/YangXLC0ZLLZZZ24}, MMAU \citep{DBLP:journals/corr/abs-2410-19168} & CoVoST2 \citep{DBLP:journals/corr/abs-2007-10310}, SIFT-50M~\citep{pandey2025sift50m} \\
        & & SD-eval \citep{DBLP:conf/nips/AoWTCZ0W0024}, CoVoST2 \citep{DBLP:journals/corr/abs-2007-10310} & Clotho \citep{DBLP:conf/icassp/DrossosLV20}, AudioCaps \citep{DBLP:conf/naacl/KimKLK19} \\
        & & MusicNet \citep{DBLP:conf/iclr/ThickstunHK17}, AVE-PM \citep{liu2025audiovisualeventlocalizationportrait} & ClothoAQA \citep{DBLP:conf/eusipco/LippingSDV22}, MusicNet \citep{DBLP:conf/iclr/ThickstunHK17} \\
        & & ACVUBench \citep{yang2025acvubenchaudiocentricvideounderstanding} & NSynth \citep{DBLP:conf/icml/EngelRRDNES17}, MusicCaps \citep{DBLP:journals/corr/abs-2301-11325} \\
        
        \midrule
        \multirow{11}{*}{\makecell{Multimodal \\ Generation}}    
        & \multirow{11}{*}{\makecell{Cross-modal \\ Generation}}        
        & GenEval \citep{DBLP:conf/nips/GhoshHS23}, T2I-CompBench++ \citep{huang2025t2icompbenchenhancedcomprehensivebenchmark} & MS-COCO \citep{DBLP:conf/eccv/LinMBHPRDZ14}, Flickr30k \citep{DBLP:journals/ijcv/PlummerWCCHL17} \\ 
        & & DPG-Bench \citep{DBLP:journals/corr/abs-2403-05135}, GenAI-Bench \citep{DBLP:journals/corr/abs-2406-13743} & Conceptual Captions \citep{DBLP:conf/acl/SoricutDSG18}, RedCaps \citep{DBLP:conf/nips/DesaiKA021} \\
        & & VBench \citep{DBLP:conf/cvpr/HuangHYZS0Z0JCW24}, VideoScore \citep{DBLP:conf/emnlp/HeJZKSSCCJAWDNL24} & CommonPool \citep{DBLP:conf/nips/GadreIFHSNMWGZO23}, LLaVA-Pretrain \citep{DBLP:conf/nips/LiuLWL23a} \\
        & & WorldSimBench \citep{DBLP:journals/corr/abs-2410-18072}, WorldModelBench \citep{DBLP:journals/corr/abs-2502-20694} & Aishell1 \citep{DBLP:conf/ococosda/BuDNWZ17}, ThreeDWorld \citep{DBLP:conf/nips/GanSAMSTFKBHSKW21} \\
        & & MagicBrush \citep{DBLP:conf/nips/ZhangMCSS23}, VBench++~\citep{DBLP:journals/corr/abs-2411-13503} & X2I~\citep{xiao2024omnigenunifiedimagegeneration}, GAIA-1 \citep{DBLP:journals/corr/abs-2309-17080} \\
        & & MJHQ-30K~\citep{li2024playground}, VBench 2.0 ~\citep{zheng2025vbench} & UniSim \citep{yang2024learninginteractiverealworldsimulators}, VidProM~\citep{DBLP:conf/nips/WangY24} \\
        & & AIGCBench~\citep{DBLP:conf/cvpr/FanZZW0H19}, EvalCrafter~\citep{DBLP:conf/cvpr/LiuC0WZCLZCS24} & LWM \citep{DBLP:journals/corr/abs-2402-08268}, Genesis \citep{authors2024genesis} \\
        & & LMM4LMM \citep{wang2025lmm4lmmbenchmarkingevaluatinglargemultimodal}, RISEBench \citep{zhao2025envisioningpixelsbenchmarkingreasoninginformed} & HQ-Edit \citep{DBLP:journals/corr/abs-2404-09990}, InstructPix2Pix \citep{DBLP:conf/cvpr/BrooksHE23} \\
        & & WikiVideo \citep{martin2025wikivideoarticlegenerationmultiple}, OmniDiff \citep{liu2025omnidiffcomprehensivebenchmarkfinegrained} & MagicBrush \citep{DBLP:conf/nips/ZhangMCSS23} \\
        & & ExtremeAIGC \citep{chandna2025extremeaigcbenchmarkinglmmvulnerability}, MCiteBench \citep{hu2025mcitebenchbenchmarkmultimodalcitation} & \\
        & & CompAlign~\citep{wan2025compalignimprovingcompositionaltexttoimage}, GODBench~\citep{lei2025godbenchbenchmarkmultimodallarge} & \\

        \bottomrule
    \end{tabular}}

\end{table}

\subsection{Multimodal Understanding}

Multimodal Understanding refers to the ability of models to process and interpret information from multiple modalities, such as visual and auditory data, to perform tasks that require comprehension, reasoning, and generation. These tasks are crucial for developing models capable of interacting with and responding to the real world in a more human-like manner. Based on the task definition, existing multimodal understanding tasks can be roughly categorized into two main areas: \textbf{1)} Visual-Centric Understanding, which encompasses the model's ability to understand and reason about visual content, and \textbf{2)} Audio-Centric Understanding, which focuses on tasks involving audio, such as speech, music, and environmental sounds.

\subsubsection{Visual-Centric Understanding} Visual-centric understanding evaluates a model's ability to comprehend and reason about visual data, such as images and videos, across a variety of specialized tasks. These tasks can be broadly categorized into the following domains: general visual understanding, document and chart interpretation, multilingual visual reasoning, video understanding, mathematical and scientific reasoning, and comprehensive benchmarks. Each domain addresses different facets of visual understanding, from object recognition and spatial reasoning in natural images to the interpretation of structured visual data, such as documents and graphs. Below, we explore each of these categories in detail, highlighting their key features and challenges.

\paragraph{General Visual Understanding}
General visual question-answering (VQA) datasets have evolved significantly in both complexity and scope. Early datasets, such as VQA~\citep{DBLP:journals/corr/KafleK16} and GQA~\citep{DBLP:conf/emnlp/AinslieLJZLS23}, primarily focused on object recognition, attribute identification, and simple spatial reasoning within natural images. These datasets typically contain image-question-answer triplets, with questions formatted simply (e.g., "What color is the car?"). The focus was largely on natural images and basic perception. More recent datasets, such as ALIGN~\citep{jia2021scalingvisualvisionlanguagerepresentation} aim to address more complex visual-language tasks, including image-text alignment and multimodal representations. Visual Genome~\citep{krishna2016visualgenomeconnectinglanguage} extends visual understanding by including relationships and object-level information, thus pushing the boundaries of reasoning. The LAION-400M dataset~\citep{schuhmann2021laion400mopendatasetclipfiltered}, one of the largest collections of image-text pairs, enables large-scale training for visual-language models. The LAION-5B dataset~\citep{schuhmann2022laion5bopenlargescaledataset} provides a strong dataset for large-scale image-text representations, and FILIP~\citep{yao2021filipfinegrainedinteractivelanguageimage} and YFCC100M~\citep{Thomee_2016} integrates both vision and language, enhancing models' performance across diverse benchmarks.

To further stress compositional and spatial reasoning capabilities, \text{MMSI-Bench} introduces a spatial reasoning benchmark focused on evaluating multi-modal understanding of object configurations and spatial relationships~\citep{yang2025mmsibenchbenchmarkmultiimagespatial}. \text{WikiMixQA} pushes models to integrate and synthesize multimodal information from textual and visual sources to answer complex queries that require cross-source reasoning~\citep{foroutan2025wikimixqamultimodalbenchmarkquestion}. In the mathematical domain, \text{VideoMathQA} presents expert-curated, video-based math problems across three reasoning categories—direct problem solving, conceptual transfer, and structured explanation—mirroring real-world educational settings~\citep{rasheed2025videomathqabenchmarkingmathematicalreasoning}.

\paragraph{Document, Chart, and OCR Visual Understanding}
Document, chart, and OCR-based VQA datasets form a specialized domain focusing on understanding structured visual information that includes textual elements. Document VQA, exemplified by DocVQA~\citep{DBLP:conf/wacv/MathewKJ21}, targets document understanding, requiring models to locate and interpret text within documents to answer questions. Chart VQA, such as DVQA~\citep{DBLP:conf/cvpr/KaflePCK18}, focuses on interpreting visual data representations, including bar charts, line graphs, and pie charts, testing the model's ability to understand these structures. OCR-VQA datasets like TextVQA~\citep{DBLP:conf/cvpr/SinghNSJCBPR19} and OCR-VQA~\citep{DBLP:conf/icdar/0001SSC19} emphasize reading and reasoning about text embedded within natural images. These datasets share several distinctive characteristics: 1) the critical integration of OCR with visual understanding, 2) multi-step reasoning that combines both textual and visual elements, and 3) domain-specific knowledge about document structures, chart conventions, or text layouts. Unlike general VQA datasets, these collections heavily emphasize the interplay between visual and textual content, requiring models to bridge modalities in more structured contexts. Additionally, datasets like AI2D~\citep{DBLP:journals/lre/HiippalaAHKLOTS21} focus on diagrams and structured visual representations, enhancing reasoning over graphical content. \textbf{WebUIBench} introduces a structured benchmark for planning and reasoning within digital interface environments. It evaluates MLLMs across four dimensions—WebUI perception, HTML programming, WebUI-HTML understanding, and UI-to-code generation—making it a comprehensive testbed for web interface grounding and control~\citep{lin2025webuibenchcomprehensivebenchmarkevaluating}.

\paragraph{Multilingual Visual Understanding}
Multilingual visual understanding datasets cater to the increasing demand for language diversity in multimodal systems. Datasets like CMMLU~\citep{DBLP:conf/acl/0002ZKY0GDB24}, C-Eval~\citep{DBLP:conf/nips/HuangBZZZSLLZLF23}, Exams-v~\citep{das2024exams}, M3exam~\citep{zhang2023m3exam}, VideoVista-CulturalLingo~\citep{chen2025videovistaculturallingo360circhorizonsbridgingcultures}, and MTVQA~\citep{DBLP:journals/corr/abs-2405-11985} cover beyond English-centric VQA systems. These datasets are characterized by: 1) integration of questions and annotations in multiple languages, covering various language families, 2) testing visual understanding and linguistic capabilities across different cultural contexts, and 3) requiring models to understand visual concepts that may have specific cultural interpretations or references. Unlike single-language VQA datasets, these multilingual datasets evaluate and enhance the cross-lingual transfer abilities of MLLMs.

New benchmarks also expand the scope of visual understanding to multilingual and real-world contexts. \text{CasualVQA} emphasizes causal and contextual reasoning via questions grounded in everyday scenarios, aiming to test models' deeper comprehension of real-world phenomena~\citep{causalvqa}. Meanwhile, \text{VLM@school} evaluates models' ability to integrate visual reasoning with curriculum-based subject knowledge in German, supporting multilingual educational applications~\citep{peinl2025vlmschoolevaluationai}.

\paragraph{Video Understanding}
Video understanding datasets, e.g., ActivityNet-QA~\citep{DBLP:conf/aaai/YuXYYZZT19} and Perception-Test~\citep{DBLP:conf/nips/PatrauceanS0RMB23}, are increasingly used for training and evaluating models in dynamic visual tasks. These datasets, compared to static image datasets, require models to address time-based understanding, involving dynamic visual features across multiple frames. They include annotations for actions, events, and temporal relationships, and cover diverse video durations, ranging from short clips to several-minute-long videos. Existing video evaluation datasets have expanded to tackle challenges such as the scientific domain (e.g., Video-MMMU~\citep{DBLP:journals/corr/abs-2501-13826}), long video domains (e.g., Video-MME~\citep{DBLP:journals/corr/abs-2405-21075}), and comprehensive video understanding and reasoning (e.g., VideoVista~\citep{DBLP:journals/corr/abs-2406-11303}). VideoVista provides a versatile benchmark featuring 14 categories of videos with durations from a few seconds to over 10 minutes and encompasses 19 understanding tasks and 8 reasoning tasks. It utilizes an automatic annotation framework powered by GPT-4o, enhancing its scalability and diversity. Datasets like YouTube8M~\citep{abuelhaija2016youtube8mlargescalevideoclassification} have become foundational for large-scale video classification and multimodal understanding. Additionally, VidGen-1M~\citep{tan2024vidgen1mlargescaledatasettexttovideo} and WebVid~\citep{bain2022frozentimejointvideo} serve as training datasets and focus on enhancing video comprehension by integrating multimodal text and visual signals.

 \begin{table}[t]
    \centering  
    \caption{Overview of Multimodal Benchmarks and Training Datasets, categorized by task: Reasoning (General Visual, Domain-Specific) and Planning (GUI, Embodied \& Simulated Environments). These benchmarks often require short or long reasoning for successful task completion, e.g., challenging visual and audio generation. }
    \label{tab:benchmark_datasets_2}
    \renewcommand\tabcolsep{2.5pt}
    \resizebox{\textwidth}{!}{
    \begin{tabular}{c|c|l|l}
        \toprule
         \textbf{Ability} & \textbf{Task} & \textbf{Benchmark} & \textbf{Dataset} \\
 \midrule
        \multirow{51}{*}{\makecell{Multimodal \\ Reasoning}}    
        & \multirow{21}{*}{\makecell{General Visual \\ Reasoning}}        
        & NaturalBench \citep{DBLP:conf/nips/LiLPNJMKKNR24}, VCR \citep{DBLP:conf/cvpr/ZellersBFC19} & VCR \citep{DBLP:conf/cvpr/ZellersBFC19}, TDIUC~\citep{kafle2017analysisvisualquestionanswering} \\
        & & PhysBench \citep{DBLP:journals/corr/abs-2501-16411}, MMBench \citep{DBLP:conf/eccv/LiuDZLZZYWHLCL24} & MMPR~\citep{DBLP:journals/corr/abs-2411-10442}, ChartQA \citep{DBLP:conf/acl/MasryLTJH22} \\
        & & MMMU \citep{DBLP:conf/cvpr/YueNZ0LZSJRSWYY24}, AGIEval \citep{DBLP:conf/naacl/ZhongCGLLWSCD24} & SWAG~\citep{zellers2018swaglargescaleadversarialdataset}, LLaVA-CoT~\citep{DBLP:journals/corr/abs-2411-10440} \\
        & & MMStar \citep{DBLP:conf/nips/ChenLDZZCDWQLZ24}, InfographicVQA \citep{DBLP:conf/wacv/MathewBTKVJ22} & CLEVR~\citep{johnson2016clevrdiagnosticdatasetcompositional}, Mulberry-260K~\citep{yao2024mulberry} \\
        & & VCRBench~\citep{qi2025vcrbench}, VisualPuzzles~\citep{song2025visualpuzzlesdecouplingmultimodalreasoning} & ShareGPT4oReasoning~\citep{DBLP:journals/corr/abs-2410-16198}, R1-Onevision~\citep{yang2025r1onevisionadvancinggeneralizedmultimodal} \\
        & & IV-Bench~\citep{ma2025ivbench}, VisuLogic~\citep{xu2025visulogicbenchmarkevaluatingvisual} & Video-R1-data~\citep{feng2025video}, Visual-CoT~\citep{shao2024visualcotadvancingmultimodal} \\
        & & FG-BMK \citep{yu2025benchmarkinglargevisionlanguagemodels}, Video-MMLU \citep{song2025videommlumassivemultidisciplinelecture}&  \\
        & & DVBench \citep{zeng2025visionllmsroadreadycomprehensive}, GeoSense \citep{xu2025geosenseevaluatingidentificationapplication} &  \\
        & & FLIP \cite{plesner2025flipreasoningchallenge}, ViLBench \citep{tu2025vilbenchsuitevisionlanguageprocess}&  \\
        & & HAVEN \citep{gao2025exploringhallucinationlargemultimodal}, MAGIC-VQA \citep{yang2025magicvqamultimodalgroundedinference}&  \\
        & & PM4Bench \citep{gao2025pm4benchparallelmultilingualmultimodal}, FAVOR-Bench \citep{tu2025favorbenchcomprehensivebenchmarkfinegrained}	 &  \\
        & & VLRMBenc \citep{ruan2025vlrmbenchcomprehensivechallengingbenchmark}, UrbanVideo-Bench \citep{zhao2025urbanvideobenchbenchmarkingvisionlanguagemodels}	 &  \\
        & & FortisAVQA \citep{ma2025fortisavqamavenbenchmarkdataset}, VideoVista-CulturalLingo \citep{chen2025videovistaculturallingo360circhorizonsbridgingcultures}	 &  \\
        & & CoMT \citep{cheng2025comtnovelbenchmarkchain}, CCHall \citep{zhang2025cchallnovelbenchmarkjoint}	 &  \\
        & & Wethink-dataset~\citep{yang2025wethinkgeneralpurposevisionlanguagereasoning}, MMMG~\citep{yao2025mmmgcomprehensivereliableevaluation} & \\
        & & Do You See Me~\citep{kanade2025multidimensionalbenchmarkevaluating}, SpaCE-10~\citep{gong2025space10comprehensivebenchmarkmultimodal} & \\
        & & GaslightingBench-R~\citep{zhu2025reasoningmodelseasilygaslighted}, MultiNet~\citep{guruprasad2025opensourcesoftwaretoolkit} & \\
        & & MANBench~\citep{zhou2025manbenchmultimodalmodelsmarter}, VGR -SFT~\citep{wang2025vgrvisualgroundedreasoning} & \\
        & & TempVS benchmark~\citep{song2025burnreadingmultimodallarge}, SFE benchmark~\citep{zhou2025scientistsexamprobingcognitive} & \\
        & & HSSBench~\citep{kang2025hssbenchbenchmarkinghumanitiessocial}, Argus Inspection~\citep{yao2025argusinspectionmultimodallarge} & \\
        & & 
        Chemtable~\citep{zhou2025benchmarkingmultimodalllmsrecognition}, KokushiMD-10~\citep{liu2025kokushimd10benchmarkevaluatinglarge} & \\

        \cmidrule(r){2-4}
        &\multirow{30}{*}{\makecell{Domain-specific \\ Reasoning}}   
        & MathVista \citep{DBLP:conf/iclr/LuBX0LH0CG024}, MATH-Vision \citep{DBLP:conf/nips/WangPSLRZZL24} & Habitat \citep{DBLP:conf/iccv/SavvaMPBKMZWJSL19}, AI2-THOR \citep{DBLP:journals/corr/abs-1712-05474} \\
        & & VLM-Bench \citep{DBLP:conf/nips/ZhengCJW22}, GemBench \citep{DBLP:journals/corr/abs-2410-01345} & Gibson \citep{DBLP:conf/cvpr/XiaZHSMS18}, GeoQA~\citep{chen2022geoqageometricquestionanswering} \\
        & & GeoQA~\citep{chen2022geoqageometricquestionanswering}, VIMA-Bench \citep{DBLP:journals/corr/abs-2210-03094} & Isaac Lab \citep{DBLP:journals/ral/MittalYYLRHYSGMMBSHG23}, ProcTHOR \citep{DBLP:conf/nips/DeitkeVHWESHKKM22} \\
        & & WorldSimBench \citep{DBLP:journals/corr/abs-2410-18072}, WorldModelBench \citep{DBLP:journals/corr/abs-2502-20694} & CALVIN \citep{DBLP:journals/ral/MeesHRB22}, SRM\&SRMEval \citep{miao2025boostingvirtualagentlearning} \\
        & & ScienceQA~\citep{DBLP:conf/nips/LuMX0CZTCK22}, ChartQA \citep{DBLP:conf/acl/MasryLTJH22} & SpaceR-151k~\citep{ouyang2025spacerreinforcingmllmsvideo}\\
        & & MathQA~\citep{DBLP:conf/naacl/AminiGLKCH19}, Habitat \citep{DBLP:conf/iccv/SavvaMPBKMZWJSL19} & \\
        & & AI2-THOR \citep{DBLP:journals/corr/abs-1712-05474}, Gibson \citep{DBLP:conf/cvpr/XiaZHSMS18} & \\
        & & iGibson \citep{DBLP:journals/corr/abs-2108-03272}, Isaac Lab \citep{DBLP:journals/ral/MittalYYLRHYSGMMBSHG23} & \\
        & & VCBench \citep{wang2025benchmarkingmultimodalmathematicalreasoning}, VCT \citep{götting2025virologycapabilitiestestvct} & \\
        & & 3MDBench \citep{sviridov20253mdbenchmedicalmultimodalmultiagent}, PuzzleBench \citep{zhang2025puzzlebenchfullydynamicevaluation} & \\
        & & ColorBench \citep{liang2025colorbench}, VisualPuzzles \citep{song2025visualpuzzlesdecouplingmultimodalreasoning} & \\
        & & Plot2XML \citep{cui2025drawthoughtunleashingmultimodal}, NoTeS-Bank \citep{pal2025notesbankbenchmarkingneuraltranscription} & \\
        & & EIBench \citep{lin2025feelbreakingboundariesemotional}, XLRS-Bench \citep{wang2025xlrsbenchmultimodalllmsunderstand} & \\
        & & STI-Bench \citep{li2025stibenchmllmsreadyprecise}, EgoToM \citep{li2025egotombenchmarkingtheorymind} & \\
        & & DomainCQA \citep{zhong2025domaincqacraftingexpertlevelqa},  MM-IQ \citep{cai2025mm-iq}& \\
        & & MMCR-Bench \citep{yan2025mmcradvancingvisuallanguage}, Misleading ChartQA \citep{chen2025unmaskingdeceptivevisualsbenchmarking}	 & \\
        & & FlowVerse \citep{chen2025mathflowenhancingperceptualflow}, VisNumBench \citep{weng2025visnumbenchevaluatingnumbersense} & \\
        & & MicroVQA \citep{burgess2025microvqamultimodalreasoningbenchmark}, MPBench \citep{xu2025mpbenchcomprehensivemultimodalreasoning} & \\
        & & Open3DVQA \citep{zhan2025open3dvqabenchmarkcomprehensivespatial}, ProBench \citep{yang2025probenchjudgingmultimodalfoundation} & \\
        & & Chart-HQA \citep{chen2025charthqabenchmarkhypotheticalquestion}, MMSciBench \citep{ye2025mmscibenchbenchmarkinglanguagemodels} & \\
        & & CharXiv\citep{DBLP:conf/nips/WangXH0LZLWLMCA24}, SPAR-Bench\citep{zhang2025flatland} & \\
        & & VSI-Bench\citep{yang2024thinking}, STI-Bench\citep{li2025stibenchmllmsreadyprecise} & \\
        & & VRBench~\citep{yu2025vrbench}, MATP-BENCH~\citep{he2025matpbenchmllmgoodautomated} & \\
        & & MMRefine~\citep{paik2025mmrefineunveilingobstaclesrobust}, MMR-V~\citep{zhu2025mmrvwhatsleftunsaid} & \\
        & & WikiMixQA~\citep{foroutan2025wikimixqamultimodalbenchmarkquestion}, SciVer~\citep{wang2025sciverevaluatingfoundationmodels} & \\
        & & MultiFinBen~\citep{peng2025multifinbenmultilingualmultimodaldifficultyaware}, RealHiTBench~\citep{wu2025realhitbenchcomprehensiverealistichierarchical} & \\
        & & VideoMathQA~\citep{rasheed2025videomathqabenchmarkingmathematicalreasoning}, GDI-Bench~\citep{li2025gdibenchbenchmarkgeneraldocument} & \\
        & & PhysicsArena~\citep{dai2025physicsarenamultimodalphysicsreasoning}, Spoken-MQA~\citep{wei2025spokenmathematicalreasoningbenchmarking} & \\
        & & ReasonSeg-Diff~\citep{wang2025pixelthinkefficientchainofpixelreasoning}, VF-Eval~\citep{song2025vfevalevaluatingmultimodalllms} & \\
        & & ChartMind~\citep{wei2025chartmindcomprehensivebenchmarkcomplex} & \\

        \midrule
        \multirow{15}{*}{\makecell{Multimodal \\ Planning}}    
        & \multirow{8}{*}{\makecell{GUI \\ Navigation}}        
        & WebArena \citep{DBLP:conf/iclr/ZhouX0ZLSCOBF0N24}, Mind2Web \citep{DBLP:conf/nips/DengGZCSWSS23} & AMEX~\citep{DBLP:journals/corr/abs-2407-17490}, RiCo~\citep{Deka:2017:Rico} \\
        & & VisualWebBench \citep{DBLP:journals/corr/abs-2404-05955}, OSWorld \citep{DBLP:conf/nips/XieZCLZCHCSLLXZ24} & WebSRC~\citep{DBLP:journals/corr/abs-2101-09465}, E-ANT~\citep{DBLP:journals/corr/abs-2406-14250} \\
        & & OmniACT \citep{DBLP:conf/eccv/KapoorBRKKAS24}, VisualAgentBench \citep{DBLP:journals/corr/abs-2408-06327} & AndroidEnv~\citep{DBLP:journals/corr/abs-2105-13231}, GUI-World~\citep{DBLP:journals/corr/abs-2406-10819} \\
        & & LlamaTouch \citep{DBLP:conf/uist/ZhangWJZYGLX24}, Windows Agent Arena \citep{DBLP:journals/corr/abs-2409-08264} & MBE-ARI \citep{noronha2025mbearimultimodaldatasetmapping} \\
        & & Ferret-UI \citep{DBLP:conf/eccv/YouZSWSNYG24}, WebShop \citep{DBLP:conf/nips/Yao0YN22} & \\
        & & SWE-bench M \citep{yang2024swe}, MineDojo \citep{DBLP:conf/nips/FanWJMYZTHZA22} & \\
        & & TeamCraft \citep{long2024teamcraftbenchmarkmultimodalmultiagent}, V-MAGE \citep{zheng2025vmagegameevaluationframework} & \\
        & & TongUI \citep{zhang2025tonguibuildinggeneralizedgui}, BEARCUBS \citep{song2025bearcubsbenchmarkcomputerusingweb} & \\
        & & MCA-Bench~\citep{wu2025mcabenchmultimodalbenchmarkevaluating}, ThinkGeo~\citep{shabbir2025thinkgeoevaluatingtoolaugmentedagents} & \\

        \cmidrule(r){2-4}
        &\multirow{7}{*}{\makecell{Embodied and Simulated \\ Environments}}   
        & MineDojo \citep{DBLP:conf/nips/FanWJMYZTHZA22}, MuEP \citep{DBLP:conf/ijcai/LiYZZZZYCSC0LT024} & MineDojo \citep{DBLP:conf/nips/FanWJMYZTHZA22}, Habitat 3.0 \citep{DBLP:conf/iclr/PuigUSCYPDCHMVG24} \\
        & & GVCCI \citep{DBLP:conf/iros/KimKKSZ23}, BEHAVIOR-1K \citep{DBLP:journals/corr/abs-2403-09227} & SAPIEN \citep{DBLP:conf/cvpr/XiangQMXZLLJYWY20}, HomeRobot \citep{DBLP:conf/corl/YenamandraRYWKG23} \\
        & & Habitat 3.0 \citep{DBLP:conf/iclr/PuigUSCYPDCHMVG24}, SAPIEN \citep{DBLP:conf/cvpr/XiangQMXZLLJYWY20} & HoloAssist \citep{DBLP:conf/iccv/WangKRPCABFTFJP23}, DrivingDojo \citep{DBLP:journals/corr/abs-2207-11432} \\
        & & HomeRobot \citep{DBLP:conf/corl/YenamandraRYWKG23}, HoloAssist \citep{DBLP:conf/iccv/WangKRPCABFTFJP23} & OmmiHD-Scenes~\citep{zheng2025omnihdscenesnextgenerationmultimodaldataset} \\
        & & DrivingDojo \citep{DBLP:journals/corr/abs-2207-11432}, WolfBench \citep{qiao2024benchmarking} & \\
        & & MBE-ARI \citep{noronha2025mbearimultimodaldatasetmapping}, VisEscape \citep{lim2025visescapebenchmarkevaluatingexplorationdriven} & \\
        & & AttachSecure-Bench~\citep{wang2025dinocompanionattachmenttheoryinformedmultimodal} & \\

        \bottomrule
    \end{tabular}}

\end{table}

\paragraph{Comprehensive Benchmarks}
Integrated evaluation benchmarks, such as MMBench~\citep{DBLP:conf/eccv/LiuDZLZZYWHLCL24}, Seed-Bench~\citep{DBLP:journals/corr/abs-2307-16125}, and MME-RealWorld~\citep{DBLP:journals/corr/abs-2408-13257}, \text{OmniBench} ~\citep{li2024omnibench}. This benchmark enables unified evaluation across reasoning, planning, and generation tasks, and supports platform-agnostic, scalable testing of MLLM capabilities. have emerged to provide a more holistic evaluation of existing multimodal models. These benchmarks test how well models integrate visual and linguistic understanding in real-world scenarios, including 1) multidimensional evaluation frameworks that assess various aspects of visual understanding, from perception to reasoning and knowledge integration, 2) carefully designed questions aimed at exploring specific abilities and identifying weaknesses, and 3) standardized evaluation pipelines for fair comparison across models. Unlike early task-specific datasets, these benchmarks offer a comprehensive measure of models' overall capabilities.

Visual-centric Understanding emphasizes models' abilities to process and reason about visual data, from basic object recognition in images to complex multimodal reasoning in videos and documents. By addressing various specialized tasks, such as general visual understanding, document interpretation, multilingual reasoning, and video comprehension, these benchmarks provide a comprehensive view of a model's visual capabilities. These evaluations are essential for ensuring that models can integrate visual perception with reasoning, which is critical for real-world applications.


\subsubsection{Audio-Centric Understanding}

Audio-Centric Understanding refers to the evaluation of models' capabilities in processing, interpreting, and responding to various forms of audio input, such as speech, environmental sounds, and music. As these modalities become increasingly integral to machine learning tasks, evaluating how well models understand and interact with audio data has become a key focus. The evaluation spans different aspects of speech, audio, and music understanding, with various benchmarks and datasets designed to assess accuracy, translation, emotion recognition, and general comprehension in audio-related tasks. These evaluations help gauge the effectiveness of models in understanding the full range of audio data encountered in real-world applications.

\paragraph{Speech Understanding}
Speech evaluation datasets play a crucial role in assessing models' performance in the audio domain. These datasets primarily measure whether a model can accurately and clearly understand human speech in real-world settings. Existing datasets evaluate speech understanding from several perspectives:
1) Accuracy of speech recognition: Librispeech~\citep{DBLP:conf/icassp/PanayotovCPK15} is a dataset of audiobooks read by various speakers, serving as a widely used evaluation metric for English speech recognition. Common Voice~\citep{DBLP:conf/lrec/ArdilaBDKMHMSTW20} collects voice recordings from volunteers globally, providing a diverse voice dataset for model training. The Aishell~\citep{DBLP:conf/ococosda/BuDNWZ17} series is the standard for Chinese speech recognition. Fleurs~\citep{DBLP:conf/slt/ConneauMKZADRRB22} evaluates speech recognition and speech-to-text translation models across multiple languages.
2) Speech multilingual translation tasks: CoVoST2~\citep{DBLP:journals/corr/abs-2007-10310} is a multilingual speech-to-text translation dataset that evaluates models' real-time speech recognition translation capabilities.
3) Emotion recognition: The MELD~\citep{DBLP:conf/acl/PoriaHMNCM19} dataset assesses models' ability to recognize emotions in speech, using emotional voices from multiple speakers in TV dramas. These datasets comprehensively assess models' ability to understand speech, considering factors such as content accuracy, diverse speech tasks, and additional acoustic information.

\paragraph{Audio Understanding}
Environmental sound understanding is another essential aspect of audio comprehension, involving the extraction and recognition of information from non-human voices. Compared to human speech, environmental sounds provide more complex and varied information. Mainstream evaluation datasets primarily assess audio understanding in two key areas:
1) Audio captioning: Clotho~\citep{DBLP:conf/icassp/DrossosLV20} contains sounds from free sound platforms, primarily used for the audio captioning task. Similarly, AudioCaps~\citep{DBLP:conf/naacl/KimKLK19}, sourced from the AudioSet dataset, also focuses on audio captioning and has a broader application scope.
2) Audio question answering (AQA): ClothoAQA~\citep{DBLP:conf/eusipco/LippingSDV22} is a crowdsourced dataset designed for the AQA task and AQUALLM~\citep{behera2023aquallm} is constructed by an automatic audio QA generation framework based on LLMs. These benchmarks include various audio types paired with questions and answers, helping models learn to understand audio content and generate accurate responses to audio-related questions.

\paragraph{Music Understanding}
Music, with its structural characteristics and complex variations, has become a significant area of research in audio understanding. Two primary directions are considered in music evaluation:
Mainstream datasets like MusicNet~\citep{DBLP:conf/iclr/ThickstunHK17} and NSynth~\citep{DBLP:conf/icml/EngelRRDNES17} evaluate models' ability to recognize music theory elements such as instruments, notes, pitches, and rhythms in the audio. Additionally, MusicCaps~\citep{DBLP:journals/corr/abs-2301-11325} and MusicBench~\citep{DBLP:conf/naacl/MelechovskyGGMH24} are used for captioning entire musical tracks, testing models' ability to understand both the detailed content and overall structure of music compositions.

\paragraph{Comprehensive Benchmarks}
As Large Audio-Language Models (LALMs) continue to evolve, more models now possess the ability to understand both speech and diverse sounds. Consequently, researchers are proposing new evaluation benchmarks to comprehensively assess models' audio understanding capabilities.
VoiceBench~\citep{DBLP:journals/corr/abs-2410-17196} focuses on models' ability to understand speech in varied contexts, including evaluations of basic capabilities, colloquial expressions, and performance in noisy environments.
AudioBench~\citep{DBLP:journals/corr/abs-2406-16020} integrates diverse speech tasks (e.g., Automatic Speech Recognition, Speech Question Answering), sound tasks (e.g., Audio Captioning, Audio Question Answering), and tasks related to human voices (e.g., accent, age, and gender).
Air-Bench~\citep{DBLP:conf/acl/YangXLC0ZLLZZZ24} and MMAU~\citep{DBLP:journals/corr/abs-2410-19168} expand upon this by including music tasks in their evaluations.
SD-eval~\citep{DBLP:conf/nips/AoWTCZ0W0024} combines speech tasks with environmental sound tasks, enabling models to understand complex, mixed audio scenarios.
These benchmarks not only incorporate earlier evaluation methods but also provide a more comprehensive framework for assessing speech understanding across a wide range of real-world applications.

 Audio-Centric Understanding offers a comprehensive framework for evaluating models' capabilities in processing and understanding audio data. It spans tasks from speech recognition to environmental sound and music interpretation. These evaluations are crucial for ensuring models' versatility and effectiveness in real-world applications, advancing their ability to handle complex audio data.

\subsection{Multimodal Generation}

Multimodal Generation is a key capability of Multimodal Reasoning Models, encompassing the creation of novel content across different data types, such as text, images, audio, or video. This generative ability is critical not only for creative applications but also for tasks where models need to communicate their understanding or reasoning results in a multimodal format.

These tasks can be broadly categorized based on how information flows between modalities and the nature of the generated output: \textbf{(1)} Cross-modal Generation, which evaluates a model's ability to generate content in one modality based on input from another; and \textbf{(2)} Joint Multimodal Generation, which assesses a model's ability to simultaneously generate content across multiple modalities.

\subsubsection{Cross-modal Generation}

Cross-modal generation involves tasks where models generate content in one modality based on input from another. This includes tasks like text-to-image, text-to-video, and text-to-speech generation, where models must effectively map one type of input (e.g., text) to a different form (e.g., image, video, or speech). These tasks challenge models to transform and align information from one modality to another, often requiring the handling of complex or conditional prompts. In this section, we explore how datasets and benchmarks have been developed to evaluate model performance across various cross-modal tasks, focusing on alignment, coherence, and semantic generation.

\paragraph{Text to Image}

The field of text-to-image generation (T2I) has seen significant advancements, driven by diverse datasets and benchmarks tailored to tasks such as text-to-image generation, editing, and conditional generation.
For text-to-image generation, datasets like MSCOCO (30K) \citep{DBLP:conf/eccv/LinMBHPRDZ14}, CC12M \citep{changpinyo2021conceptual}, and Flickr30k~\citep{DBLP:journals/ijcv/PlummerWCCHL17} offer large-scale, general-purpose image-text pairs, emphasizing everyday scenes and objects. In contrast, datasets like RedCaps~\citep{DBLP:conf/nips/DesaiKA021} and COMMONPOOL~\citep{DBLP:conf/nips/GadreIFHSNMWGZO23} introduce more complex text descriptions and higher-resolution images. Benchmarks such as GenEval~\citep{DBLP:conf/nips/GhoshHS23} and ELLA~\citep{DBLP:journals/corr/abs-2403-05135} focus on evaluating text-to-image alignment, assessing how accurately the generated images match the textual descriptions. Meanwhile, GenAI-Bench~\citep{DBLP:journals/corr/abs-2406-13743} and T2I-CompBench++~\citep{huang2023t2icompbench} emphasize the handling of complex prompts and object interactions, highlighting the need for effective compositional generation and improved semantic alignment.

For text-to-image editing, datasets like MagicBrush~\citep{DBLP:conf/nips/ZhangMCSS23}, InstructPix2Pix~\citep{DBLP:conf/cvpr/BrooksHE23}, and HQ-Edit~\citep{DBLP:journals/corr/abs-2404-09990} focus on instruction-based editing, with HQ-Edit extending tasks to high-definition images. UltraEdit~\citep{zhao2024ultraedit} and SEED-Data-Edit~\citep{ge2024seeddataedit} introduce multi-turn editing tasks, improving training for large language models (LLMs) in multi-turn dialogues. These datasets assess the varying demands of image editing, with MagicBrush evaluating creative aspects and Emu Edit~\citep{sheynin2023emu} focusing on precision and coherence in high-quality edits.

For conditional text-to-image generation, datasets like ADE20K~\citep{zhou2017scene} and CocoStuff~\citep{caesar2018coco} offer detailed segmentation maps and scene parsing annotations, enabling models to generate images with specific scene structures. UniControl~\citep{qin2023unicontrol} introduces more comprehensive data, requiring models to handle multiple conditional inputs simultaneously. Benchmarks like UniCombine~\citep{wang2025unicombine} focus on evaluating instruction execution completeness, visual coherence, and consistency with constraints.

\paragraph{Text to Video}
In text-to-video generation, high-quality datasets and comprehensive benchmarks are critical for advancing research. Datasets like VidGen-1M~\citep{tan2024vidgen1mlargescaledatasettexttovideo}, OpenVid-1M~\citep{DBLP:journals/corr/abs-2407-02371}, and VidProM~\citep{DBLP:conf/nips/WangY24} cover a wide range of video content and corresponding descriptive texts. Benchmarking tools such as AIGCBench~\citep{DBLP:conf/cvpr/FanZZW0H19}, EvalCrafter~\citep{DBLP:conf/cvpr/LiuC0WZCLZCS24}, and VBench~\citep{DBLP:conf/cvpr/HuangHYZS0Z0JCW24} evaluate models across various metrics like relevance, coherence, and visual quality. Specialized benchmarks like VideoScore~\citep{DBLP:conf/emnlp/HeJZKSSCCJAWDNL24}, WorldSimBench~\citep{DBLP:journals/corr/abs-2410-18072}, and WorldScore~\citep{duan2025worldscore} expand evaluation to cover video quality and real-world accuracy, with VideoScore assessing user satisfaction.

\paragraph{Text to Speech}

Text-to-speech (TTS) generation has benefited from high-caliber datasets and benchmarks that enable the development of Large Audio-Language Models (LALMs). Early models used synthetic datasets to evaluate speech dialogue capabilities, employing datasets like LlaMA-Questions~\citep{Nachmani2024llamaq}, Web Questions~\citep{Berant2013webq}, and Trivia QA~\citep{Joshi2017TriviaQA}. Evaluations were based on comparing word error rates and accuracy between text and audio outputs. Recent benchmarks like ADU-Bench~\citep{Gao2024ADUBench} assess speech dialogue capabilities across regular, professional, multilingual, and ambiguous scenarios, while URO-Bench~\citep{Yan2025UROBench} includes evaluations of speech style, such as intonation and emotion.

\paragraph{Robotics}

In robotics, datasets and benchmarks provide high-fidelity, multi-modal environments for evaluating model performance. Datasets like ThreeDWorld~\citep{DBLP:conf/nips/GanSAMSTFKBHSKW21} and GAIA-1~\citep{DBLP:journals/corr/abs-2309-17080} offer interactive simulation platforms for robotics tasks like autonomous driving. On the benchmark side, Genesis~\citep{engelcke2019genesis} provides a standardized evaluation framework to assess models across a range of robotics tasks, ensuring real-world applicability.

In summary, cross-modal generation is a pivotal area of multimodal AI, focusing on tasks such as text-to-image, text-to-video, and text-to-speech generation. These tasks challenge models to transform and align information across modalities. As advancements continue, the focus is on improving the handling of complex prompts, multi-step reasoning, and semantic alignment, with models poised to perform increasingly sophisticated transformations and interactions across modalities.

\subsubsection{Joint Multimodal Generation}

Joint multimodal generation refers to the simultaneous creation of content across multiple modalities, such as generating both text and images or combining text, audio, and video into a cohesive output. This presents additional complexity as models must ensure coherence and alignment between the generated modalities. Tasks like text-to-interleaved image-text and text-to-multimodal output exemplify this, requiring models to generate content that complements and fits within the broader context of the narrative. Specialized datasets and benchmarks have been developed to support these tasks, providing a rich environment for training models to create contextually relevant multimodal outputs.

\paragraph{Text to Interleaved Image-Text}

The development of multimodal large language models (MLLMs) has significantly advanced interleaved image-text generation, with datasets like MM-Interleaved~\citep{tian2024mm} and ANOLE~\citep{chern2024anole} supporting model training with high-quality annotated image-text pairs. These datasets emphasise the need for models to generate contextually relevant and visually coherent content. Benchmarks like InterleavedEval~\citep{liu2024holistic} and OpenLEAF~\citep{an2024openleaf} focus on evaluating models’ ability to generate coherent and aligned image-text pairs, while OpenING~\citep{zhou2024GATE} provides a more diverse set of tasks to assess interleaved image-text generation.

\paragraph{Text to Multimodal Output}

Recent developments in text-to-multimodal output focus on enhancing multimodal generation by combining cross-modal and joint multimodal data. Models like NextGPT~\citep{wu24next} and DreamFactory~\citep{xie2024dreamfactory} leverage training-free approaches to transform text into multimodal stories, integrating video evaluation benchmarks like Vbench. Other models, such as EVA~\citep{chi2024eva}, incorporate embodied world models to simulate and anticipate events in video sequences based on text inputs.

In summary, joint multimodal generation involves the simultaneous creation of content across multiple modalities, requiring models to maintain coherence and alignment between them. As research advances, future developments will likely focus on improving intermodal coherence, adaptability, and seamless generation, opening up new possibilities for dynamic, multi-dimensional content creation and interactive user experiences.

\subsection{Multimodal Reasoning}

Multimodal reasoning goes beyond simple understanding or generation by requiring models to integrate information from multiple modalities. This allows them to make inferences, solve problems, and answer complex questions that demand a deeper comprehension of the relationships between different types of data.

We can broadly categorize multimodal reasoning models into two primary categories: \textbf{(1)} General Visual Reasoning, which evaluates a model’s ability to understand visual content and apply general knowledge, logic, and common sense to solve tasks; and \textbf{(2)} Domain-specific Reasoning, which evaluates specific, often more technical, reasoning abilities such as mathematical problem-solving based on visual input.

\subsubsection{General Visual Reasoning}

General visual reasoning is one of the most critical capabilities in Multimodal Reasoning Models. It requires models not only to perceive visual information but also to comprehend, analyse, and reason about it using extensive knowledge, logical deduction, and common sense across a variety of scenarios.

To rigorously assess this ability, a wide range of benchmarks has been developed, each targeting distinct aspects of visual reasoning. Moving beyond simple question answering tasks (e.g., VQA), Visual Commonsense Reasoning benchmarks like VCR \citep{DBLP:conf/cvpr/ZellersBFC19}, and specialised datasets like PhysBench \citep{DBLP:journals/corr/abs-2501-16411} for physical reasoning, and VideoPhy \citep{DBLP:journals/corr/abs-2406-03520} for understanding physical common sense in videos, challenge models to apply everyday knowledge to interpret visual situations.

Ambitions for broader AI capabilities are reflected in Multimodal General Intelligence Benchmarks. These include comprehensive evaluations like MMBench \citep{DBLP:conf/eccv/LiuDZLZZYWHLCL24} (covering multilingual aspects), MMMU \citep{DBLP:conf/cvpr/YueNZ0LZSJRSWYY24} (spanning diverse disciplines), AGIEval \citep{DBLP:conf/naacl/ZhongCGLLWSCD24} (focused on human-centric evaluation), VideoVista \citep{DBLP:journals/corr/abs-2406-11303} and MMStar \citep{DBLP:conf/nips/ChenLDZZCDWQLZ24} (video-centric). These benchmarks incorporate visual reasoning as a key component alongside other modalities and tasks. Additionally, visual reasoning over diagrams and structured visuals is crucial, with benchmarks like AI2D \citep{DBLP:conf/eccv/KembhaviSKSHF16} and InfographicVQA \citep{DBLP:conf/wacv/MathewBTKVJ22} challenging models to interpret spatial layouts, understand relationships, and extract information from diagrams, charts, and infographics.

A critical element in these benchmarks is the datasets used for training and evaluating models. Several datasets, such as SWAG \citep{zellers2018swaglargescaleadversarialdataset}, are designed to train models to predict the likely continuation of actions in visual scenes. The LLava-CoT dataset \citep{DBLP:journals/corr/abs-2411-10440} enables models to reason about visual commonsense tasks by integrating large language models. CLEVR \citep{johnson2016clevrdiagnosticdatasetcompositional} challenges models to perform complex reasoning on synthetic images of everyday objects. Other datasets like Mulberry-260K \citep{yao2024mulberry} and ShareGPT4oReasoning \citep{DBLP:journals/corr/abs-2410-16198} further train models for visual commonsense reasoning and multimodal dialogues, respectively.

Video-R1-data \citep{feng2025video} helps train models for reasoning about dynamic visual content in video sequences. Finally, Visual-CoT \citep{shao2024visualcotadvancingmultimodal} supports training models requiring both visual understanding and reasoning across a variety of tasks. This dynamic and ever-evolving landscape of benchmarks and datasets is essential for advancing multimodal reasoning models.

\subsubsection{Domain-specific Reasoning}

Domain-specific reasoning benchmarks play a crucial role in evaluating the specialized reasoning capabilities of multimodal models in specific fields. For mathematical reasoning, datasets like MathVista \citep{DBLP:conf/iclr/LuBX0LH0CG024} and MATH-Vision \citep{DBLP:conf/nips/WangPSLRZZL24} assess a model's ability to solve mathematical problems in visual contexts, requiring both visual understanding and mathematical inference. Similarly, benchmarks like ChartQA \citep{DBLP:conf/acl/MasryLTJH22}, ScienceQA \citep{DBLP:conf/nips/LuMX0CZTCK22} and CharXiv \citep{DBLP:conf/nips/WangXH0LZLWLMCA24} focus on reasoning in specific domains.

In robotics, several benchmarks assess different aspects of embodied AI with a strong emphasis on reasoning. Simulation environments such as Habitat \citep{DBLP:conf/iccv/SavvaMPBKMZWJSL19}, AI2-THOR \citep{DBLP:journals/corr/abs-1712-05474}, and iGibson \citep{DBLP:journals/corr/abs-2108-03272} require agents to reason about navigation, interaction, and spatial understanding in complex 3D settings. Benchmarks like Isaac Lab \citep{DBLP:journals/ral/MittalYYLRHYSGMMBSHG23} and ProcTHOR \citep{DBLP:conf/nips/DeitkeVHWESHKKM22} focus on reasoning for manipulation tasks in diverse environments. Others, such as WebArena \citep{DBLP:conf/iclr/ZhouX0ZLSCOBF0N24}, test reasoning about web content, while language-guided reasoning is evaluated through benchmarks like CALVIN \citep{DBLP:journals/ral/MeesHRB22}.

For physical reasoning, datasets like PhysBench \citep{DBLP:journals/corr/abs-2501-16411}, VideoPhy \citep{DBLP:journals/corr/abs-2406-03520}, and CRAVE \citep{sun2025contentrichaigcvideoquality} assess models' understanding of physical laws and common sense across visual and video contexts. Finally, benchmarks like GAIA-1 \citep{DBLP:journals/corr/abs-2309-17080} and RoboGen \citep{DBLP:conf/icml/WangXCWWFEHG24} support the development of world models by evaluating how well models can simulate and reason about real-world dynamics and interactions.

Several datasets target reasoning in specific high-value domains. \text{WeThink-Dataset} contributes a richly annotated dataset with explicit reasoning paths, enabling instruction tuning and reinforcement learning for vision-centric reasoning tasks~\citep{yang2025wethinkgeneralpurposevisionlanguagereasoning}. In the IT domain, \text{MMMG} introduces a novel dataset for multimodal knowledge graph generation, advancing knowledge-structured reasoning in technical contexts~\citep{yao2025mmmgcomprehensivereliableevaluation}. \text{SciVer} provides 3{,}000 expert-labeled examples for multimodal scientific claim verification across four distinct reasoning types derived from real research papers~\citep{wang2025sciverevaluatingfoundationmodels}.

These domain-specific benchmarks are crucial for pushing the boundaries of multimodal reasoning in specialized areas, enabling the development of more capable and intelligent multimodal reasoning models for specific applications.

In summary, multimodal reasoning is a critical area of AI that requires models to integrate and reason across multiple modalities, such as text, images, and video, to solve complex tasks. It is divided into General Visual Reasoning, which applies logic and common sense to visual content, and Domain-specific Reasoning, which evaluates specialized reasoning abilities in fields like mathematics, robotics, and physical laws. These tasks continually push multimodal reasoning models to evolve and approach human-level reasoning. As the field progresses, the future of multimodal reasoning will focus on creating more integrated systems capable of generalizing across diverse tasks and real-world scenarios, enabling more adaptive, intelligent, and versatile AI solutions.

\subsection{Multimodal Planning}

Multimodal planning benchmarks are essential for evaluating agents' abilities to integrate and process diverse inputs—such as visual, textual, and interactive data—while performing complex, multi-step tasks. These benchmarks cover a wide range of challenges, including web navigation, graphical user interfaces (GUIs), embodied environments, and open-ended simulations. By testing planning, reasoning, and adaptability, they provide a comprehensive view of an agent's capabilities. We categorize these benchmarks into two key areas to highlight their unique contributions and innovations.

\subsubsection{GUI Navigation}

Benchmarks in GUI navigation assess agents' abilities to plan and execute tasks across digital interfaces, requiring robust visual-language grounding and multi-step reasoning. {WebArena}~\citep{DBLP:conf/iclr/ZhouX0ZLSCOBF0N24} and {Mind2Web}~\citep{DBLP:conf/nips/DengGZCSWSS23} offer realistic web environments for navigation and information extraction, with {Mind2Web} further introducing cross-website tasks to test generalizability. {VisualWebBench}~\citep{DBLP:journals/corr/abs-2404-05955} advances visual-intensive planning with 1.5K tasks focused on cross-page integration and element localisation. {Windows Agent Arena}~\citep{DBLP:journals/corr/abs-2409-08264} evaluates cross-application planning in desktop environments, while {Ferret-UI}~\citep{DBLP:conf/eccv/YouZSWSNYG24} focuses on grounded UI understanding for executing multi-step instructions. Benchmarks like {WebShop}~\citep{DBLP:conf/nips/Yao0YN22} test visual-language grounding in simulated e-commerce environments. Similarly, {OSWorld}~\citep{DBLP:conf/nips/XieZCLZCHCSLLXZ24} and {OmniACT}~\citep{DBLP:conf/eccv/KapoorBRKKAS24} provide real desktop OS environments, supporting cross-application workflows such as file manipulation and data processing. {VisualAgentBench}~\citep{DBLP:journals/corr/abs-2408-06327} extends this paradigm by systematically evaluating large multimodal models across GUI, embodied, and visual design tasks, establishing a unified benchmark for planning and acting in visually rich digital environments. This is complemented by benchmarks like {LlamaTouch}~\citep{DBLP:conf/uist/ZhangWJZYGLX24}, which scales mobile UI automation with 495 tasks, testing multi-step operations such as app navigation.

\subsubsection{Embodied and Simulated Environments}

Embodied and simulated environments emphasize planning in dynamic, interactive settings, where agents must adapt to physical or virtual worlds. {MineDojo}~\citep{DBLP:conf/nips/FanWJMYZTHZA22} provides an open-ended benchmark in Minecraft, enabling the training and evaluation of generalist agents across diverse tasks in a rich, interactive environment. Its flexibility supports multimodal planning for object interaction, navigation, and resource management. {MuEP}~\citep{DBLP:conf/ijcai/LiYZZZZYCSC0LT024} focuses on embodied planning with visual-language inputs for tasks like path planning in simulated environments. {GVCCI}~\citep{DBLP:conf/iros/KimKKSZ23} introduces a lifelong learning framework that generates synthetic data to enhance visual grounding for language-guided robotic manipulation, achieving significant performance gains without human supervision. {BEHAVIOR-1K}~\citep{DBLP:journals/corr/abs-2403-09227} offers a dataset of 1,000 household activities, enabling robots to plan complex tasks by integrating visual, semantic, and action data. {Habitat 3.0}~\citep{DBLP:conf/iclr/PuigUSCYPDCHMVG24} advances human-robot collaboration in simulated homes, supporting multimodal planning for navigation and interaction. {SAPIEN}~\citep{DBLP:conf/cvpr/XiangQMXZLLJYWY20} provides a high-fidelity environment for part-based object manipulation, enhancing robotic planning precision. {HomeRobot}~\citep{DBLP:conf/corl/YenamandraRYWKG23} and its {OpenVocabManip} benchmark~\citep{DBLP:journals/corr/abs-2407-06939} pioneer open-vocabulary mobile manipulation, combining language, perception, and action for generalizable tasks. {HoloAssist}~\citep{DBLP:conf/iccv/WangKRPCABFTFJP23} captures egocentric human-robot interactions, facilitating planning for real-world collaborative tasks. {DrivingDojo}~\citep{DBLP:journals/corr/abs-2207-11432} tests dynamic decision-making in real-time driving scenarios using video and multi-agent data. Finally, {V-MAGE}~\citep{zheng2025vmagegameevaluationframework} presents a game-based evaluation framework to assess Multimodal Large Language Models (MLLMs) on tasks like positioning, trajectory tracking, and visual memory, offering a novel approach to quantify planning abilities.

Multimodal planning benchmarks have made significant progress in evaluating agents across diverse tasks, from web navigation to embodied environments. However, challenges remain, such as long-horizon planning, handling noisy inputs, and real-world adaptability. Future benchmarks should focus on open-world environments, real-time human feedback, and collaborative planning, particularly in multi-agent or human-AI scenarios. Addressing these gaps will help advance the development of agents capable of handling unpredictable, real-world tasks with greater flexibility and generalization.

\subsection{Evaluation Method}


The mainstream evaluation methods currently include Exact/Fuzzy Match, Option Matching, LLM/MLLM Scoring, and Agentic Evaluation. 

\paragraph{Exact/Fuzzy Matching} Exact/Fuzzy Matching is primarily used in general open-ended VQA tasks including VQAv2~\citep{VQA}, OKVQA~\citep{DBLP:conf/cvpr/MarinoRFM19}. These evaluation datasets typically provide multiple human-annotated candidate answers, and the predicted answers, processed by rules, are matched against the candidate answers either exactly or fuzzily. The final evaluation score is then calculated based on certain rules. For example, in VQAv2~\citep{VQA} evaluation, a match with a single candidate answer is worth only 1/3 of a point, and a full score of 1 point requires a match with all three candidate answers; DocVQA~\citep{DBLP:conf/wacv/MathewKJ21}, on the other hand, uses Levenshtein distance to measure the accuracy of the predicted results.

\paragraph{Options Matching}
Due to the diversity of answers, exact and fuzzy matching methods are often unable to encompass all candidate options. To ensure fairness and accuracy in evaluation, the Options Matching approach was introduced. In this method, the system prompt includes several candidate options, and the model is required to select the most appropriate one. Moreover, to reduce the possibility of the model exhibiting a preference for a specific option during the selection process, works such as MMBench~\citep{DBLP:conf/eccv/LiuDZLZZYWHLCL24} have adopted the CircularEval methodology to minimise stochastic variations in the evaluation.

\paragraph{LLM/MLLM Scoring}
Although option selection ensures fairness, it deviates considerably from the nature of open-ended questions and real-world scenarios. As a result, LLM-based evaluation methods have been introduced into the assessment of open-ended questions~\citep{fu-etal-2024-gptscore,zhang2023gpt}. This approach involves inputting specific prompts, questions, standard answers, and model predictions into an LLM or MLLM, such as GPT-4o, to generate scores~\citep{mllm_as_judge,xu-etal-2024-multiskill,saad-falcon-etal-2024-ares}. The prompts typically include scoring guidelines, reference examples, and other information designed to guide the model toward providing fair and balanced scores.

\paragraph{Agentic Evaluation}
During the evaluation process, the capabilities of a single model are inherently limited, which may lead to shortcomings when processing diverse multimodal information. 
To this end, agent-based approaches can leverage tools to mitigate the inherent limitations of the model itself. 
For instance, CIGEval~\citep{wang2025cigeval} expands the visual understanding capabilities of MLLMs by integrating a multi-functional toolbox, thereby enabling more fine-grained evaluation.
Moreover, multi-agent discussions have shown effectiveness in downstream tasks by fostering consensus and producing more robust solutions~\citep{xu2023reasoninglargelanguagemodels,chen-etal-2024-reconcile,xu2025filmagent},
a benefit that also extends to evaluation settings.
Methods that leverage cooperative or adversarial interactions among multiple agents to assess outputs have demonstrated more reliable and interpretable evaluations~\citep{chan2024chateval,li2024prd,zhao2024autoarena,liang-etal-2024-abseval}.

\section{Conclusion}
\label{sec: conclusion}

In this paper, we survey the evolution of multimodal reasoning models, highlighting pivotal advancements and paradigm-shifting milestones in the field. While current models predominantly adopt a ‌language-centric reasoning paradigm‌, delivering impressive results in tasks like visual question answering, visual math, and video understanding. Notably, ‌visual-centric long reasoning‌ (e.g., understanding 3D contexts, addressing complex visual information-seeking questions) and ‌interactive multimodal reasoning‌ (e.g., dynamic cross-modal dialogue or iterative feedback loops) remain underdeveloped frontiers requiring deeper exploration.

Building on empirical evaluations and experimental insights, we propose a forward-looking concept for ‌inherently  large multimodal models‌ that transcend language-dominated architectures. Such models should prioritize three core capabilities:
‌Multimodal Agentic Reasoning‌: Enabling proactive environmental interaction (e.g., embodied AI agents that learn through real-world trial and error).
‌Omini-Modal Understanding and Generative Reasoning: Integrating any-modal semantics (e.g., aligning abstract concepts across vision, audio, and text) while resolving ambiguities in complex, open-world contexts; Producing coherent, context-aware outputs across modalities (e.g., generating diagrams from spoken instructions or synthesizing video narratives from text).
By addressing these dimensions, future models could achieve human-like contextual adaptability, bridging the gap between isolated task performance and generalized, real-world problem-solving.




\newpage
\bibliographystyle{lmrm}
\bibliography{refs}

\end{document}